\documentclass{article}

\usepackage[utf8]{inputenc}
\usepackage[T1]{fontenc}
\usepackage{graphicx}
\usepackage[textwidth=6in,textheight=9in,centering]{geometry}
\usepackage{amsmath, amssymb}
\usepackage{booktabs}
\usepackage{comment}
\usepackage{dirtytalk}
\usepackage{longtable} %
\usepackage{array}
\usepackage{caption} %
\usepackage{adjustbox} %
\usepackage{appendix} %
\newcolumntype{H}{>{\setbox0=\hbox\bgroup}c<{\egroup}@{}}
\usepackage{multirow} %
\usepackage{makecell}
\usepackage{enumitem}
\usepackage[load-configurations=version-1]{siunitx}
\usepackage[round,authoryear]{natbib}

\usepackage[colorlinks=true,linkcolor=blue,citecolor=blue,urlcolor=blue]{hyperref}
\usepackage{xcolor}

\title{Assessing the Real-World Utility of Explainable AI for Arousal Diagnostics: An Application-Grounded User Study}

\author{%
  Stefan Kraft\thanks{Also: University of Tübingen, GER} \\
  IT-Designers Gruppe\\
  Esslingen am Neckar, GER \\
  \texttt{stefan.kraft@it-designers.de} \\
  \and
  Andreas Theissler\\
  University of Giessen\\
  Giessen, GER\\
  \texttt{https://orcid.org/0000-0003-0746-0424} \\
  \and
  Vera Wienhausen-Wilke \\
  Klinikum Esslingen, Klinik für Kardiologie, Pneumologie und Angilologie \\
  Esslingen am Neckar, GER \\
  \texttt{v.wienhausen-wilke@klinikum-esslingen.de} \\
  \and
  Gjergji Kasneci \\
  Technical University of Munich \\
  Munich, GER \\
  \texttt{gjergji.kasneci@tum.de} \\
  \and
  Hendrik Lensch \\
  University of Tübingen\\
  Tübingen, GER \\
  \texttt{hendrik.lensch@uni-tuebingen.de} \\
}

\date{}

\begin{document}

\maketitle

\begin{abstract}
Artificial intelligence (AI) systems increasingly match or surpass human experts in biomedical signal interpretation. However, their effective integration into clinical practice requires more than high predictive accuracy. Clinicians must discern \textit{when} and \textit{why} to trust algorithmic recommendations.
This work presents an application-grounded user study with eight professional sleep medicine practitioners, who score nocturnal arousal events in polysomnographic data under three conditions: (i) manual scoring, (ii) black-box (BB) AI assistance, and (iii) transparent white-box (WB) AI assistance. Assistance is provided either from the \textit{start} of scoring or as a post-hoc quality-control (\textit{QC}) review.
We systematically evaluate how the type and timing of assistance influence event-level and clinically most relevant count-based performance, time requirements, and user experience. When evaluated against the clinical standard used to train the AI, both AI and human-AI teams significantly outperform unaided experts, with collaboration also reducing inter-rater variability. Notably, transparent AI assistance applied as a targeted QC step yields median event-level performance improvements of approximately 30\% over black-box assistance, and QC timing further enhances count-based outcomes.
While WB and QC approaches increase the time required for scoring, start-time assistance is faster and preferred by most participants. Participants overwhelmingly favor transparency, with seven out of eight expressing willingness to adopt the system with minor or no modifications.
In summary, strategically timed transparent AI assistance effectively balances accuracy and clinical efficiency, providing a promising pathway toward trustworthy AI integration and user acceptance in clinical workflows.
\end{abstract}

\section{Introduction}\label{sec:introduction}

Artificial Intelligence (AI) has demonstrated impressive diagnostic capabilities in medicine, often achieving or surpassing expert-level performance in fields such as radiology, pathology, dermatology, ophthalmology, and cardiology~\citep{topol2019high}. However, successful deployment in clinical settings depends not only on high predictive accuracy but also on meaningful integration into medical workflows~\citep{fawzy2023ethics}.

In sleep medicine, one key diagnostic task is the analysis polysomnographic (PSG) data, where the scoring of arousal events -- brief electroencephalographic (EEG)-based disruptions of sleep -- is essential for diagnosing sleep disorders such as obstructive sleep apnea (OSA) or periodic limb movement syndrome~\citep{berry2012rules, franklin2015obstructive}. OSA alone affects approximately 20\% of the population and is associated with severe health consequences~\citep{franklin2015obstructive,wetter2012elsevier}.

Arousal scoring, traditionally performed by trained experts, is a time-consuming process with high inter-rater variability, particularly across different institutions~\citep{chylinski2020validation,ehrlich2024state}. 
This variability in scoring quality, combined with high demand for PSG examinations, leading to long waiting times, creates an urgent need for reliable clinical decision support systems (CDSS) that can augment human expertise while maintaining clinical standards.
In response, deep learning models like \textit{DeepSleep} have emerged, achieving state-of-the-art performance on public PSG datasets and showing promise for clinical application~\citep{li2021deepsleep,ehrlich2024state}.

Despite these advances, real-world adoption of AI-based CDSS is hindered by the lack of interpretability, understandability, and transparency, which undermine accountability and reduce trust in predictive outcomes~\citep{fawzy2023ethics}.

\paragraph{Explainability and the evaluation gap.}
Explainable AI (XAI) offers the potential to enhance decision-makers' validation capabilities, improve outcome quality, foster trust, and facilitate accountability in collaborative decision-making~\citep{mersha2024explainable}.

However, the effectiveness of explanations in clinical settings remains poorly understood, with conflicting evidence from existing studies.
Recent XAI research reveals that explanations don't always improve human-AI collaboration performance. For instance, \citet{bansal2021does} found that explanations are beneficial only when AI systems outperform humans working independently. Additionally, studies by \citet{poursabzi2021manipulating} and \citet{panigutti2022understanding} indicate that explanations can hinder error detection due to automation bias. Furthermore, \citet{schmidt2020calibrating} demonstrated that explanations of incorrect predictions can increase algorithmic bias among risk-averse users.

The evaluation of the utility of AI explanations is therefore a crucial aspect of explainable AI (XAI) research, and there is ongoing debate about the best approach to measure the effectiveness of these explanations~\citep{amarasinghe2023explainable}. One of the most widely accepted methods is the use of \textit{application-grounded} evaluations, which involve real users performing real tasks in real-world environments~\citep{amarasinghe2023explainable, doshi2017towards}. They are often considered the gold standard for evaluating AI explanations~\citep{gunning2021darpa} because they offer the most authentic assessment of how explanations affect real-world outcomes. By involving domain experts in actual tasks, these studies provide insights into the practical utility of explanations in professional environments, where the stakes are often high.

Yet, only about 20\% of XAI evaluations apply user studies~\citep{rong2022towards, nauta2023anecdotal} at all and only about 5\% conduct application-grounded user studies~\citep{rong2022towards, nauta2023anecdotal}. The rest of the evaluations are \textit{functionally-} or \textit{human-grounded} evaluations with proxy tasks or lay participants or are even just evaluated with anecdotal evidence~\citep{nauta2023anecdotal}. This scarcity of real-world, task-specific evaluations leaves a significant gap in understanding how well AI explanations actually perform in practice, especially in high-stakes environments where expert users are essential.

We assert that compelling evidence for the utility of AI explanations demands \textit{application-grounded} studies that integrate objective task metrics with subjective user feedback within the intended environment. This methodology should adhere to guidelines derived from best practices in human-AI interaction, as discussed for example by \citet{rong2022towards}. The significance of these best practices is underscored in recent research by \citet{amarasinghe2024importance}, who illustrates how selecting evaluation metrics that align with business objectives and acknowledge real-world decision-making options, such as deferring a decision to a senior expert, can fundamentally alter the conclusions drawn about the utility of explanations.

\paragraph{Contributions of this work.}
This study addresses this gap in the evaluation of the utility of AI explanations for the high-stakes task of arousal scoring and makes three specific contributions:
\begin{enumerate}
    \item We incorporate \textit{DeepSleep}, a state-of-the-art arousal detection tool, into a web-based Decision Support System, specifically designed for the task of arousal scoring. This system can function as a black box or provide comprehensive explanations 
    presented with varying levels of detail.
    \item We design and conduct an application-grounded user study involving eight professional sleep scorers. For each participant, the study spans approximately twelve hours of physiological timeseries data for arousal event scoring and includes detailed questionnaire data.
    \item We examine how \textit{transparency} (white- vs. black box) and \textit{timing} (assistance from the start vs.\ post-hoc quality control) influence (i) diagnostic performance against two complementary ground truths, (ii) time efficiency, and (iii) user-centred factors such as trust, comfort, and perceived plausibility.
\end{enumerate}

\paragraph{Research questions.}
Our investigation is structured around three research questions:
\begin{description}
    \item[RQ1:] Does AI assistance improve human arousal scoring performance and efficiency relative to unaided experts?
    \item[RQ2:] Does transparent (\textit{white box}) assistance outperform opaque (\textit{black box}) assistance in terms of objective performance and subjective acceptance?
    \item[RQ3:] When is assistance most beneficial? During the initial scoring pass or as a post-hoc quality control step?
\end{description}

\subsection{Related Work.}

\paragraph{Evolution of Automated Arousal Detection.}

Early approaches to automated arousal detection were based on traditional machine learning techniques using hand-crafted features~\citep{zan2023multi}. However, these methods were quickly outperformed by deep learning models utilizing end-to-end architectures~\citep{li2021deepsleep}.
A major catalyst in the field was the 2018 PhysioNet Challenge on arousal detection~\citep{ghassemi2018you, goldberger2000physiobank}, where the winning solution introduced \textit{DeepSleep}, a fully convolutional neural network architecture~\citep{li2021deepsleep}, setting new standards for segmentation accuracy and throughput. 
Subsequent research has built on these foundations, with continued innovation in deep learning architectures driving further improvements in detection performance~\citep{zan2023multi,kuo2023machine, badiei2023novel}.

Despite these technical advances, the evaluation of arousal detection models has historically prioritized methodological benchmarks over clinical applicability. Only recently has attention turned to assessing these models in real-world settings~\citep{ehrlich2024state,pmlr-v287-kraft25a}. Notably, previous evaluation protocols often disregarded prevalent clinical practices, such as annotating only arousal onset points rather than complete events, and were not guided by clinically relevant considerations~\citep{pmlr-v287-kraft25a}. 
This resulted in fragmented evaluation practices, relying on technically motivated window-based or pointwise schemes, while overlooking clinically important factors such as accurate event counts and the selection of meaningful decision thresholds~\citep{pmlr-v287-kraft25a}.

\paragraph{Most Similar Prior Work: Ehrlich et al. 2024.}
Among existing studies, the work by \citet{ehrlich2024state} represents the most thorough and clinically-relevant assessment of automated arousal detection to date. Their research bridges the divide between technical performance and clinical utility by comprehensively evaluating a U-Net-based detector derived from the \textit{DeepSleep} architecture~\citep{li2021deepsleep}. Their evaluations span both a large clinical dataset and several public datasets, with evaluation measures such as the arousal index (ArI) error employed to reflect real-world relevance.

This study overcame several limitations of prior research, specifically restricted and non-diverse patient samples and the use of clinically irrelevant evaluation metrics. They addressed the gap in understanding model generalizability across patient groups by leveraging a large, heterogeneous, real-world dataset collected from routine sleep laboratory practice~\citep{ehrlich2024state}.
Findings from robust cross-dataset testing revealed significant performance shifts, underscoring the necessity of multi-center data and rigorous external validation, while the strategy of dataset mixing was noted to be promising for enhancing generalization in the future. 
By selecting the ArI error as their primary evaluation measure, the authors strengthened the methodological alignment of their study with clinical practice.

While their results reaffirm the considerable promise of automated arousal detection, they also emphasize persistent challenges: low inter-rater reliability, limited diversity among patient populations, and pronounced variation in cross-dataset performance. 

The relevance of this prior work to our study is considerable, given that we likewise adopt the \textit{DeepSleep} architecture, determine the accuray of total arousal count estimation (which is similar to them utilizing the ArI error measure), and maintain a focus on clinical effectiveness, including similar observations regarding scorer reliability and inter-standard performance divergence. Nonetheless, our approach extends this foundation by conducting a fully application-grounded user study, an element not addressed by \citet{ehrlich2024state}. Furthermore, our research explicitly examines dimensions of AI transparency, such as post-hoc attribution and the visualization of confidence scores and decision threshold.
Moreover, in contrast to their study, our previous work~\citep{pmlr-v287-kraft25a} made our collected dataset publicly available.
Finally, our evaluation uniquely incorporates direct feedback from professional scorers, thereby enabling a holistic assessment of clinical utility and user acceptance. This allows us to systematically quantify the impact of transparency and the timing of AI assistance on accuracy, bias, efficiency, and user acceptance, factors directly tied to clinical utility.

\paragraph{The Unclear Value of Explainability and the Need for Application-Grounded Evaluations.}
The prevailing argument suggests that enhancing the explainability of AI-driven decision support systems can foster user trust and optimize human-AI collaboration in medical contexts~\citep{yu2025enhancing}. Yet, empirical evidence emphasizes that the benefits of explainability cannot be assumed to be universal or automatic~\citep{rosenbacke2024explainable, gaube2023non}. For example, a recent systematic review by \citet{rosenbacke2024explainable} found that clinicians' trust in AI systems often increases with clear and clinically relevant explanations, particularly when such explanations resonate with medical intuition. Conversely, ambiguous or overly technical explanations may have the opposite effect, leading to diminished trust.

Beyond trust, explainability can also directly influence diagnostic performance and collaborative efficacy. In an AI-assisted radiography study, \citet{prinster2024care} demonstrated that the format of the explanation plays a critical role: radiologists who received local, feature-based explanations improved in both accuracy and receptivity to the AI’s recommendations compared to those provided global, prototype-based explanations. Notably, this influence on reliance and trust occurred largely below participants’ conscious awareness, indicating that explanatory modality subtly shapes clinician behavior. Meanwhile, research by \citet{gaube2023non} showed that clinicians with less experience derived the greatest benefit, both objectively and subjectively, from AI explanations, with experienced experts not receiving any significant benefit from explanations.

On the other hand, explainability can inadvertently encourage over-reliance. \citet{buccinca2021trust} observed that users often deferred to AI suggestions, even erroneous ones, unless the user interface incorporated ``cognitive forcing'' mechanisms to prompt deliberate review. The complexity of evaluating explanations is further heightened by phenomena such as the perception-behavior gap~\citep{gaube2021ai}, where clinicians may report skepticism towards AI-generated advice but nonetheless rely on it in practice, revealing a divergence between stated preferences and actual behaviors.

Collectively, these studies underscore the importance of engaging healthcare professionals in controlled, context-rich experimental designs that mirror real-world workflow. Doing so not only surfaces usability and interaction issues potentially overlooked in technical evaluations, but also more accurately documents how explanations affect clinically relevant outcomes. The literature increasingly advocates for application-grounded evaluation strategies that consider workflow compatibility, cognitive burden, and concrete decision impacts, rather than presuming that additional explanation invariably produces greater benefit~\citep{rosenbacke2024explainable, gaube2021ai}.

In this context, our work aims to address these challenges by systematically investigating how AI transparency, contrasting transparent (white-box) with opaque (black-box) assistance, affect trust, efficiency, and diagnostic accuracy among professional sleep scorers in arousal scoring.

\subsection{Paper outline.}
This paper is organized as follows:
Section~\ref{sec:methodologies} elaborates on the methodologies employed in this study. Section~\ref{sec:user-study-design} describes the experimental design, including participant recruitment, data and model selection, and the study protocol. Section~\ref{sec:results} presents both quantitative and qualitative findings for each research question. Section~\ref{sec:discussion} interprets the results comprehensively, discusses limitations,
and concludes the study.

\section{Methodologies}\label{sec:methodologies}

This section presents the core methodologies applied in this study. Section~\ref{sec:ml-model} offers an overview of the machine learning model architecture, training, and inference strategies for arousal detection. Section~\ref{sec:explanation-methods} introduces the explanation approaches, encompassing both local (Section~\ref{sec:local-explanations}) and global (Section~\ref{sec:global-explanations}) post-hoc feature attribution techniques. Section~\ref{sec:consensus-ground-truth} details the construction of a consensus ground truth from multiple annotators, including annotation clustering (Section~\ref{sec:data-preparation}) and the Expectation-Maximization method for joint estimation of true labels and annotator quality (Section~\ref{sec:em-truth-quality}). Section~\ref{sec:performance-evaluation} describes the performance evaluation framework, addressing both event-based (Section~\ref{sec:per-event-performance}) and count-based (Section~\ref{sec:total-event-count-performance}) metrics. Section~\ref{sec:pairwise-inference-cps-primary} explains the pairwise inference procedures using the CPS ground truth and consensus labels. Finally, Section~\ref{sec:statistical-analysis-framework} outlines the statistical analysis framework, including experimental design, modeling strategies, repeated-measures ANOVA, permutation testing, effect size estimation, and simple effects analysis. Table~\ref{tab:table_of_notation} (Appendix~\ref{appendix:table-of-notation}) provides a table of notation.

\subsection{Machine Learning Model}\label{sec:ml-model}
Detecting arousal events in sleep is challenging due to their brief, subtle nature. Classical machine learning (ML) models often require extensive feature engineering and may not generalize well~\citep{zan2023multi}.
To address these issues, we employ a ML model designed for robust, end-to-end arousal onset detection.
Specifically, we use the top-performing model, referred to as \textit{D4} in our prior research~\citep{pmlr-v287-kraft25a}. 

This section provides a concise overview of the model's architecture, training, and inference processes. For comprehensive technical details, including hyperparameter configurations, readers are directed to our previously published work~\citep{pmlr-v287-kraft25a}.

Our model is a modified version of the DeepSleep architecture, specifically adapted for detecting arousal onsets in sleep data~\citep{li2021deepsleep}. As refined by \citet{fonod2022deepsleep}, it employs a streamlined U-Net architecture, reducing the depth from 11 to 5 layers to decrease computational demands while maintaining performance. 
A weighted binary cross-entropy (BCE) loss function addresses data imbalance, with arousal events defined as 10-second intervals around the onset, in accordance with clinical scoring guidelines~\citep{berry2012rules}. Detailed descriptions of the model architecture and training process are available in earlier publications~\citep{fonod2022deepsleep, pmlr-v287-kraft25a}.

During inference, the model produces confidence scores for each time step, representing the likelihood of an arousal event initiation. To mitigate noise and reduce false positives, the output is smoothed using a 3-second averaging filter, following the methodology outlined in~\citet{pmlr-v287-kraft25a}.

Binary predictions are derived from these confidence scores using the Approximate Localization and Precise Event Count (ALPEC) framework~\citep{pmlr-v287-kraft25a}. ALPEC facilitates post-processing and performance evaluation for arousal detection by ensuring precise event localization and counting. The ALPEC variant applied here determines the decision threshold based on the F2-score from the training set, prioritizing the minimization of missed events (false negatives). This focus is crucial in clinical contexts, where recalling arousal events is more critical than precision to prevent missing significant occurrences. The F2-score is highlighted as the primary metric for performance evaluation, as it aligns with the operational objectives of clinical decision support systems, where minimizing false negatives is essential, allowing clinicians to address false positives, thereby optimizing resource use while maintaining high efficacy~\citep{deo2015machine}. Following thresholding, predicted intervals within 10 seconds of each other are merged, intervals exceeding 60 seconds are discarded, and a 15-second temporal buffer is applied on both sides of the ground-truth annotations to ensure clinically relevant alignment.

For an in-depth understanding of the ALPEC framework's rationale and its alignment with clinical standards, refer to~\citet{pmlr-v287-kraft25a}.

\subsection{Explanation Methods}\label{sec:explanation-methods}
The DeepLift method~\citep{shrikumar2017learning} is employed to generate both local and global post-hoc feature attributions. It is a gradient-based interpretability technique recognized for its effectiveness in neural time series classifiers~\citep{vsimic2025comprehensive}. Also, it provides notable advantages in computational efficiency and memory usage over many alternative methods.
DeepLift operates by comparing the activation of a neuron to a reference activation, with the resulting difference used to assess the significance of input features in the model's decision-making process. The choice of the reference activation will be discussed in the context of the study design in Section~\ref{sec:local-explanations-design}.
The following subsections briefly describe the methodologies applied in this study. The underlying rationale and illustrative examples are presented in Sections~\ref{sec:local-explanations-design} and~\ref{sec:global-explanation-design}, respectively.

\subsubsection{Local Explanations}\label{sec:local-explanations}
To provide local explanations for individual arousal events while minimizing computational demands and complexity, we focus on calculating attributions solely for the time point identified by the model as the most probable event onset. We present visualizations of the attributions for the top 10 channels deemed most significant to the model's decision-making process. Channel importance is determined by aggregating attributions across all time points.
We display attributions for a 60-second window centered symmetrically around the predicted event onset. Furthermore, we exclusively present positive attributions that exceed a predefined threshold, thereby simplifying the explanation.

\subsubsection{Global Explanation}\label{sec:global-explanations}
In this study, a global explanation of the model's decision-making process is derived by aggregating feature attributions across all events from subjects included in the test set that are selected in the user study. The attributions are summed over all time points and all events for each channel. Subsequently, channels are ranked based on their importance for the model's decision. The 20 channels with the highest importance are depicted in a bar chart for visualization.

\subsection{Consensus Ground Truth Derivation}\label{sec:consensus-ground-truth}
To adequately assess the performance of both human annotators and the AI model in identifying arousal events, we want to establish a consensus ground truth that encapsulates the collective expertise of multiple annotators. Given the temporal nature of arousal events and the potential misalignment of annotations from different experts, we adopt a two-step approach: (1) clustering annotations based on their temporal proximity, and (2) employing an Expectation-Maximization (EM) algorithm to estimate both the consensus ground truth and the performance of individual annotators. Both steps are described in the following subsections.

\subsubsection{Clustering of Annotations}\label{sec:data-preparation}
Each annotator provides a series of arousal event start times, denoted as $\{ t_{k,i} \}$, where $k$ represents the annotator and $i$ signifies their respective events. To group annotations that likely correspond to the same underlying event, we employ temporal clustering.

Given the irregular temporal distribution of arousal events, fixed temporal windows are unsuitable, and overlapping windows would unnecessarily complicate the analysis. Therefore, we organize the annotations into clusters $\{ C_j \}$ of similar events, with each cluster $C_j$ comprising annotations from one or more annotators that are temporally close to each other.

To mitigate potential bias from selecting a single clustering algorithm, we utilize two distinct methods: DBSCAN~\citep{ester1996density}, a density-based clustering algorithm, and agglomerative clustering~\citep{sokal1958statistical}, a hierarchical clustering algorithm. The event start times $t_{k,i}$ constitute the one-dimensional feature space for clustering.

\begin{description}[leftmargin=2em,style=nextline]
\item[DBSCAN] The DBSCAN algorithm effectively identifies clusters of densely packed annotations while designating isolated points as noise. The optimal value for the $\varepsilon$ parameter is determined through the k-distance method in conjunction with the Kneedle algorithm~\citep{satopaa2011finding}.
To ensure robust parameterization, we evaluate multiple sensitivity values $S$ for the Kneedle algorithm, which modulate the algorithm's responsiveness to knee point detection. Lower sensitivity values make the algorithm less responsive, typically identifying knee points later in the curve (resulting in larger $\varepsilon$), whereas higher sensitivity values detect knee points earlier (yielding smaller $\varepsilon$).
Subsequent post-processing ensures that each cluster includes no more than one annotation per annotator and that all clusters meet a minimum annotator count.

\item[Agglomerative Clustering] Agglomerative clustering is employed to iteratively merge annotations based on their temporal proximity. This process utilizes Ward's linkage~\citep{ward1963hierarchical} as the distance metric. Post-processing steps are implemented to eliminate duplicate annotations from the same annotator within a cluster and to exclude clusters that lack a sufficient number of unique annotators.
\end{description}

\subsubsection{Expectation-Maximization for Simultaneous Truth and Annotator-Quality Estimation}\label{sec:em-truth-quality}
Following the clustering phase, we utilize the Expectation-Maximization (EM) technique to simultaneously estimate the true occurrences of events and evaluate annotator performance. This process involves adapting the well-established Dawid-Skene EM algorithm~\citep{dawid1979maximum} to our specific context of clustered, time-stamped event annotations. Our aim is to determine the probability that each cluster represents a true event while also assessing the sensitivity and specificity of each annotator. This is accomplished through a two-class latent-variable model, wherein the latent variable signifies the presence of an event within a cluster. Each cluster is modeled to potentially contain either zero or one latent true event.

\begin{description}[leftmargin=2em,style=nextline]
\item[Initialization]
Sensitivities ($s_k$) and specificities ($p_k$) for each annotator $k$ are initialized to 0.99, reflecting an initial assumption of high performance while avoiding computational issues associated with starting at 1.0.
The initial probability that a cluster $C_j$ represents a true event is set to $P_j = 0.5$.

\item[EM Iterations]

\textbf{E-Step (Expectation)}: For each cluster $C_j$, we compute the probability $P_j$ that it represents a true event, given the current estimates of annotator sensitivities $s_k$ and specificities $p_k$. For each annotator, we define $A_{k,j} = 1$ if they contributed to cluster $C_j$, and $0$ otherwise.
The log-likelihood of the observed annotations in $C_j$ if an event is truly present is given by
\begin{equation}
\log(L_j) = \sum_{k=1}^{K} \left[ A_{k,j} \log(s_k) + (1 - A_{k,j}) \log(1 - s_k) \right].
\end{equation}

Conversely, the log-likelihood if no event is present is given by
\begin{equation}
\log(M_j) = \sum_{k=1}^{K} \left[ A_{k,j} \log(1 - p_k) + (1 - A_{k,j}) \log(p_k) \right].
\end{equation}

The posterior probability for an event in cluster $C_j$ is then

\begin{equation}
P_j = \frac{\exp(\log(D_{1j}))}{\exp(\log(D_{1j})) + \exp(\log(D_{2j}))},
\end{equation}

where the log-posterior probabilities for an event being present and absent are given by $\log(D_{1j}) = \log(L_j) + \log(P_j^{\text{prev}})$ and $\log(D_{2j}) = \log(M_j) + \log(1 - P_j^{\text{prev}})$.

Here, $P_j^{\text{prev}}$ and $1 - P_j^{\text{prev}}$ are the prior probabilities of an event being present / absent in cluster $C_j$ from the previous iteration.

\textbf{M-Step (Maximization)}: We update the sensitivities $s_{k}$ and specificities $p_{k}$ of the annotators based on the current estimates of cluster probabilities as
\begin{equation}
s_k = \frac{\sum_{j} P_j A_{k,j}}{\sum_{j} P_j},
\end{equation}
and
\begin{equation}
p_k = \frac{\sum_{j} (1 - P_j)(1 - A_{k,j})}{\sum_{j} (1 - P_j)},
\end{equation}

These updates reflect the expected true positive and true negative rates, weighted by the current cluster probabilities.

\textbf{Convergence Criteria}: 
The algorithm iterates until the maximum change in any cluster probability falls below a set tolerance (e.g., $1 \times 10^{-7}$), or a maximum number of iterations is reached.

\textbf{Probability Bounds}: 
After each update, all probabilities and performance parameters are clipped to the $[0, 1]$ interval to ensure validity.

\textbf{Consensus Event Selection}:
After convergence, clusters with $P_j \geq \tau$ are considered consensus events, where typically $\tau = 0.5$
The consensus event time for each cluster is computed as the mean of the annotated times within the cluster.

\textbf{Implementation Notes}:
In our implementation, all calculations are performed in log-space to ensure numerical stability, effectively mitigating potential underflow and overflow issues associated with extremely small or large numbers.

Furthermore, the noise cluster is deliberately excluded from all Expectation-Maximization (EM) updates to preserve the integrity of the results.

The method is implemented in Python, utilizing the robust functionalities of libraries such as NumPy and scikit-learn.

To further enhance the robustness of our calculations, we introduce small constants, such as $1 \times 10^{-10}$, during logarithmic computations or divisions. This precautionary measure prevents undefined operations, such as the logarithm of zero or division by zero, thereby ensuring the reliability of the computational process.

\item[Critical Considerations]
The accuracy of the consensus is fundamentally dependent on the efficacy of the clustering process. Poorly constructed clusters can substantially compromise the precision of the consensus outcomes.

Moreover, the algorithm presupposes the independence of annotators, a common assumption in analogous methodologies. Nonetheless, this assumption may not consistently reflect real-world conditions, thereby potentially affecting the reliability of the results.

\end{description}

\subsection{Performance Evaluation}\label{sec:performance-evaluation}
Performance is assessed in two primary ways: the evaluation of individual event scoring, detailed in Section~\ref{sec:per-event-performance}, and the evaluation based on the total event count, elaborated in Section~\ref{sec:total-event-count-performance}.

\subsubsection{Event-Based Performance Evaluation}\label{sec:per-event-performance}
The evaluation of an expert's performance, which may be a human annotator, the AI model, or a human-AI collaborative team, is conducted by assessing the alignment between expert annotations and a specified ground-truth.

For human experts, the onset point of each annotated event serves as the reference point for comparison. In the case of the AI model, the argmax values of the model's confidence score output for each event are utilized as reference points. For human-AI teams, reference points are determined by either the onset points of events marked by humans or the argmax value of an AI event that has been accepted by the human annotator.

We proceed depending upon the type of ground-truth employed:

\begin{description}
\item[Consensus Ground-Truth]
The center point of a cluster is calculated as the mean value of the annotations constituting the cluster, which is then used as the reference point for comparison.

For the evaluation of human solo performance, calculations could be based on the existing clustering structure. However, for novel annotations not previously included in the clustering procedure, a method must be established to associate them with either a consensus cluster or a noise cluster. A pragmatic approach involves utilizing the distance of an annotation to the centroids of consensus clusters to determine the nearest cluster. 
This is done by defining a distance threshold, derived empirically from clustering statistics, which determines whether an annotation should be 
assigned to a consensus cluster or 
classified as noise.
For consistency, human solo performance will also be evaluated using this methodology.

\item[CPS Ground Truth]
An externally determined standard, in the form of ground-truth event annotations, may also be employed for comparison. In this context, the ground truth from the CPS dataset, that was used to train and evaluate the AI model, is utilized. The onset of an event annotation serves as the reference point for comparison to expert annotations.

\end{description}

The evaluation involves two binary sequences: one for expert predictions and another for ground-truth annotations, where events are denoted by a single 1 at the reference point indicating an event start, and 0 elsewhere.

Matches between expert and ground-truth annotations are determined as follows: a predicted 1 is counted as a true positive (TP) if it occurs within a specified time window, defined by the distance threshold on either side of a ground-truth 1. Predicted 1s that fall outside this window are counted as false positives (FP). Conversely, any ground-truth 1 that does not have a corresponding predicted 1 within the threshold window is counted as a false negative (FN).

Unlike the AI model, where the primary optimization and performance evaluation metric was the F2 score (refer to Section~\ref{sec:ml-model}), the F1 score, which represents the harmonic mean of precision and recall, is now employed for a more balanced evaluation. Additionally, the F2 score, precision, recall, as well as TP, FP, and FN counts are reported for a comprehensive performance analysis.

\paragraph{Benefit Ratio.}
The \textit{benefit ratio} is a normalized alignment score utilized to evaluate the performance of a human-AI team in comparison to a fixed ground truth. The benefit ratio is applicable in scenarios where the AI outperforms the unaided human, that is, when $F1^{\text{AI}} > F1^{\text{HU}}$. The benefit ratio is calculated by normalizing the team's performance relative to the performance of an unaided human and the AI's performance.

The formula for the benefit ratio $\mathcal{B}$ is defined as:
\begin{equation}
\mathcal{B} = \frac{F1^{\text{HU+AI}} - F1^{\text{HU}}}{F1^{\text{AI}} - F1^{\text{HU}}},
\end{equation}
where $F1^{\text{HU+AI}}$ is the F1 score of the human-AI team, $F1^{\text{HU}}$ is the F1 score of the unaided human (solo performance). 
A value of $\mathcal{B} > 1$ would indicate that the human-AI team has performed better than the AI alone, while a value of $\mathcal{B} < 0$ would indicate that the human-AI team has performed worse than the human alone.
In the case, where $F1^{\text{HU}} < F1^{\text{HU+AI}} < F1^{\text{AI}}$, the benefit ratio is bounded between 0 and 1. In this case, an elevated benefit ratio signifies that the human-AI team has effectively harnessed the AI's capabilities to enhance performance beyond the level attainable by the human alone, in comparison to the AI's maximum potential contribution.

In scenarios where the AI model is trained on the CPS ground truth, the benefit ratio provides insight into the extent to which the human participant has adopted the AI's alignment towards this standard.

\paragraph{Relative F1 Scores.}
In cases where we don't have access to the human solo performance, we can still determine the alignment of the human-AI team to the AI's performance.
Comparing the raw F1 scores of a human-AI team across different experimental human-AI interaction regimes is methodologically flawed when the baseline performance of the AI varies between these regimes. For example, if the AI demonstrates sub-optimal performance in one condition, the human-AI team may appear less effective in absolute terms. This apparent weakness is not necessarily due to inadequate human performance but rather because the AI's enhanced performance reduces the potential for further improvement.

To address this issue, we use the relative F1 scores $\mathcal{R}$, defined as the ratio of the F1 score of the human-AI team to that of the AI alone.

\begin{equation}
\mathcal{R} = \frac{F1_{\text{HU+AI}}}{F1_{\text{AI}}}
\end{equation}

This normalization quantifies the human-AI team's capacity to leverage the AI's potential relative to its stand-alone performance, thereby controlling for regime-specific differences in baseline difficulty or suitability. 
This interpretation of alignment is most appropriate in scenarios where the AI outperforms the human-AI team, that is, when $F1^{\text{AI}} > F1^{\text{HU+AI}}$.
Nevertheless, it should be interpreted with caution: a high alignment score neither guarantees greater absolute accuracy nor implies proximity to a clinical ground truth. Instead, it indicates that human corrections remain close to the AI's initial proposal, which may reflect trust, compliance, or alignment, but not necessarily improvement. In this sense, \(\mathcal{R}\) is best understood as a proxy for \textit{calibrated cooperation} rather than performance per se. Its validity rests on the critical assumption that the AI baseline approximates a well-curated, consensus-driven ground truth.

\subsubsection{Count-Based Performance Measures}\label{sec:total-event-count-performance}

This section introduces count-based measures per recording

\paragraph{Notation.}
For each recording let $C_{\mathrm{GT}}$ be the ground-truth arousal count.
For each source $x \in \{\mathrm{AI},\, \mathrm{HU{+}AI},\, \mathrm{HU}\}$, let $C_x$ denote its corresponding arousal count. Here, $\mathrm{AI}$ refers to the AI-only output, $\mathrm{HU{+}AI}$ to the human-AI team output, and $\mathrm{HU}$ to the human-solo output, which is available exclusively in QC regimes.
The primitive absolute count deviation used throughout is defined as
\begin{align}
D_{x\to\mathrm{GT}}&=\bigl|C_x-C_{\mathrm{GT}}\bigr|.
\label{eq:primitive_distances}
\end{align}

\paragraph{Count Accuracy (Bounded APE).}
As a bounded, interpretable score we use the inverse absolute percentage error (APE) transform
\begin{align}
A_{x,\mathrm{GT}}
=\frac{1}{1+\mathrm{APE}_{x,\mathrm{GT}}},
\qquad
\mathrm{APE}_{x,\mathrm{GT}}
=\frac{D_{x\to\mathrm{GT}}}{\max\{C_{\mathrm{GT}},\epsilon\}},
\quad \epsilon>0.
\label{eq:gt-accuracy}
\end{align}
Here, we set $\epsilon=10^{-8}$ to ensure numerical stability. The accuracy measure $A_{x,\mathrm{GT}}$ ranges within $(0,1]$, attaining a value of $1$ under perfect agreement and decreasing monotonically as the absolute percentage error increases.

\paragraph{AI-Baseline Improvement Ratio.}
Our primary improvement measure compares how much the team closes the GT gap relative to the AI on a ratio scale:
\begin{align}
R_{\mathrm{GT}}
=\frac{D_{\mathrm{AI}\to\mathrm{GT}}+\delta}{D_{\mathrm{HU+AI}\to\mathrm{GT}}+\delta},
\qquad \delta>0.
\label{eq:rgt}
\end{align}
We use $\delta=10^{-6}$ to ensure numerical stability and analyze its log form
\begin{align}
y_{\mathrm{RGT}}=\log R_{\mathrm{GT}}
=\log\!\bigl(D_{\mathrm{AI}\to\mathrm{GT}}+\delta\bigr)
 -\log\!\bigl(D_{\mathrm{HU+AI}\to\mathrm{GT}}+\delta\bigr),
\label{eq:yrgt}
\end{align}
so that $y_{\mathrm{RGT}}>0$ ($R_{\mathrm{GT}}>1$) indicates the team is closer
to ground truth than the AI by a multiplicative factor $R_{\mathrm{GT}}$, 
$y_{\mathrm{RGT}}=0$ parity, and $y_{\mathrm{RGT}}<0$ a worse team outcome.
For reporting we back-transform point estimates and confidence intervals via $\exp(y_{\mathrm{RGT}})$.

\paragraph{Percentage Error.}
To assess systematic count bias toward over- or under-counting, we report the team's percentage error:
\begin{align}
    \mathrm{PE}=\frac{C_{\mathrm{HU+AI}}-C_{\mathrm{GT}}}{\max\{C_{\mathrm{GT}},\epsilon\}},
\label{eq:bias}
\end{align}
We set $\epsilon=10^{-8}$ to ensure numerical stability. A value of $\mathrm{PE}>0$ indicates over-counting, while $\mathrm{PE}<0$ reflects under-counting relative to the ground truth. In contrast to $R_{\mathrm{GT}}$, which quantifies only the \emph{magnitude} of improvement, the percentage error characterizes both the \emph{magnitude} and \emph{direction} of the residual count error.

\paragraph{QC team--solo comparison (applicable only in QC regimes).}
In QC regimes, where a human-solo baseline is available, we directly compare $A_{\mathrm{HU+AI},\mathrm{GT}}$ and $A_{\mathrm{HU},\mathrm{GT}}$ for the same patient recordings. The paired difference, defined as $\Delta A_{\mathrm{GT}} = A_{\mathrm{HU+AI},\mathrm{GT}} - A_{\mathrm{HU},\mathrm{GT}}$, is computed to quantify the extent of improvement achieved by the team relative to the solo human performance.

\paragraph{Hypotheses used in inference.}
All tests are performed at the participant level and then aggregated per atomic regime (Start/QC $\times$ WB/BB): (i) for $y_{\mathrm{RGT}}$ we use the directional hypothesis $H_1:\,\mu_{y_{\mathrm{RGT}}}>0$ (team improves over AI). (ii) For bias we use the two-sided hypothesis $H_1:\,\mu_{B}\neq 0$.

\subsection{Pairwise Inference under the CPS Ground Truth (Primary) and Consensus (Sensitivity)}
\label{sec:pairwise-inference-cps-primary}

\paragraph{Scope.}
All confirmatory, pairwise tests use the \emph{CPS ground truth} as the authoritative annotation source. Analyses based on the human \emph{consensus} labels are treated as sensitivity checks and are \emph{not} part of any multiplicity-controlled family.
Because the CPS ground truth is prespecified as the sole confirmatory annotation source, each of the three questions of the second primary objective \textit{PO2} (performance comparison of human-AI team vs.\ human solo, human-AI team vs.\ AI, and human solo vs.\ AI, see Section~\ref{sec:objectives}) forms a family of size one, so there is no multiplicity adjustment.

\paragraph{Outcome scales.}
We conduct the same inferential procedure on two evaluation scales:
(i) \emph{event-level} performance, where the per-participant outcome is the F1-score, and
(ii) \emph{count-level} performance, where the outcome is count accuracy.
Let $M(\cdot)$ denote a generic per-participant performance metric (either F1-score or accuracy).

\subsubsection{Human-AI Team vs.\ Human Solo (paired)}
\label{sec:team-vs-solo-primary}
For each participant $i=1,\dots,n$ we form the paired difference
\begin{align}
d_i \;=\; M_{\text{HU+AI},i}^{\text{CPS}} - M_{\text{HU},i}^{\text{CPS}}.
\end{align}
We test $H_0\!:\,\mu_d=0$ (where $\mu_d$ denotes the population mean difference) using (a) the paired $t$-test and (b) an exact, two-sided sign-flip permutation test on the \emph{mean} difference $\bar d$.
With $n \in \{8, 9\}$ across both standards, we enumerate all $2^n$ sign configurations exactly.
We report $\bar d$ with a nonparametric percentile bootstrap \mbox{95\% CI} based on 10{,}000 resamples of $\{d_i\}_{i=1}^n$.
The identical workflow is applied to the consensus labels for sensitivity only.

\subsubsection{Human-AI Team and Human Solo vs.\ AI (one-sample vs.\ scalar)}
\label{sec:team-solo-vs-ai-primary}
Let $a_{\text{AI}}^{\text{CPS}}$ be the AI's performance under CPS ground truth (a single scalar).
For each participant $i$ we define two comparisons:
\begin{align}
d_i^{\text{HU+AI}} \;=\; M_{\text{HU+AI},i}^{\text{CPS}} - a_{\text{AI}}^{\text{CPS}}, \\
d_i^{\text{HU}} \;=\; M_{\text{HU},i}^{\text{CPS}} - a_{\text{AI}}^{\text{CPS}}.
\end{align}
For each comparison, we test $H_0\!:\,\mu_d=0$ (where $\mu_d$ denotes the population mean difference) using a one-sample $t$-test and an exact two-sided sign-flip permutation test on $\bar d$, and report $\bar d$ CIs and $d_z$ with bootstrap CIs.
Consensus-based sensitivity analyses contrast $M_{\text{HU+AI},i}^{\text{cons}}$ and $M_{\text{HU},i}^{\text{cons}}$ to $a_{\text{AI}}^{\text{cons}}$ in the same manner.

\subsection{Statistical Inference for Team Performance over all Regimes under the CPS Ground Truth}
\label{sec:statistical-analysis-framework}

\subsubsection{Experimental Design and Scope of Inference}\label{sec:experimental-design-and-scope-of-inference}
All inferential analyses employ a $2{\times}2$ within-participant factorial design, with the factors \textit{Timing} (levels: Start and QC) and \textit{AI transparency} (levels: WB and BB; hereafter referred to as \textit{AI}). The study includes eight participants who provide complete data across all four experimental conditions. Inferential results are confined to these four atomic regimes (Start/QC $\times$ WB/BB). Composite regimes, which combine data from multiple phases or conditions (indicated with a ``+'' symbol, e.g. ``Start,BB+WB''), are used solely for descriptive purposes and are excluded from hypothesis testing to preserve statistical independence and maintain a clearly defined family of tests.

\subsubsection{Statistical Modeling and Analysis Scale}\label{sec:statistical-modeling-and-analysis-scale}
All statistical models are fitted on the logarithmic scale to stabilize variance and facilitate additive modeling of multiplicative effects. Outcome variables are log-transformed with a small offset ($\varepsilon=10^{-8}$) to address zero values. All point estimates and 95\% confidence intervals are reported as back-transformed ratios using $\exp(\cdot)$. Sphericity is not a concern, as each within-subject term in the $2{\times}2$ design has one degree of freedom.

\subsubsection{Repeated-Measures ANOVA and Planned Contrasts}\label{sec:repeated-measures-anova-and-planned-contrasts}
Repeated-measures ANOVA (RM-ANOVA) is conducted, 
with subject as the repeated-measures factor and \textit{Timing} and \textit{AI} as within-participant factors. To test the study hypotheses, standard orthogonal contrasts are defined for the $2{\times}2$ design and applied to participant-wise cell means: (i) the main effect of \textit{AI} (WB vs BB), (ii) the main effect of \textit{Timing} (Start vs QC), and (iii) the interaction effect.

\subsubsection{Permutation Testing and Multiple Comparisons}\label{sec:permutation-testing-and-multiple-comparisons}
For each contrast, participant-wise contrast scores are computed and their means are tested using within-subject contrast t-tests. Exact sign-flip permutation $p$-values are provided, with Holm correction applied within contrast families. Specifically, Holm correction is used for the two main effects, while the interaction effect is evaluated using unadjusted permutation $p$-values. This approach ensures robustness in the presence of small sample sizes and effectively controls the familywise error rate.

\subsubsection{Effect Sizes and Confidence Intervals}\label{sec:effect-sizes-and-confidence-intervals}
Effect ratios (back-transformed contrast means) and the within-subject effect size $d_z$ are reported. Confidence intervals for both measures are obtained via nonparametric bootstrap (10{,}000 resamples of participant-level contrast scores, using the percentile method). Effect sizes are presented as both partial eta-squared ($\eta^2_{\mathrm{p}}$) from ANOVA and Cohen's $d_z$ from contrasts: $\eta^2_{\mathrm{p}}$ quantifies the proportion of variance explained by the factor, while $d_z$ represents the standardized mean difference for the contrast.

\subsubsection{Simple Effects Analysis}\label{sec:simple-effects-analysis}
When significant interactions are identified, simple effects are examined using paired $t$-tests with sign-flip permutation tests and Holm correction within the family of simple effects. This procedure clarifies the nature of the interaction by comparing conditions within each level of the other factor.

\section{User Study Design}\label{sec:user-study-design}

This section provides a comprehensive overview of the user study design. We begin by outlining the general design considerations that guided the development of the study in Section~\ref{sec:user-study-design-aspects}. Next, we describe the overall structure and objectives of the arousal scoring task in Section~\ref{sec:task-overview}. The design and implementation of the Decision Support System (DSS) are detailed in Section~\ref{sec:decision-support-system}, followed by an explanation of the different modes of explainability available to participants in Section~\ref{sec:modes-of-explainability}. 
Section~\ref{sec:dataset-and-samples} details the dataset and sample selection process, while Section~\ref{sec:study-protocol} provides an in-depth description of the study protocol including objectives, recruitment and participants and the conrete procedure of the study. The overarching evaluation approach and its underlying rationale are then outlined in Section~\ref{sec:evaluation-approach-rationale}.

\subsection{General Design Considerations}\label{sec:user-study-design-aspects}

As highlighted in Section~\ref{sec:introduction}, application-grounded user studies are vital for assessing the practical utility of AI explanations. Consequently, \citet{amarasinghe2023explainable} emphasize that evaluations should be conducted within the context of the specific task the AI system is intended to support.
To achieve this, we utilized contemporary patient data sourced from a sleep clinic, enlisted the expertise of professional sleep scorers, developed a web-based application enabling these scorers to execute the arousal scoring task, and employed evaluation measures pertinent to the clinical environment.

\subsection{Task Overview}\label{sec:task-overview}
Participants in our study are tasked with scoring arousal events using polysomnography data. 

This task is an essential element of sleep diagnostics, based on the American Academy of Sleep Medicine guidelines~\citep{berry2012rules}. Professional sleep scorers, who are trained in visually analyzing polysomnography (PSG) data, execute this task by identifying the onset of arousal events. In clinical settings, typically only the onset of an arousal event is marked, as its duration is generally deemed clinically irrelevant. For simplicity, we will refer to these as arousal events, although the emphasis is on identifying the onset.

The AI-powered Decision Support System (DSS) identifies regions where the onset of an arousal event is most probable. If a scorer agrees that an arousal begins within the suggested region, they may accept the event without further modification. Alternatively, they may create an event independently. If the AI-suggested region is ambiguous and the scorer wishes to accept the AI event, they must adjust the region to ensure it encompasses the correct onset of the arousal. In the absence of an AI-determined region near the scorer's perceived arousal onset, they must manually create a new arousal event.

Upon completion of the scoring process, all manually created arousal events, along with accepted AI-suggested events, are included in the final scoring. There is no obligation to explicitly reject AI-suggested events that the scorer disagrees with.

Furthermore, for each identified arousal event, whether manually created or accepted from AI suggestions, participants are encouraged to provide a confidence assessment. This assessment is a subjective evaluation of the scorer's confidence in the accuracy of their identification and can be selected from the following five levels: ``very uncertain'', ``uncertain'', ``confident'', ``very confident'', and ``fully confident'', with ``confident'' as the default. The definitions of these levels, which are covered during the initial training session of the user study, are as follows:

\begin{description}
    \item[Very uncertain] signifies a slight inclination towards believing the event is an arousal but with considerable doubt.
    \item[Confident] denotes a standard level of confidence in the event being an arousal.
    \item[Fully confident] indicates the scorer has absolute certainty that the event is an arousal, with an expectation that this assessment is highly unlikely to be incorrect.
    \item[Uncertain and Very confident] serve as intermediate levels, offering gradations between the other confidence levels.
\end{description}

In the context of this user study, the arousal scoring task is simplified by not requiring causal differentiation between various arousal event types, and arousal scoring is concluded at 1 AM for each patient. According to the guidelines of the American Academy of Sleep Medicine, all arousals, regardless of their cause, must be visible in the electroencephalogram channel. Therefore, identifying a generic arousal class remains a realistic task routinely performed by professional sleep scorers.

\subsection{Decision Support System}\label{sec:decision-support-system}
\begin{figure}[!htbp]
\centering
\includegraphics[width=\textwidth]{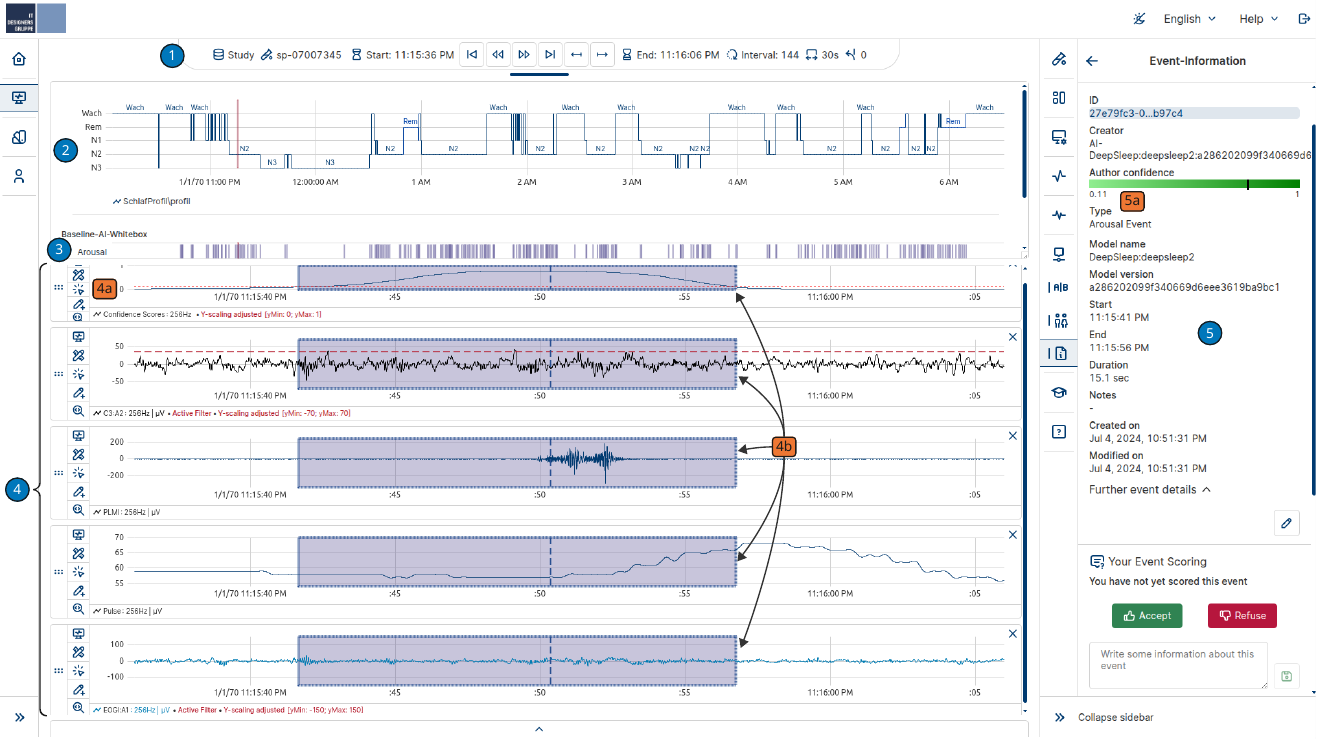}
\caption{\textbf{Main interface of the Decision Support System} for the arousal scoring task. The interface is organized into several key areas, each marked with blue circles: 
\textbf{1)} Control bar providing access to basic information, and navigation tools; 
\textbf{2)} Overview graph displaying the hypnogram (sleep stages) for the entire recording, with the currently selected interval highlighted by an orange vertical bar; 
\textbf{3)} Timeline summarizing AI-annotated arousal event positions across the recording; 
\textbf{4)} Visualization of selected polysomnography data channels for the current interval, including overlays for AI-suggested arousal regions; 
\textbf{5)} Event information panel presenting details of the currently selected arousal event, including options to accept or reject the event;
Additional transparency elements, highlighted with orange ovals, include: 
\textbf{4a)} Confidence score channel, visualizing the AI model's confidence for arousal onset at each time point; 
\textbf{4b)} Shaded regions on each channel, indicating intervals where an arousal onset is likely, with the most probable onset marked by a vertical dashed line (corresponding to the maximum confidence score);
\textbf{5a)} Bar indicator showing the maximum confidence score for the selected event, visualized with a green gradient.}
\label{fig:dss-main-interface}
\end{figure}

The Decision Support System (DSS) is a web-based platform that facilitates the arousal scoring task through an intuitive user interface. It is designed to enable users to conduct the arousal scoring task with or without AI assistance. Figure~\ref{fig:dss-main-interface} shows the main interface of the DSS.

The development of the DSS was conducted in collaboration with a seasoned sleep medicine expert, who also possesses extensive experience as a trainer of medical scorers.

The design of the Decision Support System (DSS) was guided by several key principles to ensure both functionality and usability. Central to the system is the integration of an efficient machine learning (ML) model for the detection of arousals during sleep, as described in Section~\ref{sec:ml-model}. This model was trained on patient data from the Comprehensive Polysomnography Dataset~\citep{kraft2024cps}, ensuring clinical relevance and robustness. In addition to its predictive capabilities, the DSS is equipped to provide comprehensive explanations for the ML model's predictions, a feature that is further elaborated in Section~\ref{sec:modes-of-explainability}. 

A primary objective in the system's development was to achieve efficient time behavior, enabling rapid responses to user inputs and timely data processing. The platform was designed to be intuitive and easy to learn, with an attractive user interface that minimizes the learning curve and enhances the overall user experience. To further support user efficiency, the interface was engineered to minimize the number of required interactions, presenting all relevant information within a single view and offering keyboard shortcuts for common actions as well as multiple navigation options, such as quickly moving between intervals or events. 

Customization was another important consideration, allowing users to tailor the interface to their preferences by adjusting channel configurations, color schemes, and other settings. These configurations can be saved as profiles and reloaded as needed, supporting both flexibility and consistency in user workflows. The DSS also incorporates specific features such as auxiliary lines within channels and filtering options for time series data, including high-pass and low-pass filters. Finally, the system provides comprehensive event management capabilities, enabling users to either manually annotate events or accept or refuse AI-suggested events, thereby supporting a seamless integration of human expertise and AI assistance.

\subsection{Modes of explainability}\label{sec:modes-of-explainability}

The DSS incorporates multiple complementary modes of explainability to support AI-assisted arousal scoring, each designed to enhance user understanding and trust in the system’s predictions. 

\begin{enumerate}
    \item \textbf{Local explanations} are provided via post-hoc feature attributions for the most probable onset of arousal events, enabling users to interpret the model’s decision at the individual event level. For a detailed description and illustration, see Section~\ref{sec:local-explanations-design} and Figure~\ref{fig:local-explanations}.
    \item A \textbf{global explanation} is presented as a bar chart summarizing the overall importance of each data channel in the model’s decision-making process, offering users insight into which physiological signals most strongly influence predictions across the dataset. Further details are provided in Section~\ref{sec:global-explanation-design} and Figure~\ref{fig:global-explanation}.
    \item The DSS enables users to select a data channel that visualizes the AI model’s \textbf{confidence scores} for each time point, highlighting regions with a high likelihood of arousal onset. This functionality is depicted in Figure~\ref{fig:dss-main-interface}, area~4a).
    \item The \textbf{decision threshold} used by the model is explicitly visualized within the confidence score channel, allowing users to see how predicted scores relate to the threshold for arousal detection (see Figure~\ref{fig:dss-main-interface}, area~4a)).
    \item The \textbf{most probable onset} of an arousal event is clearly marked within the predicted interval, providing a precise temporal reference for users (see Figure~\ref{fig:dss-main-interface}, area~4b)).
    \item The DSS also offers a concise visualization of the \textbf{confidence value} associated with the most likely onset of a predicted arousal event, supporting rapid assessment of prediction certainty (see Figure~\ref{fig:dss-main-interface}, area~5a)).
\end{enumerate}

\subsubsection{Local Explanations}\label{sec:local-explanations-design}

\paragraph{Stage of Explanations: Post-hoc.}
We have chosen to employ post-hoc explanations that offer feature attributions for predictions. This targeted approach aligns well with the application-grounded nature of our user study, as investigating multiple forms of explanations or methods for generating post-hoc feature attributions could detract from the authenticity of a real-world setting. Furthermore, providing a singular type of explanation simplifies the process for participants, enabling us to concentrate on other critical elements, such as the timing of AI assistance. Post-hoc explanations are particularly pertinent to our study, given their prevalent use in the healthcare sector~\citep{gupta2024comparative}, thereby enhancing the relevance and applicability of our findings.

\paragraph{Method of Explanations: DeepLIFT.}
In this study, we utilize DeepLIFT for model explainability (see Section~\ref{sec:explanation-methods}). A critical aspect of DeepLIFT is the choice of the reference activation. In other modalities of data processing, missing features are typically represented as black pixels in images or as zero/mean values in tabular datasets. However, in the context of time series data, defining a missing feature is more complex~\citep{rezaei2024explanation}. To address this challenge, we employ a random Gaussian noise baseline with a mean of zero and a standard deviation of one. While this approach may introduce artifacts in the explanations for certain channels due to the different characteristics of the large number of channels in the PSG data, it remains consistent with the objectives of our study. Our aim is not to identify new biomarkers for arousal detection. Rather, if any explanations are unfaithful, it provides an opportunity to assess the participants' ability to recognize such inconsistencies. Still, a comparison between DeepLIFT and GradientSHAP is provided in Appendix~\ref{appendix:comparison-of-feature-attribution-methods}.

\paragraph{Balancing Completeness and Interpretability.}
Following the reasoning of~\citet{gilpin2018explaining}, we recognize the necessity of balancing completeness and interpretability in explanations. While overly complex explanations can impede interpretability, offering overly simplified explanations without transparency is considered unethical. To address this, we implement three threshold levels for feature attributions. A lower threshold results in more complex yet comprehensive explanations. This information is communicated transparently to the participants.

\paragraph{Illustration of Local Explanations.}

\begin{figure}[!htbp]
\centering
\includegraphics[width=\textwidth]{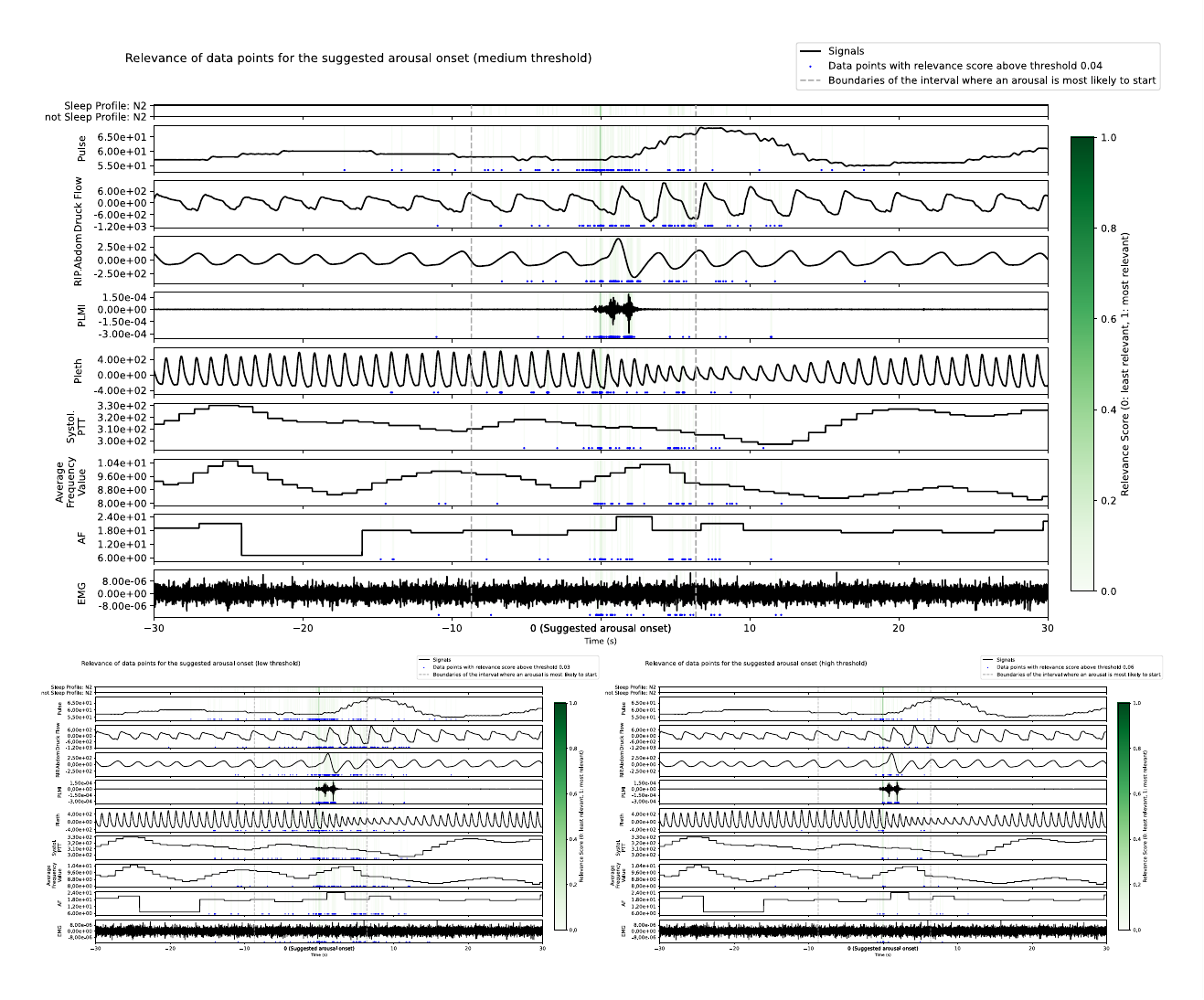}
\caption{\textbf{Local explanations of the AI model's arousal prediction at three levels of detail} for the arousal depicted in Figure~\ref{fig:dss-main-interface}.
Each panel visualizes the ten most relevant channels for a predicted arousal onset, with the x-axis representing time relative to the suggested arousal onset (vertical line at $t=0$). Blue dots indicate data points with relevance scores above the current threshold, and the color bar to the right encodes the magnitude of feature relevance (dark green: high relevance, light green: lower relevance). Gray dashed lines mark the interval boundaries where an arousal is most likely to start. The top panel shows the medium detail level, while the bottom left and bottom right panels display the low and high detail levels, respectively, corresponding to different thresholds for feature attribution. Legends clarify the graphical elements.}
\label{fig:local-explanations}
\end{figure}

Figure~\ref{fig:local-explanations} illustrates the local explanation of the arousal event shown in Figure~\ref{fig:dss-main-interface}. During the training phase, participants are given the following info text to explain the local explanation:

\begin{quote}
\textbf{The present graphic explains the AI model's prediction for an arousal.} The explanation allows for an understanding of the importance of individual data points and signals in predicting an arousal in the time series. It consists of several elements, described below:
\begin{itemize}
    \item \textbf{Signals:} The graphic shows the ten channels that have the highest overall relevance for the prediction, in descending order from top to bottom.
    \item \textbf{Data points with relevance score above threshold X:} Blue points are placed along the signals. They mark the data points relevant to the prediction, i.e., whose relevance score is above a certain threshold (Threshold X).
    \item \textbf{Threshold:} Depending on the choice of the explanation filter, more or fewer data points are marked as relevant. Choosing brief explanations/medium detail/detailed explanations leads to different thresholds. For example, a low threshold means that many data points are marked as relevant, allowing for more detailed analysis.
    \item \textbf{Boundaries of the interval where an arousal is most likely to start:} Gray dashed lines indicate the interval where the AI model predicts an arousal is most likely to begin.
    \item \textbf{Relevance score:} To the right of the signals is a colored scale indicating the relevance score of the data points. Dark green indicates high relevance, while lighter shades of green indicate lower relevance.
    \item \textbf{Time:} The x-axis represents time, with time 0 marking the suggested arousal onset, which is the most likely time for the arousal to begin. The marked relevant data points relate to this time point.
\end{itemize}
\end{quote}

\subsubsection{Global Explanation}\label{sec:global-explanation-design}

\begin{figure}[!htbp]
\centering
\includegraphics[width=\textwidth]{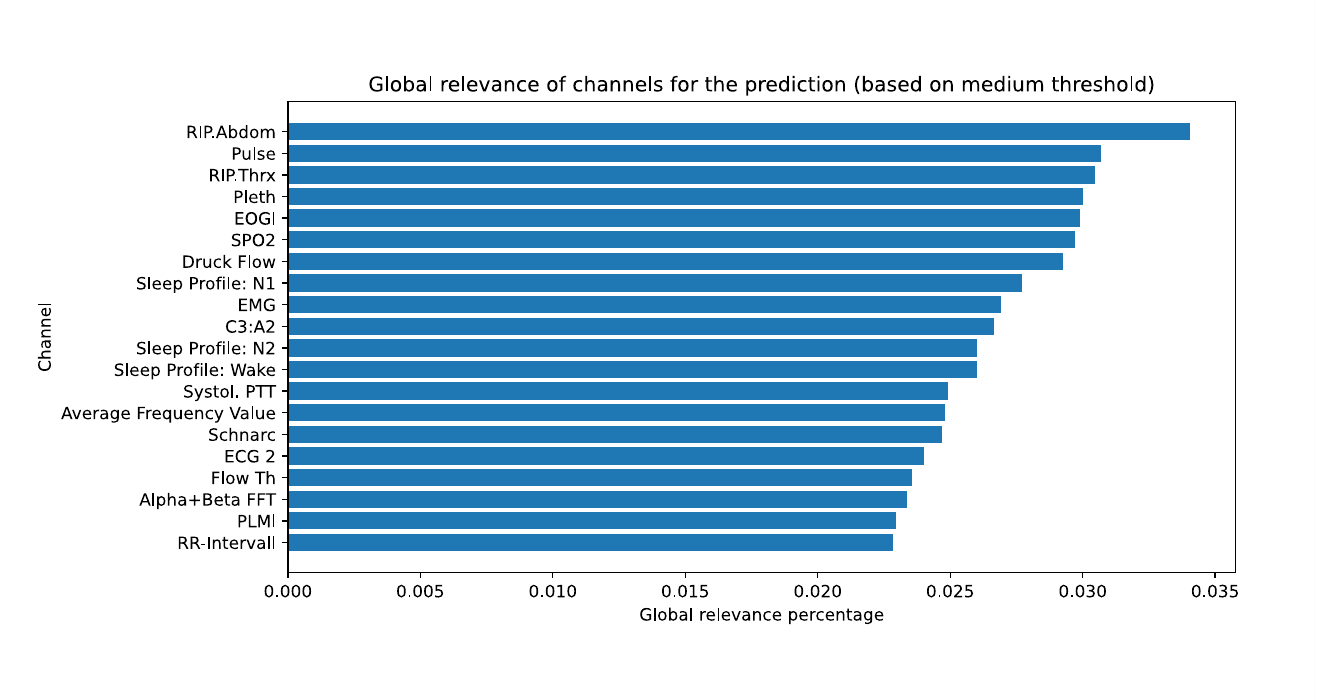}
\caption{\textbf{Global explanation for the AI model's decision making process.}
The bar chart displays the global relevance percentage of each channel for the AI model's arousal prediction, aggregated across the dataset. Channels are ranked in descending order of their contribution, with \textit{RIP Abdomen}, \textit{Pulse}, and \textit{RIP Thorax} showing the highest relevance. This visualization helps users understand which physiological signals most strongly influence the model's decisions at a global level.}
\label{fig:global-explanation}
\end{figure}

The global explanation of the AI model, as described in Section~\ref{sec:global-explanations}, is presented to participants as a bar chart (see Figure~\ref{fig:global-explanation}). Participants are informed that this visualization is generated by aggregating the channel relevance of all local explanations for all arousal events in the dataset, using the medium threshold setting (see Section~\ref{sec:local-explanations-design}).

\subsection{Choice of Dataset and Samples}\label{sec:dataset-and-samples}
The dataset employed in this study is the Comprehensive Polysomnography Dataset (CPS), comprising 113 full-night sleep studies conducted during clinical operations at Klinikum Esslingen, Germany, between 2021 and 2022. This dataset encompasses up to 36 raw data channels and 23 derived channels, capturing a broad spectrum of sleep-related physiological signals. Additionally, it includes annotations for 81 distinct event types, such as arousals induced by respiratory disturbances, limb movements, and spontaneous arousals. For the purposes of this study, all arousal event types were consolidated into a single category to facilitate binary classification. This dataset, along with comprehensive documentation, is publicly accessible on PhysioNet~\citep{kraft2024cps, goldberger2000physiobank}.

Utilizing a recently acquired real-world dataset that closely mirrors the conditions encountered by participants in their professional practice enhances the study's relevance to their expertise and the clinical environment.

In terms of preprocessing, the raw data was subjected to noise reduction using a third-order Butterworth bandpass filter, followed by normalization. The derived channels were resampled to a uniform rate of 256 Hz and normalized to a range between 0 and 1. Padding was applied to ensure uniformity across recordings. Further technical details regarding data preprocessing, hyperparameter configurations, and channel selection are available in previously published work~\citep{pmlr-v287-kraft25a}.

\paragraph{Sample Selection.}

A sample refers to a complete overnight recording of polysomnographic (PSG) data from a patient.
For this study, one sample is designated for the training phase, while four samples are used for the primary segment of the user study. Table \ref{tab:sample-selection} presents the distribution of arousal events and associated performance metrics within the test set of the CPS dataset, ranked according to the F2-score.

\begin{table}[htbp]
\caption{\textbf{Performance Metrics and Statistics of the AI Model for Patient Records in the CPS Dataset Test-Fold.} Refer to Appendix~\ref{appendix:mapping-of-subject-ids} for the mapping to actual sample IDs. Samples are ranked by their F2-score. Detailed phase descriptions are provided in Section~\ref{sec:procedure}.}\label{tab:sample-selection}
\begin{center}
\setlength{\tabcolsep}{3.8pt} %
\begin{tabular}{llllllllll}
\toprule
ID & F2 & F1 & Precision & Recall & TP & FP & FN & \#Events & Usage \\

\midrule

S1 & 0.91 & 0.85 & 0.77 & 0.95 & 349 & 107 & 19 & 475 & Phase 2 (Training) \\
S2 & 0.85 & 0.73 & 0.60 & 0.95 & 198 & 133 & 11 & 342 & Phase 4 (QC,WB) \\
S3 & 0.80 & 0.77 & 0.73 & 0.82 & 192 & 71  & 41 & 304 & Phase 5 (Start,BB) \\
S4 & 0.79 & 0.72 & 0.63 & 0.85 & 245 & 146 & 45 & 436 & Phase 3 (QC,BB) \\ %
S5 & 0.79 & 0.63 & 0.47 & 0.95 & 184 & 212 & 9  & 405 & - \\
S6 & 0.79 & 0.64 & 0.48 & 0.94 & 107 & 114 & 7  & 228 & - \\
S7 & 0.74 & 0.72 & 0.68 & 0.75 & 173 & 81 & 57 & 311 & Phase 6 (Start,WB) \\
S8 & 0.73 & 0.53 & 0.37 & 0.97 & 74  & 128 & 2 & 204 & - \\
S9 & 0.71 & 0.52 & 0.36 & 0.95 & 173 & 314 & 9  & 496 & - \\
S10 & 0.70 & 0.67 & 0.63 & 0.72 & 216 & 125 & 86 & 427 & - \\
S11 & 0.68 & 0.71 & 0.77 & 0.66 & 98  & 30  & 51 & 179 & - \\
S12 & 0.67 & 0.66 & 0.65 & 0.68 & 83  & 45  & 40 & 168 & - \\
S13 & 0.66 & 0.55 & 0.43 & 0.76 & 119 & 159 & 38 & 316 & - \\
S14 & 0.62 & 0.62 & 0.62 & 0.62 & 58  & 35  & 36 & 129 & - \\

\bottomrule
\end{tabular}
\end{center}
\end{table}

The selection of samples for the user study was guided by two primary criteria: (1) The overall performance of the samples should be sufficiently high to facilitate effective learning from the AI assistance by participants. (2) The performance characteristics of the AI model across the samples should be as consistent as possible to minimize bias in the results.

A significant performance disparity in the F2-score of the AI was observed between the top-performing sample (\textit{S1}) and the subsequent samples. Consequently, \textit{S1} was chosen for the initial training phase of the user study. The AI model demonstrated relatively similar F2-score performance on the next three samples, warranting their inclusion in the study. However, two samples (\textit{S5} and \textit{S6}) were excluded due to the AI model's lower precision compared to the already selected samples. Thus, the next best-performing sample (\textit{S7}) was selected for the study. The remaining samples were not included in the study.

\subsubsection{Questionnaires}\label{sec:questionnaires}
Participants were required to complete two detailed questionnaires designed to assess the usability, user experience, and perceived effectiveness of the AI-powered Decision Support System (DSS). These evaluations were conducted under different AI interaction regimes, specifically contrasting black box and white box modes, as well as varying the timing of AI assistance (either from the onset or as a quality control measure) within the context of the arousal scoring task. The surveys examined both the manual and the AI-assisted scoring process. The questionnaires included Likert scale items alongside open-ended questions to facilitate qualitative feedback.

The initial questionnaire was administered following the training session to collect demographic data, document any issues encountered during training, and evaluate the overall usability and user experience of the DSS. This assessment employed the standardized \textit{AttrakDiff} questionnaire, as developed by \citet{hassenzahl2008user}, which evaluates the system's pragmatic and hedonic qualities, as well as its attractiveness.

The second questionnaire was distributed at the study's conclusion. It concentrated on the perceived utility of AI assistance, participants' trust in AI predictions, their comfort level when interacting with AI, and the ease of validating AI predictions. Furthermore, it examined the plausibility of AI explanations, user preferences for upfront explanations versus AI-based quality control, and specific system functionalities, such as preferences regarding the level of detail in AI transparency and insights into potential improvements.

Collectively, these surveys offered a comprehensive perspective on user interactions with the DSS, their confidence in the system, and their views on its transparency and functionality.

\subsection{Study Protocol}\label{sec:study-protocol}

\subsubsection{Main Objectives}\label{sec:objectives}
As highlighted by \citet{mohseni2021multidisciplinary} and \citet{rong2022towards}, it is crucial to explicitly define design goals and evaluation criteria. In this study, we split our objectives into primary and secondary categories. The primary objectives are quantified through objective performance metrics, while the secondary objectives are evaluated via participant questionnaires. Each objective is assigned a unique identifier, such as \textbf{PO1} for the first primary objective and \textbf{SO1} for the first secondary objective. 
Collectively, these objectives aim to address the research questions outlined in Section~\ref{sec:introduction}.

\paragraph{Primary Objective.}
The primary aim of this study is to evaluate the performance of human-AI collaboration in arousal diagnostics. This involves a comparative analysis of transparent AI assistance, black box AI assistance, and manual scoring as a baseline. The specific objectives are to:

\begin{description}
    \item[\textbf{PO1}] Evaluate the level of agreement among human study participants.
    \item[\textbf{PO2}] Assess the performance of human solo, AI solo, and human-AI collaboration using two benchmarks: (1) the consensus ground truth among study participants and (2) the CPS dataset ground truth.
    \item[\textbf{PO3}] Investigate the impact of different timings (AI support from the start versus AI support as quality control) and transparency modes (black box versus white box AI assistance) on the alignment of human-AI teams with the CPS dataset standard.
    \item[\textbf{PO4}] Analyze the time efficiency in the arousal scoring task across varying timing and transparency modes.
\end{description}

\paragraph{Secondary Objectives.}
For transparent AI assistance, we also want to assess:
\begin{description}
    \item[\textbf{SO1}] The degree of trust participants place in AI predictions.
    \item[\textbf{SO2}] The perceived comfort level of participants when interacting with AI assistance.
    \item[\textbf{SO3}] The ease with which participants can validate AI predictions.
    \item[\textbf{SO4}] The plausibility of the explanations provided, including participants' preferred modes of explanation (refer to Section \ref{sec:modes-of-explainability}).
    \item[\textbf{SO5}] The overall satisfaction of participants when utilizing AI assistance.
    \item[\textbf{SO6}] The most beneficial level of detail in explanations for participants: This involves determining whether participants favor explanations highlighting only the most salient features of the model's decision or more comprehensive explanations.
    \item[\textbf{SO7}] The extent to which participants gain insights into the AI's general decision-making process.
    \item[\textbf{SO8}] The appropriateness of the AI model's decision threshold, including whether it should be adjusted to a lower, higher, or user-configurable level.
    \item[\textbf{SO9}] The temporal accuracy of AI predictions: This involves assessing whether the predicted arousal start intervals' proximity to the true arousal starts, as well as the intervals' duration, are sufficient for participants to evaluate the prediction's accuracy.
\end{description}

\subsubsection{Recruitment and Participants}\label{sec:recruitment-and-participants}
In studies requiring domain-specific expertise, \citet{rong2022towards} highlight that evaluations by laypersons may not adequately reflect the practical effectiveness of explanations. Although lay participants are more accessible, expert participants possess the requisite skills to assess the accuracy and reliability of AI predictions, which is particularly crucial in safety-critical domains. However, recruiting experts poses challenges, often resulting in a smaller participant pool, thereby limiting the generalizability of findings, as noted by \citet{rong2022towards}.

During the initial recruitment phase, we encountered notable challenges which are worth mentioning. A significant number of experts declined participation, primarily due to concerns about potential job displacement by AI and the risk of disclosing expert knowledge to the public.

Despite these obstacles, we successfully recruited eight sleep scorers for the user study. The majority are employed by reputable companies such as NRI Medizintechnik GmbH and Löwenstein Medical, which offer sleep diagnostic services to hospitals and sleep laboratories. Others are independent sleep scorers. Each participant received a compensation of 300 EUR. 

\begin{figure}[!htbp]
\centering
\includegraphics[width=1.0\textwidth]{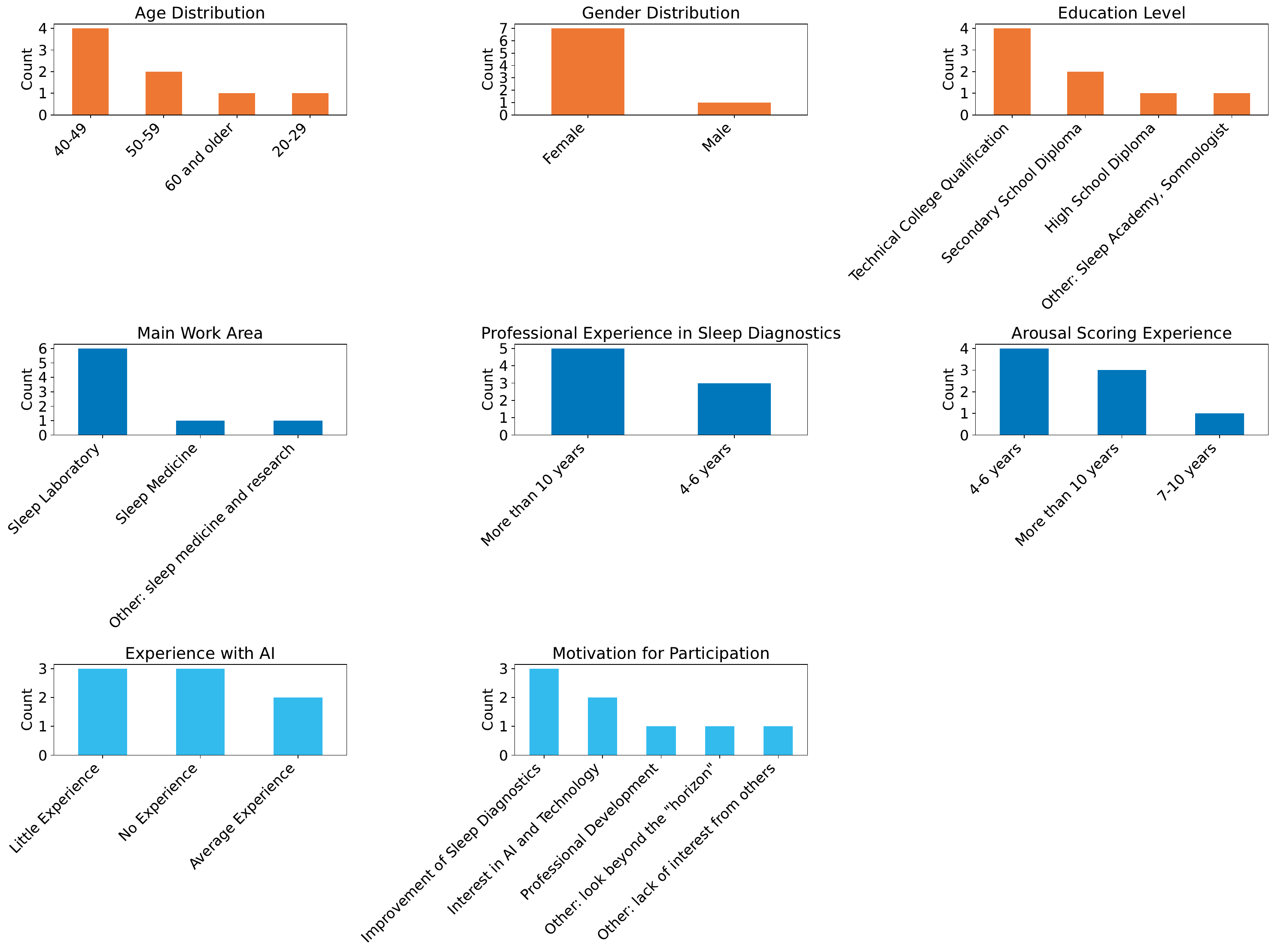}
\caption{Detailed demographic profile of study participants, illustrating basic demographic information (orange), professional experience in sleep diagnostics (blue), and both AI experience and motivation for participation (light blue).}
\label{fig:demographics}
\end{figure}

Figure~\ref{fig:demographics} provides an overview of the study cohort's demographic composition, highlighting several notable characteristics.

The age distribution is relatively balanced, with most participants falling in the 30--40 and 41--50 age brackets, and a smaller representation from the 51--60 and 18--29 groups. Gender distribution is skewed towards female participants, reflecting the composition of the field.
Regarding education, the majority of participants hold either a vocational qualification or a university degree, with a smaller subset possessing advanced academic degrees. 

Most participants are employed in sleep laboratories, while a minority work in other medical contexts. 
Professional experience in sleep diagnostics is substantial: all participants have at least 4 years of experience, and more than half have more than 10 years. Arousal scoring experience closely mirrors this, with most participants reporting between 4 and 10 years, and several exceeding 10 years.

Experience with AI is mixed: while some participants have prior exposure to AI-based systems, others have little or no experience. 
Motivations for participation are diverse, with the most common being the improvement of sleep diagnostics, followed by interest in AI and technology. Additional motivations include professional development, a desire to look beyond one's immediate field, and participation due to a lack of interest from others.

This diversity in background and motivation enriches the study, providing a comprehensive view of user needs and expectations in the context of AI-assisted sleep diagnostics.

\subsubsection{Procedure}\label{sec:procedure}
The user study was conducted remotely, with participants typically working from their home offices. They accessed the Decision Support System (DSS) via a remote desktop connection to a server running the DSS. The study was facilitated through a screen-sharing session, enabling the moderator to observe and communicate with participants. The moderator intervened solely in instances of technical difficulties, protocol deviations, or to address questions about usage and configuration of the DSS.

The study adhered to a structured protocol comprising multiple phases. Participants were provided with written instructions, and new polysomnographic records were utilized in each phase to mitigate learning effects. The study employs a within-subject design, where the same test subjects are employed for all participants across the phases, allowing for a direct comparison of the different AI interaction regimes and timings per participant.

\begin{figure}[!htbp]
\centering
\includegraphics[width=1.0\textwidth]{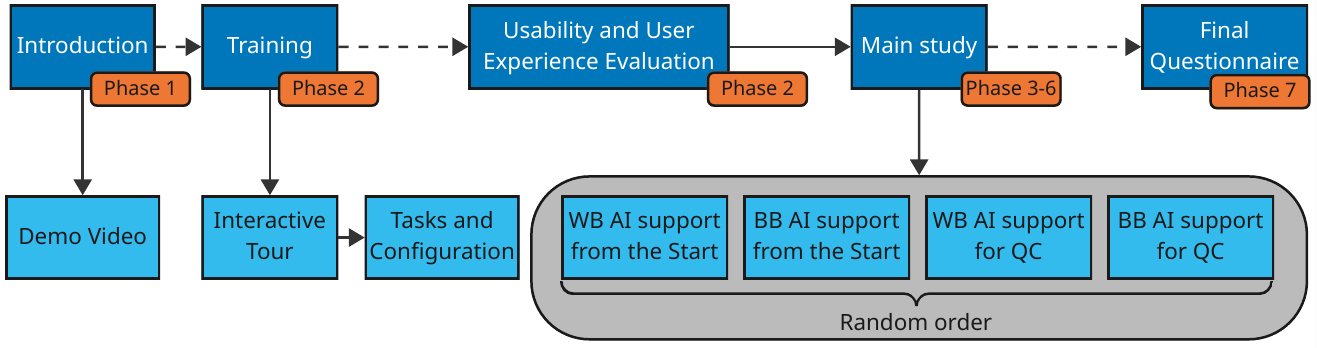}
\caption{\textbf{Overview of the Study Phases.} The phases are annotated in orange. Dashed lines indicate that the next phase is only started after all steps of the previous phase are completed. Abbreviations: \textit{QC} denotes Quality Control, \textit{WB} refers to White Box, and \textit{BB} means Black Box.}
\label{fig:study-phases}
\end{figure}

Figure~\ref{fig:study-phases} provides an overview of the study phases, with a comprehensive description presented subsequently.

Phases 1 and 2, which included the introduction and training, were consistently conducted at the outset. To minimize potential fatigue and learning effects, a counter-balanced design was implemented, randomizing the sequence of Phases 3 through 6 for each participant. Furthermore, to keep the study's time requirements manageable, participants scored subjects only from the beginning of the measurement until 01:00 AM during Phases 3 to 6, corresponding to approximately three hours of sleep time per patient. This approach aligns closely with the application-grounded objective, as participants are not abruptly introduced to an intermediate section of a scoring. Additionally, arousals predominantly occur in the first third of the night~\citep{wetter2012elsevier}. Phase 7, which involved the final questionnaire, was invariably the concluding phase.

\paragraph{Phase 1: Welcoming and General Introduction.} The moderator initiated the session by welcoming participants and verifying their access to the Decision Support System (DSS) and accompanying materials. A pre-recorded video was presented, offering a comprehensive overview of the DSS's features and fundamental operations.

\paragraph{Phase 2 (Training): Training Session and Usability Evaluation.} Participants engaged in a series of tasks aimed at acquainting them with the DSS and allowing for system customization based on individual preferences. This phase incorporated interactive tutorials built into the DSS and practice exercises to facilitate navigation and feature utilization within the DSS. This phase was additionally designed to allow participants the opportunity to tailor the DSS settings according to their preferences. Upon completion, participants were required to fill out a questionnaire assessing usability and user experience.

\paragraph{Phase 3 (QC,BB): Manual Arousal Scoring and Re-Assessment with Black Box AI Assistance.} Participants conducted manual arousal scoring on a new patient record, establishing a baseline for subsequent comparison. They then re-evaluated the same record using the DSS with black box AI assistance, which provided event suggestions without supplementary explanations. This setup enabled participants to adjust their manual scoring based on the AI-generated suggestions.

\paragraph{Phase 4 (QC,WB): Manual Arousal Scoring and Re-Assessment with Transparent AI Assistance.} Participants performed manual arousal scoring on a new patient record, creating a baseline for later comparison. Subsequently, they re-assessed the same record using the DSS with transparent AI assistance. The system offered explanations and visualizations alongside its predictions, including confidence scores, decision thresholds, and both local and global feature importance for each proposed arousal event onset (refer to Section~\ref{sec:modes-of-explainability}). This allowed participants to adjust their manual scoring based on the AI-generated suggestions.

\paragraph{Phase 5 (Start,BB): Arousal Scoring with Black Box AI Assistance.} Participants conducted the arousal scoring task on a new patient record using the DSS with black box AI assistance, which provided event suggestions devoid of additional explanations.

\paragraph{Phase 6 (Start,WB): Arousal Scoring with Transparent AI Assistance.} Participants utilized the DSS to perform arousal scoring with transparent AI assistance, which offered explanations and visualizations alongside its predictions. The system offered explanations and visualizations alongside its predictions, including confidence scores, decision thresholds, and both local and global feature importance for each proposed arousal event onset (refer to Section~\ref{sec:modes-of-explainability}).

\paragraph{Phase 7: Final Questionnaire.} In the concluding phase, participants completed a final questionnaire to evaluate their overall experience with the DSS, the perceived utility of the system, and their experiences in working with the AI-assistance. They were also encouraged to provide feedback and suggestions for potential improvements.

\subsubsection{Provided Resources}\label{subsec:provided-resources}
Participants were provided with the following resources:

\begin{description}
    \item[Written Instructions] Comprehensive written instructions for the user study.
    \item[Moderator] A moderator was present throughout the study to provide guidance.
    \item[Introductory Video] An introductory video that showcased the primary features of the Decision Support System (DSS) and its AI assistance capabilities.
    \item[Interactive Tour] An interactive tour integrated into the DSS.
    \item[Questionnaires] A series of questionnaires was administered to gather participant feedback.
\end{description}

\subsection{Evaluation Approach and Rationale}\label{sec:evaluation-approach-rationale}
\begin{figure}[!htbp]
\centering
\includegraphics[width=\textwidth]{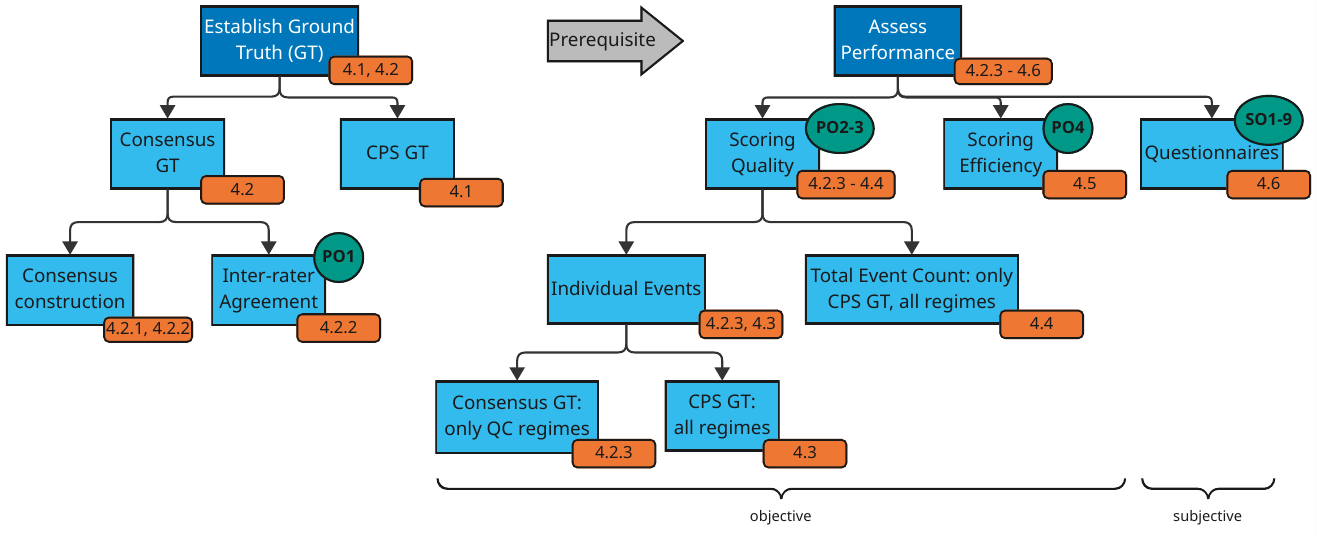}
\caption{\textbf{Illustration of the Evaluation Process.} Sections displaying the results of each step are marked in orange, while the objectives are marked in teal. Abbreviations used: \textit{CPS} denotes the Comprehensive Polysomnography dataset~\citep{kraft2024cps} (see Section~\ref{sec:dataset-and-samples}), \textit{QC} stands for Quality Control, \textit{PO} indicates Primary Objective, and \textit{SO} means Secondary Objective (see Section~\ref{sec:objectives}).}
\label{fig:overview-evaluation}
\end{figure}

The evaluation approach is illustrated in Figure~\ref{fig:overview-evaluation}.

Establishing an adequate ground-truth (GT) is the initial task for our evaluation which is of critical importance~\citep{mohseni2021multidisciplinary}. We will utilize two variants: the \textit{consensus GT} constructed from unaided scoring results of the study participants and the \textit{CPS GT} from the CPS dataset. During this process, we also examine the inter-rater agreement among study participants (\textit{PO1}), which is instrumental in constructing the consensus GT.

Having a GT specifically for human solo performance is essential. Merely comparing different explanation techniques is insufficient, as it does not provide insights into the utility of explanations versus no explanations~\citep{rong2022towards}.

To address the primary objectives \textit{PO2-PO3}, we conduct objective performance evaluations. These evaluations are performed at two levels: per-event and total event count.

The \textit{QC} scoring phases serve a dual purpose. Firstly, they allow us to evaluate the timing of AI assistance (\textit{PO3}) by comparing it with the \textit{Start} scoring phases. Secondly, they enable us to use the manual scoring baseline to construct the consensus GT. This construction helps us assess the unaided inter-rater agreement (\textit{PO1}) and compare the performance of human-AI collaboration with human solo performance (\textit{PO2}). Consequently, the \textit{QC} phases can only be evaluated using the consensus GT (refer to Figure~\ref{fig:overview-evaluation}).

The secondary objectives \textit{SO1-SO9} (see Section~\ref{sec:objectives}) are assessed through participants' subjective experiences. These are gathered using questionnaires that include both quantitative responses, such as Likert scale ratings, and qualitative responses in the form of open-ended questions. Thematic analysis will be applied to open-ended responses to extract deeper insights and identify areas for improvement.

By integrating objective performance metrics with subjective questionnaire responses, such as trust or perceived utility, we aim to provide a comprehensive evaluation of each objective and the overarching research questions. This discussion is carried out in Section~\ref{sec:discussion}.

\subsection{Data Recording and Privacy}\label{sec:data-recording-and-privacy}
Participants are assured of anonymity throughout the study. In addition to collecting scoring data and questionnaire 
responses, the study involves screen and audio recordings, to 
which participants have consented. However, to maintain participant anonymity, these recordings will not be disclosed.

\section{Results}\label{sec:results}

This section presents the findings of the user study, addressing both the primary objectives (\textit{PO1--PO4}) and secondary objectives (\textit{SO1--SO9}) as defined in Section~\ref{sec:objectives}. General considerations about the evaluation approach are discussed in Section~\ref{sec:evaluation-approach-rationale} and the structure is visualized in Figure~\ref{fig:overview-evaluation}.

We begin by outlining the process of ground truth selection in Section~\ref{sec:ground-truth-intro}, where both the consensus ground truth and the CPS ground truth are introduced. 
Section~\ref{sec:consensus-results} begins by analyzing the inter-rater agreement among study participants (\textit{PO1}), which is fundamental to the development of the consensus ground truth. Building on this, the section further investigates how human solo annotators, the AI system, and human-AI teams perform when evaluated against the consensus ground truth. In contrast, Section~\ref{sec:original-results} extends this performance comparison to the CPS ground truth, enabling a thorough assessment of each approach relative to both reference standards (\textit{PO2}) and it further investigates how different timing and transparency modes affect the alignment of human-AI teams (\textit{PO3}). 
In contrast to the preceding event-level analyses, Section~\ref{sec:count_based_eval} extends the evaluation of \textit{PO2} and \textit{PO3} on the CPS ground truth by shifting focus to aggregate arousal counts, providing a clinically oriented perspective on how human, AI, and human-AI teams compare in terms of total event detection across different collaboration regimes.
An analysis of time efficiency (\textit{PO4}) follows in Section~\ref{sec:time-demand}. Finally, Section~\ref{sec:eval-questionnaires} presents the questionnaire results, evaluating the usability and user experience of the web application, providing additional insights into \textit{PO2} and \textit{PO3}, and addressing the secondary objectives (\textit{SO1--SO9}) related to AI assistance. Furthermore, a comparison of feature attributions between DeepLIFT and GradientSHAP is presented in Appendix~\ref{appendix:comparison-of-feature-attribution-methods}.

\subsection{Establishing the Ground Truth and Research Goals}\label{sec:ground-truth-intro}

A central challenge in evaluating performance for the arousal annotation task is the establishment of an appropriate ground truth. In this study, we employ two distinct reference standards: (1) the \textit{CPS ground truth}, which comprises annotations from multiple human experts, with each patient recording scored by a single expert as part of routine clinical practice in the Comprehensive Polysomnography (CPS) dataset; and (2) a \textit{consensus ground truth}, derived from manual, unaided scoring results of user study participants through clustering and probabilistic weighting.

The CPS ground truth reflects the cumulative expertise of about five medical professionals who contributed to the dataset over time during routine clinical practice. While each patient record is annotated by a single expert, the dataset as a whole incorporates the practices of several individuals. This reference standard was also used to train the AI model, potentially predisposing the model to replicate its inherent patterns.

In contrast, the consensus ground truth is designed to mitigate individual biases by synthesizing the judgments of multiple annotators into a unified standard. Specifically, it is constructed by combining the annotations for the two subjects initially scored manually without AI assistance (subjects S2 and S4 in Table~\ref{tab:sample-selection}; see also Section~\ref{sec:procedure}). This process includes annotations from all study participants as well as the CPS ground truth annotations. The resulting consensus, formed via clustering and expectation-maximization weighting (cf.\ Section~\ref{sec:consensus-results}), provides a more balanced and arguably less biased reference than the CPS ground truth.

Accordingly, we assess the performance of humans, AI, and human-AI teams from two perspectives. First, we compare results against the consensus ground truth, which is less likely to favor any single approach and is thus more suitable for claims of objective correctness. Second, we evaluate performance relative to the CPS ground truth, which represents a distinct, albeit potentially more biased, standard. This dual approach enables us to examine whether human-AI teams tend to align more closely with a given standard (the CPS ground truth) compared to unaided human annotators, and how this alignment differs when evaluated against a neutral consensus reference.
For the purposes of statistical inference, we designate the CPS ground truth as the definitive reference for addressing the primary objectives (\textit{PO2} and \textit{PO3}). In contrast, we employ the consensus ground truth to conduct sensitivity analyses, as detailed in Section~\ref{sec:pairwise-inference-cps-primary}.

\subsection{Performance Under the Consensus Ground Truth}\label{sec:consensus-results}

\subsubsection{Clustering Manual Human Annotations}\label{sec:clustering-manual-annotations}
When employing the consensus as the evaluation standard, our objective is to assess the extent to which human solo annotators, the AI system, and human-AI teams align with a collectively derived reference for correct annotations. This approach is grounded in the premise that, in cases where true correctness is ambiguous, a consensus among multiple human experts can serve as a balanced ground truth.

Given the imperfect alignment among human annotations, we first cluster these annotations to delineate regions of agreement and disagreement. 
Recognizing that different clustering algorithms may yield varying results, we selected two representative and conceptually distinct methods for comparison: DBSCAN, a density-based clustering algorithm, and agglomerative clustering, a hierarchical approach (cf.\ Section~\ref{sec:data-preparation}).

For both clustering methods, we set the minimum cluster size to $m^{\text{Cluster}} = 2$, requiring that at least two annotators mark a temporal region for it to be considered a cluster. This conservative threshold is justified by the subsequent probabilistic weighting step, which enables us to filter out unlikely clusters of size two or greater.

Our DBSCAN-based clustering approach is described in Section~\ref{sec:data-preparation}. The key parameters to determine are the epsilon value $\varepsilon$ and the minimum cluster size $m^{\text{Cluster}}$.

\begin{figure}[!htbp]
\centering
\includegraphics[width=1.0\linewidth]{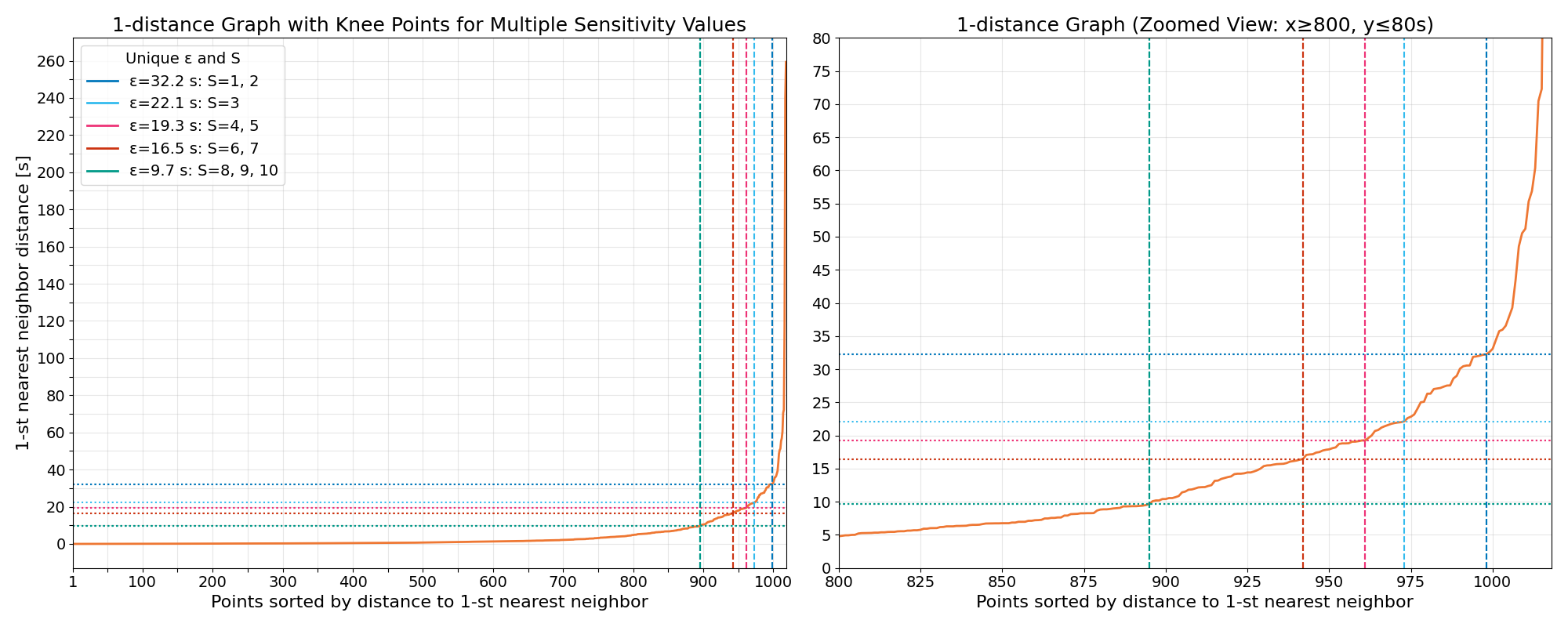}
\caption{\textbf{1-Distance Graph with Knee Points for Multiple Sensitivity Values.} The left graph displays the 1-distance values sorted by distance to the 1st nearest neighbor, with knee points identified for sensitivity values $S \in \{1, \ldots, 10\}$. The right graph provides a zoomed view of the relevant range. These knee points inform the selection of the optimal epsilon parameter $\varepsilon$ for DBSCAN clustering, offering a spectrum of candidate values for robust parameterization.}
\label{fig:images-kneedle}
\end{figure}

To select $\varepsilon$, we apply the Kneedle algorithm to the 1-distance graph for multiple sensitivity values $S$ (see Section~\ref{sec:data-preparation}). This multi-parameter strategy produces a range of candidate $\varepsilon$ values, facilitating manual selection informed by domain expertise and visual inspection of clustering outcomes. The results of this analysis are presented in Figure~\ref{fig:images-kneedle}.

Because the $\varepsilon$ parameter in DBSCAN does not impose a strict upper bound on cluster size, allowing clusters to span multiple $\varepsilon$ distances, we select a relatively low value of $\varepsilon = 9664$ ms, as indicated by sensitivities $S\in\{8,9,10\}$. Here, $\varepsilon$ is best interpreted as the minimum gap between two clusters. Notably, this value aligns well with the American Academy of Sleep Medicine guidelines~\citep{berry2012rules}, which stipulate a minimum of 10 seconds of sleep between arousals.

For agglomerative clustering, we utilize the Scikit-learn implementation and specify a distance threshold $\varepsilon$ above which clusters are no longer merged, using the Ward linkage criterion. As this threshold is absolute, it is logical to base it on the typical duration of an arousal event, estimated by \citet{boselli1998effect} at $14.6 \pm 2.5$ s, irrespective of age. This is consistent with the 15-second temporal tolerance buffer applied in our previous work~\citep{pmlr-v287-kraft25a} for matching predicted and ground-truth events. Given that this is a statistical average, we set the threshold slightly higher at $\varepsilon = 19251$ ms, which remains within the 95\% confidence interval of the arousal duration estimate.

In accordance with the minimum cluster size criterion ($m^{\text{Cluster}} = 2$), all annotations from a single annotator are assigned to a noise cluster (see Section~\ref{sec:data-preparation}).

\subsubsection{Deriving the Consensus Clusters}\label{sec:consensus-clusters}

To identify consensus clusters, i.e.\ those most likely to represent true events, it is instructive to first examine the degree of inter-rater agreement for the different clustering approaches.

\textbf{Inter-Rater Agreement.}
Table~\ref{tab:inter-rater-agreement} reports Krippendorff's Alpha, Fleiss' Kappa, and the mean pairwise Cohen's Kappa for both clustering methods.

\begin{table}[htbp]
\centering
\caption{Inter-rater Agreement for Different Clustering Methods}
\label{tab:inter-rater-agreement}
\begin{tabular}{lccc}
\toprule
Clustering Method & \makecell[tl]{Krippendorff's\\Alpha} & \makecell[tl]{Fleiss'\\Kappa} & \makecell[tl]{Mean Pairwise\\Cohen's Kappa}\\
\midrule
DBSCAN & 0.11 & 0.11 & 0.12\\ 
Agglomerative Clustering & 0.09 & 0.09 & 0.10\\
\bottomrule
\end{tabular}
\end{table}

\begin{figure}[!htbp]
\centering
\includegraphics[width=1.0\linewidth]{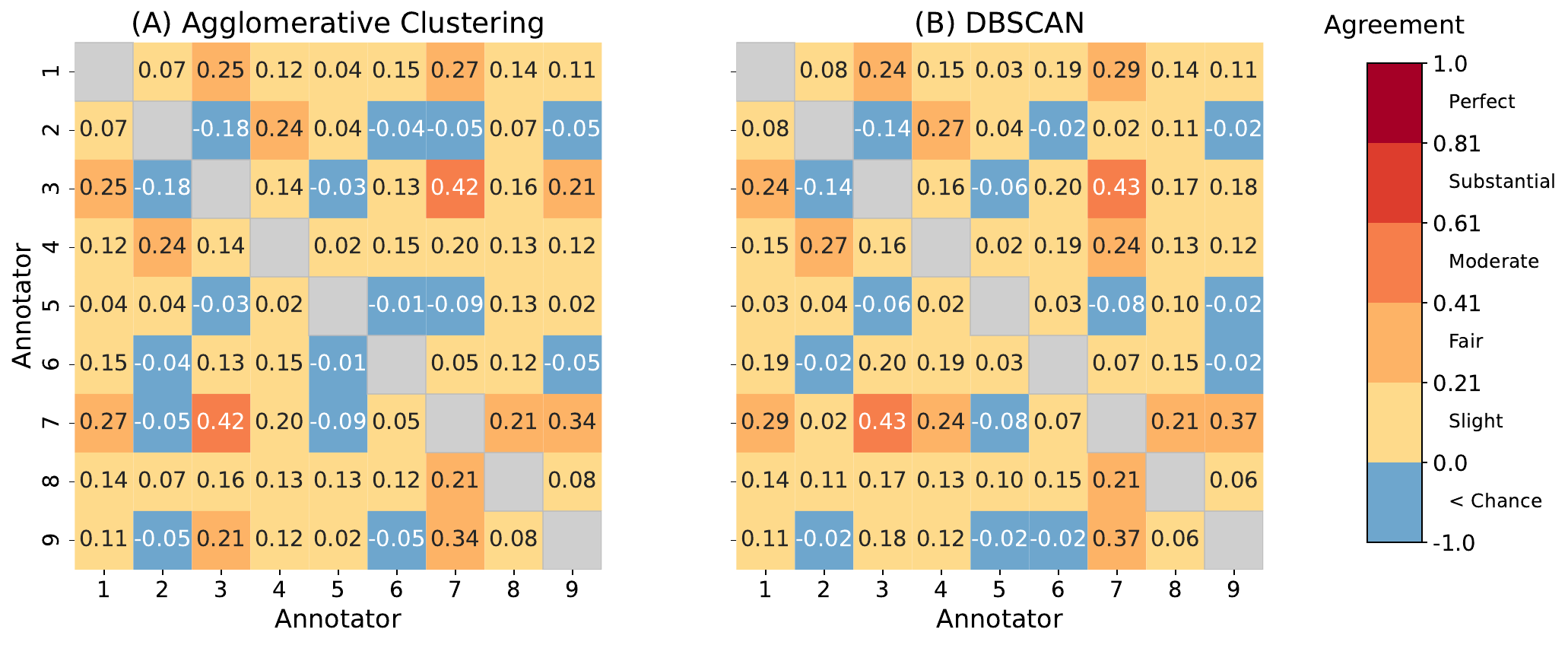}
\caption{\textbf{Pairwise Cohen's Kappa Heatmaps} for the different clustering methods: (A) Agglomerative Clustering and (B) DBSCAN. The color scale indicates the level of agreement, ranging from perfect agreement (dark red) to less than chance agreement (light blue). Annotator 9 corresponds to the human annotations which served as the CPS ground truth for training the AI model.}
\label{fig:pairwise-cohen-kappa-heatmaps}
\end{figure}

All three metrics are marginally higher for DBSCAN than for agglomerative clustering, suggesting slightly greater annotator agreement within DBSCAN-derived clusters. Nevertheless, the absolute values remain low for both methods, indicating only minimal agreement above chance among annotators. To further investigate the inter-rater agreement, we examine pairwise agreement, as visualized in Figure~\ref{fig:pairwise-cohen-kappa-heatmaps}.

The heatmaps in Figure~\ref{fig:pairwise-cohen-kappa-heatmaps} reveal two key findings. First, the overall agreement patterns are similar across both clustering methods. Second, despite generally low agreement, certain annotator pairs demonstrate fair ($\kappa$ between 0.21 and 0.40) or even moderate ($\kappa$ between 0.41 and 0.60) agreement. Concretely, in the DBSCAN results, fair agreement is observed for six annotator pairs (1\&3, 1\&7, 2\&4, 4\&7, 7\&8, 7\&9), and moderate agreement for one pair (3\&7). Conversely, several annotator pairs exhibit very low, or even below-chance, agreement, most notably involving annotator 5.
These findings indicate that, despite low overall agreement, there remains potential to derive a meaningful consensus from the human annotations.

\begin{figure}[!htbp]
\centering
\includegraphics[width=\linewidth]{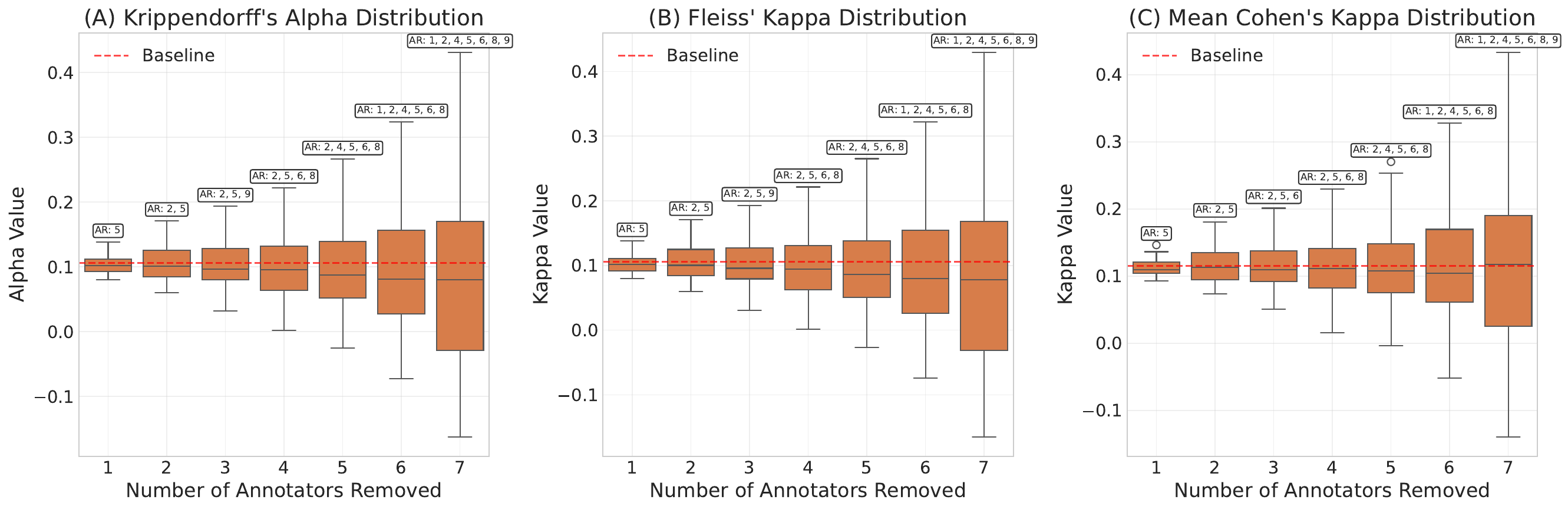}
\caption{\textbf{Sensitivity of inter-rater agreement to systematic annotator removal.}
For each $k\in\{1,\dots,7\}$, box plots show the distribution of Krippendorff’s $\alpha$ (A), Fleiss’ $\kappa$ (B), and mean pairwise Cohen’s $\kappa$ (C), computed across all $\binom{9}{k}$ possible subsets formed by removing $k$ annotators, with the unit (event) set fixed. The red dashed line marks the baseline value when all $n=9$ annotators are included. The labels annotated above the boxplots indicate the particular set of annotators whose removal yielded the highest value for each agreement statistic (``AR'' stands for ``annotators removed''). Thus, while each box shows the spread of agreement for all combinations, only the set achieving the maximum is called out. As $k$ increases, the variability in agreement widens, reflecting both reduced rater numbers and differences across subsets.}
\label{fig:sensitivity-analysis}
\end{figure}

To more rigorously assess sensitivity of the inter-rater agreement statistics, we examine how they vary as subsets of annotators are systematically excluded from the analysis.
Let $n=9$ be the total number of annotators and let each unit correspond to a clustered event.
For each removal size $k\in\{1,\dots,7\}$ we enumerate all $\binom{n}{k}$ combinations of removed annotators. 
For a given combination, we 
recalculate Krippendorff's Alpha, Fleiss' Kappa, and mean pairwise Cohen's Kappa between the remaining $m=n-k$ annotators. 
The same three statistics computed with all $n$ annotators serve as the baseline.
The results of this analysis are presented in Figure~\ref{fig:sensitivity-analysis}.

Across removal sizes the medians remain close to the baseline, while the spread widens markedly for larger $k$.
This pattern is expected: as the number of remaining raters $m=n-k$ decreases, inter-rater agreement estimates become more variable across subsets. 
Importantly, absolute levels across different $k$ are not strictly comparable as each $k$ corresponds to a different $m$. Since Alpha- and Kappa-based statistics are chance-corrected proportions, their magnitude depends not only on who remains but also on $m$ and the label marginals. 
Within a fixed $k$, however, the interquartile range reveals meaningful heterogeneity and supports our former analysis: some subsets of annotators yield substantially higher agreement than others.
Notably, removing annotators 5 and 2 leads to a marked increase in agreement across all three statistics, while the highest overall agreement is achieved between annotators 3 and 7, a pattern that aligns with the observations from Figure~\ref{fig:pairwise-cohen-kappa-heatmaps}.

Nevertheless, to avoid introducing selection bias, we opted not to exclude any annotators from the consensus construction. Instead, we employ a probabilistic weighting scheme based on annotator agreement, utilizing an expectation-maximization algorithm (see Section~\ref{sec:em-truth-quality}).

\begin{figure}[!htbp]
\centering
\includegraphics[width=1.0\linewidth]{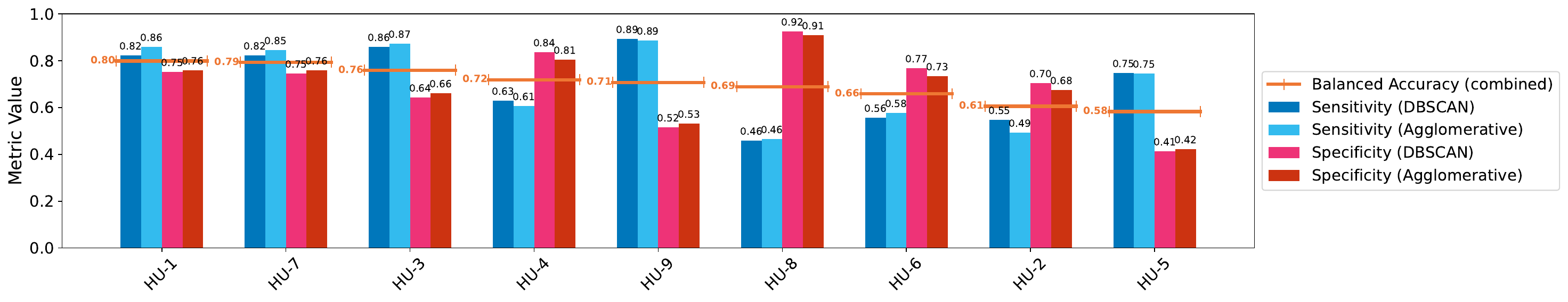}
\caption{\textbf{Annotator Sensitivities, Specificities, and Balanced Accuracies for DBSCAN and Agglomerative Clustering.} For each annotator, sensitivities (blue bars) and specificities (red bars) are shown for both DBSCAN and agglomerative clustering. Annotators are sorted by their overall balanced accuracy, which is defined as the average of sensitivity and specificity for each clustering method, then averaged across both methods. The orange line indicates the balanced accuracy per annotator, highlighting variability in annotation performance. Values for both clustering approaches are closely aligned for most annotators, suggesting robustness of these performance metrics to the choice of clustering method.}
\label{fig:annotator-sensitivities-specificities}
\end{figure}

This approach enables estimation of each annotator's sensitivity and specificity, providing a measure of trustworthiness that informs the consensus annotation process. The analysis is conducted separately for DBSCAN and agglomerative clustering, as illustrated in Figure~\ref{fig:annotator-sensitivities-specificities}.

Both clustering methods yield broadly similar outcomes, yet they also highlight pronounced differences in annotator reliability as reflected by inter-rater agreement. Notably, annotators 1, 7, and 3 achieve the highest balanced accuracy, underscoring their overall trustworthiness. In contrast, annotator 5 stands out for markedly low specificity and the lowest balanced accuracy, signaling a persistent pattern of unreliability relative to its peers.
This low specificity reduces annotator 5's influence when voting against a cluster being an arousal event. However, due to high sensitivity, this annotator may still contribute meaningfully to identifying true arousal events.
Notably, these results are in close agreement with the sensitivity analysis shown in Figure~\ref{fig:sensitivity-analysis}: retaining annotators 1, 3, and 7 produced the highest agreement when five annotators were excluded, while removing annotator 5 led to a notable improvement in agreement. This convergence reinforces the robustness of our findings across analytical approaches.

The sensitivities and specificities from Figure~\ref{fig:annotator-sensitivities-specificities} are now incorporated as weights in the consensus voting scheme.

\begin{figure}[!htbp]
\centering
\includegraphics[width=1.0\linewidth]{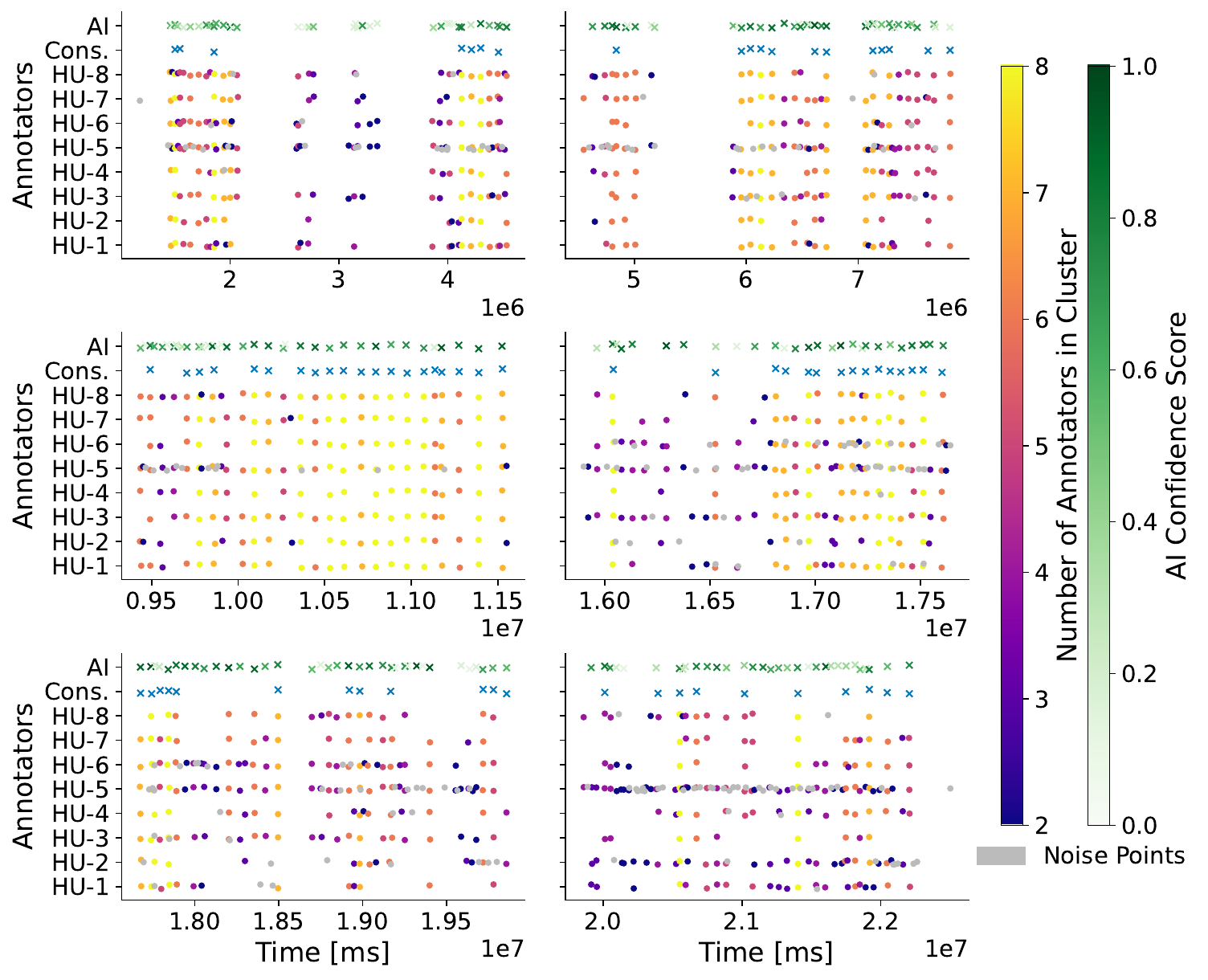}
\caption{\textbf{Faceted Timeline Plot of Clustering Results using DBSCAN Clustering}. The plot displays human solo annotations across multiple panels, with data from patient samples \textit{S2} and \textit{S4} concatenated into approximately six hours of recording time. Annotations of the same color and vertical alignment belong to the same cluster. The horizontal displacement is for visual clarity only. Large intervals without annotations have been omitted. Consensus annotations (\textit{Cons.}) are marked as blue crosses, while AI annotations are marked as green crosses, with the hue representing the AI model's confidence score. The color scale on the right indicates the number of annotators in each cluster. Annotator \textit{HU-9} represents the human annotations used as the CPS ground truth for AI model training.}
\label{fig:faceted-timeline-plot-consensus-dbscan}
\end{figure}

Figure~\ref{fig:faceted-timeline-plot-consensus-dbscan} presents all human annotations, their DBSCAN-based clustering structure, the resulting consensus annotations, and the AI-generated annotations. A corresponding visualization for agglomerative clustering is provided in Appendix~\ref{appendix:consensus-ground-truth}.

Several observations merit emphasis. First, all annotators consistently agree on periods without annotations, which likely correspond to wakefulness, as the American Academy of Sleep Medicine guidelines~\citep{berry2012rules} restrict arousal annotation to sleep periods. Second, distinct annotation \textit{styles} are evident: some annotators (e.g., 2, 4, and 8) annotate sparsely, while others (e.g., 3, 7, and 9) annotate more densely. This heterogeneity may explain the presence of higher pairwise agreements that do not translate into high overall agreement. Finally, the AI annotations visually align closely with those of annotator 9, who provided the CPS ground truth for model training. Notably, the AI system appears to produce few false negatives but a relatively higher number of false positives compared to the consensus annotations, which is consistent with expectations, as it was optimized for the F2-score.

\paragraph{Comparison of Clustering Approaches.}
To assess the reliability and suitability of the consensus event clusters derived from different clustering algorithms, we compare the results of agglomerative clustering and DBSCAN using statistics and pairwise clustering similarity measures. The results are summarized below.

\begin{figure}[!htbp]
\centering
\includegraphics[width=1.0\linewidth]{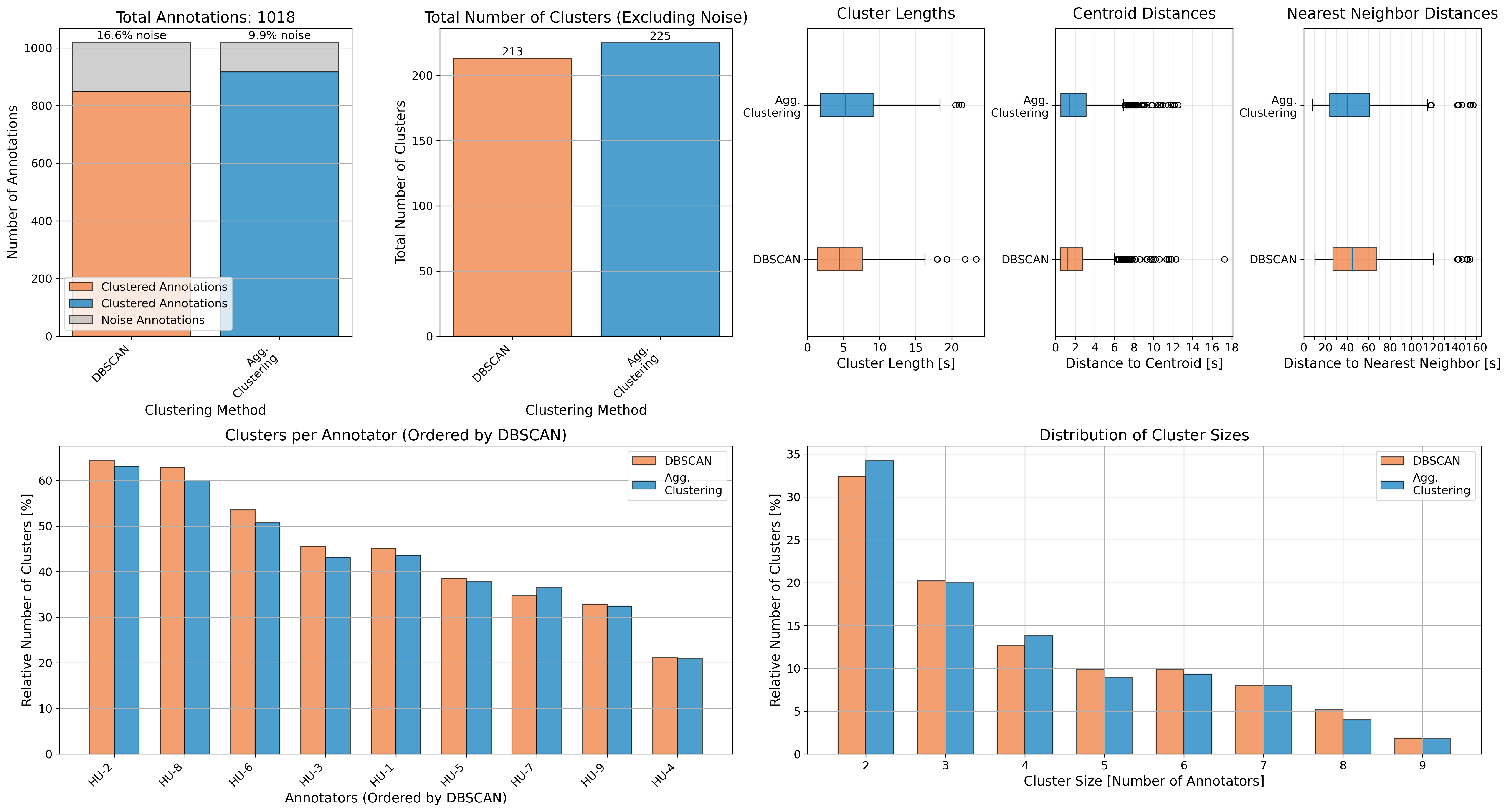}
\caption{\textbf{Clustering Statistics} for DBSCAN and Agglomerative Clustering methods, illustrating various cluster characteristics. From top left to bottom right: Number of clustered versus noise annotations, total number of clusters, cluster length distributions, distances of annotations to their respective cluster centroids or the centroids of their nearest neighbor cluster, relative number of clusters each annotator is part of, and the distribution of cluster sizes. The statistics for both clustering methods are similar, suggesting that the choice of clustering method does not significantly impact the cluster structure.}
\label{fig:clustering-statistics}
\end{figure}

Figure~\ref{fig:clustering-statistics} shows statistics for the clustering approaches.

The analysis highlights several distinctions between DBSCAN and agglomerative clustering. Notably, DBSCAN classifies a greater number of annotations as noise, thereby focusing attention on more meaningful clusters (top left). It also produces slightly fewer, yet more compact, clusters (top right), which may enhance the delineation of cluster boundaries. Furthermore, DBSCAN tends to incorporate a larger number of annotators per cluster (bottom left and bottom right), potentially reflecting a broader consensus among annotators. Despite these differences, both methods yield clusters with broadly similar characteristics, underscoring the robustness of the consensus clusters irrespective of the clustering algorithm employed.

This conclusion is further supported by pairwise similarity measures between the clusterings, as shown in Table~\ref{tab:pairwise-measures}.

\begin{table}[htbp]
\centering
\caption{\textbf{Pairwise Measures for Clustering Methods DBSCAN and Agglomerative Clustering}, including Adjusted Rand Index (ARI) and Normalized Mutual Information (NMI). The initial comparison is between the two clustering methods, the second comparison is between the consensus clusters derived from each method.}
\label{tab:pairwise-measures}
\begin{tabular}{lccc}
\toprule
Comparison & ARI & NMI \\
\midrule
Initial & 0.57 & 0.95 \\
Consensus & 0.90 & 0.95 \\
\bottomrule
\end{tabular}
\end{table}

The results indicate strong to very strong agreement between the clustering methods, as reflected by the Adjusted Rand Index (ARI) and Normalized Mutual Information (NMI) metrics. Specifically, the direct comparison between DBSCAN and agglomerative clustering yields an ARI of 0.57, signifying strong agreement, and an NMI of 0.95, indicating very strong agreement. This suggests that both algorithms identify similar cluster structures, with only minor differences in the assignment of individual data points.

When comparing the consensus clusters derived from each method, the agreement becomes even more pronounced, with an ARI of 0.90 and an NMI of 0.95, both indicative of very strong concordance. This heightened consistency underscores the effectiveness of the consensus-building approach in capturing the underlying structure of the data. The consensus process not only enhances agreement between methods but also demonstrates that the resulting cluster structure is robust and largely independent of the specific clustering algorithm chosen.

\paragraph{Rationale for Selecting DBSCAN Consensus Clusters.}
Our analysis demonstrates that both DBSCAN and Agglomerative Clustering reliably identify well-defined, natural groupings within the data, as evidenced by consistently high values of Normalized Mutual Information (NMI) and Adjusted Rand Index (ARI). These metrics indicate that the choice of clustering algorithm does not substantially impact the resulting consensus events, and the consensus-building process further improves consistency between methods by reducing noise and enhancing cluster quality. The stability of the fundamental cluster structure, reflected in the high NMI values, provides confidence that the clustering results capture genuine patterns in the annotation data rather than artifacts of algorithm selection. Nevertheless, DBSCAN offers specific advantages that render it particularly well-suited to our application, motivating our choice to adopt DBSCAN consensus clusters for subsequent analyses.

\textbf{Enhanced inter-rater agreement}: DBSCAN yields higher inter-rater reliability metrics—including Krippendorff's Alpha, Fleiss' Kappa, and mean pairwise Cohen's Kappa—signifying improved clustering quality and greater annotator consensus.

\textbf{Superior cluster properties}: By discarding a greater proportion of annotations as noise, DBSCAN produces fewer and more compact clusters. This not only sharpens cluster boundaries but also increases the number of annotators represented within each cluster, reflecting stronger expert agreement.

Accordingly, we adopt the consensus clusters generated by the DBSCAN algorithm for all subsequent analyses.

\subsubsection{Human and AI Solo Performance}\label{sec:human-ai-solo-performance}

Using the consensus ground truth, we evaluate the performance metrics of human solo annotators (HU), human-AI teams (HU\&AI), and the AI model.

\paragraph{Distance Threshold Selection.}

As outlined in Section~\ref{sec:per-event-performance}, it is necessary to define a distance threshold to determine whether an annotation should be considered part of a consensus cluster or classified as noise. Figure~\ref{fig:clustering-statistics} (top right) presents the distribution of distances from human solo annotations to the centroids of their assigned consensus clusters, as well as to the centroids of their nearest neighboring clusters. The minimum distance to the nearest neighbor cluster (distinct from the assigned cluster) is approximately 10 seconds, which closely aligns with the chosen DBSCAN parameter of $\varepsilon=9664$ ms. Conversely, only a small number of outlier annotations are located farther than $\varepsilon$ from their own cluster centroid. Consequently, we adopt $\varepsilon$ as the distance threshold for classifying annotations as either part of a consensus cluster or as noise. For human solo annotations, this approach results in only a few annotations being incorrectly labeled as noise, while effectively preventing misattribution of annotations to incorrect clusters.

\paragraph{Results.}

The corresponding results are summarized in Table~\ref{tab:performance-metrics-consensus}.

\begin{table}[htbp]
\centering
\caption{\textbf{Performance comparison with consensus ground truth}. Performance metrics for each human annotator (HU-n), human-AI team (HU-n\&AI) which are aggregated across black box and white box AI assistance for quality control (i.e., phases 3 and 4 as described in Section~\ref{sec:procedure}), and the AI model. HU and HU\&AI denote the averages across all human annotators and human-AI teams, respectively. HU-9 corresponds to the human annotations used as the CPS ground truth for AI model training.}
\label{tab:performance-metrics-consensus}
\begin{tabular}{lccccccc}
\toprule
\textbf{Expert} & \textbf{F1 Score} & \textbf{F2 Score} & \textbf{Precision} & \textbf{Recall} & \textbf{TP} & \textbf{FP} & \textbf{FN} \\
\midrule
HU-1 & 0.70 & 0.76 & 0.61 & 0.82 & 62 & 40 & 14 \\
HU-2 & 0.46 & 0.50 & 0.40 & 0.53 & 40 & 59 & 36 \\
HU-3 & 0.65 & 0.76 & 0.52 & 0.86 & 65 & 59 & 11 \\
HU-4 & 0.62 & 0.62 & 0.62 & 0.62 & 47 & 29 & 29 \\
HU-5 & 0.39 & 0.54 & 0.26 & 0.74 & 56 & 156 & 20 \\
HU-6 & 0.48 & 0.51 & 0.43 & 0.54 & 41 & 54 & 35 \\
HU-7 & 0.69 & 0.75 & 0.60 & 0.80 & 61 & 41 & 15 \\
HU-8 & 0.56 & 0.50 & 0.73 & 0.46 & 35 & 13 & 41 \\
HU-9 (CPS GT) & 0.56 & 0.71 & 0.41 & 0.87 & 66 & 94 & 10 \\
\midrule
HU Avg. & 0.57 & 0.63 & 0.51 & 0.69 & 53 & 61 & 23 \\
\midrule
HU-1\&AI & 0.61 & 0.73 & 0.48 & 0.84 & 64 & 69 & 12 \\
HU-2\&AI & 0.45 & 0.53 & 0.37 & 0.59 & 45 & 78 & 31 \\
HU-3\&AI & 0.62 & 0.75 & 0.48 & 0.88 & 67 & 73 & 9 \\
HU-4\&AI & 0.63 & 0.71 & 0.53 & 0.78 & 59 & 53 & 17 \\
HU-5\&AI & 0.39 & 0.61 & 0.24 & 0.99 & 75 & 236 & 1 \\
HU-6\&AI & 0.56 & 0.72 & 0.41 & 0.88 & 67 & 95 & 9 \\
HU-7\&AI & 0.68 & 0.78 & 0.55 & 0.87 & 66 & 53 & 10 \\
HU-8\&AI & 0.62 & 0.75 & 0.49 & 0.86 & 65 & 67 & 11 \\
\midrule
HU\&AI Avg. & 0.57 & 0.70 & 0.44 & 0.84 & 64 & 90 & 12 \\
\midrule
AI & 0.51 & 0.71 & 0.35 & 0.96 & 73 & 137 & 3 \\
\bottomrule
\end{tabular}
\end{table}

On average, human solo annotators (HU) achieved an F1 score of approximately 0.57, while the AI model alone attained an F1 score of 0.51. 

A one-sample t-test (see Table~\ref{tab:combined-t-tests-consensus}) comparing the AI's F1 score to the distribution of human F1 scores did not reject the null hypothesis of equal performance ($p=0.15$), indicating no statistically significant difference between the AI model and human solo annotators.

\begin{table}[htbp]
\centering
\caption{\textbf{Event-level performance (F1 score) evaluated against the consensus ground truth.}
Statistical methods are described in Section~\ref{sec:pairwise-inference-cps-primary}. Here, $n$ denotes the number of participants. $p$-values are calculated using exact sign-flip tests, and 95\% confidence intervals (CIs) are obtained via percentile bootstrap. Across all comparisons, the null hypothesis of equal performance cannot be rejected.}
\label{tab:combined-t-tests-consensus}
\begin{tabular}{@{}lcccccc@{}}
\toprule
Comparison (A vs B) & $n$ & A mean & B mean & Diff (A$-$B) [95\% CI] & $t(\mathrm{df})$ & $p_{\text{perm}}$ \\
\midrule
HU+AI vs HU & 8 & 0.57 & 0.57 & 0.0031 [$-0.030$, 0.037] & 0.17 (7) & 0.92 \\
HU+AI vs AI & 8 & 0.57 & 0.51 & 0.06 [$-0.0079$, 0.12] & 1.7 (7) & 0.16 \\
HU vs AI    & 9 & 0.57 & 0.51 & 0.06 [$-0.019$, 0.12] & 1.6 (8) & 0.15 \\
\bottomrule
\end{tabular}
\end{table}

\begin{figure}[!htbp]
\centering
\includegraphics[width=1.0\linewidth]{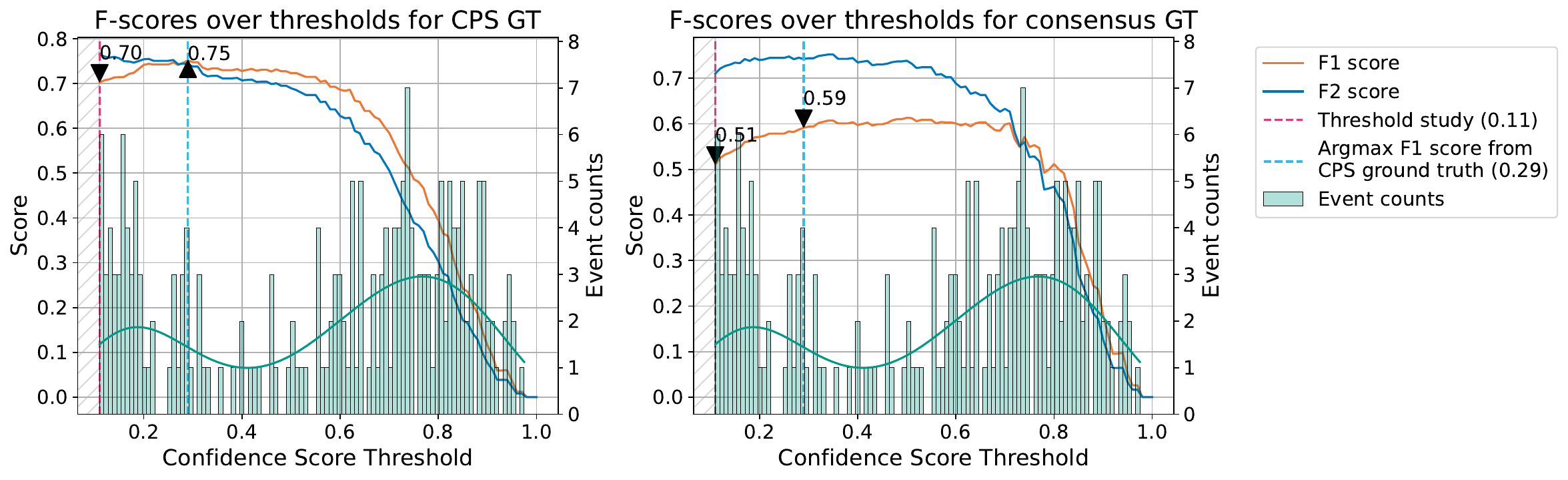}
\caption{\textbf{F1 and F2 Scores over Confidence Score Thresholds} for both CPS ground truth (left) and consensus ground truth (right). The vertical lines indicate the threshold used in the study (0.11) and the threshold corresponding to the maximum F1 score on the CPS ground truth, which is also applied to the consensus ground truth. Maximum F1 score values at these thresholds are annotated. The distribution of event counts is also shown. The AI model demonstrates superior calibration on the CPS ground truth compared to the consensus ground truth.}
\label{fig:f_scores_over_thresholds}
\end{figure}

Several factors should be considered when interpreting these findings. The AI model was optimized for the F2 score, as discussed in~\citet{pmlr-v287-kraft25a}, under the assumption that this would be advantageous for human-AI collaboration compared to optimizing for the F1 score. As a result, the AI's solo performance is somewhat disadvantaged when evaluated using the F1 metric. The potential for a higher F1 score is evident in Figure~\ref{fig:f_scores_over_thresholds}. The left panel displays F1 and F2 scores across a range of decision thresholds, including values higher than the threshold used in the study (0.11), which was selected to maximize the F2 score on the training data. The threshold that maximizes the F1 score is 0.29. Applying this threshold to the AI model on the consensus ground truth (right panel) increases the F1 score from 0.51 to 0.59, slightly surpassing the average human solo performance and further supporting the conclusion of no significant difference between AI and human annotators.

Although, in practice, such a threshold would not be selected based on test data as in this illustrative example, these results demonstrate that optimizing the threshold for the F1 score on training data typically leads to improved F1 performance on the consensus ground truth as well. This observation underscores the argument that the AI model's performance can be further enhanced through appropriate calibration.

It is also important to note that the samples analyzed here (S2 and S4 in Table~\ref{tab:sample-selection}) are among those where the AI performed best in the test set. While human performance may also vary across samples, the AI model may not generalize as robustly as human annotators to other, more challenging cases.

In summary, these findings indicate that the DeepSleep AI model, trained on annotations from multiple human scorers (one per sample), is capable of achieving human-level performance when evaluated against the consensus of expert annotators.

\subsubsection{Human-AI Team Performance}\label{sec:human-ai-team-performance}
We next examine the performance of human-AI teams relative to human solo performance, as assessed against the consensus ground truth annotations. In this setting, human scorers first completed manual scoring and were subsequently permitted to revise their annotations based on AI-generated suggestions for quality control (see phases 3 and 4 in Section~\ref{sec:procedure}).

\begin{figure}[!htbp]
\centering
\includegraphics[width=1.0\linewidth]{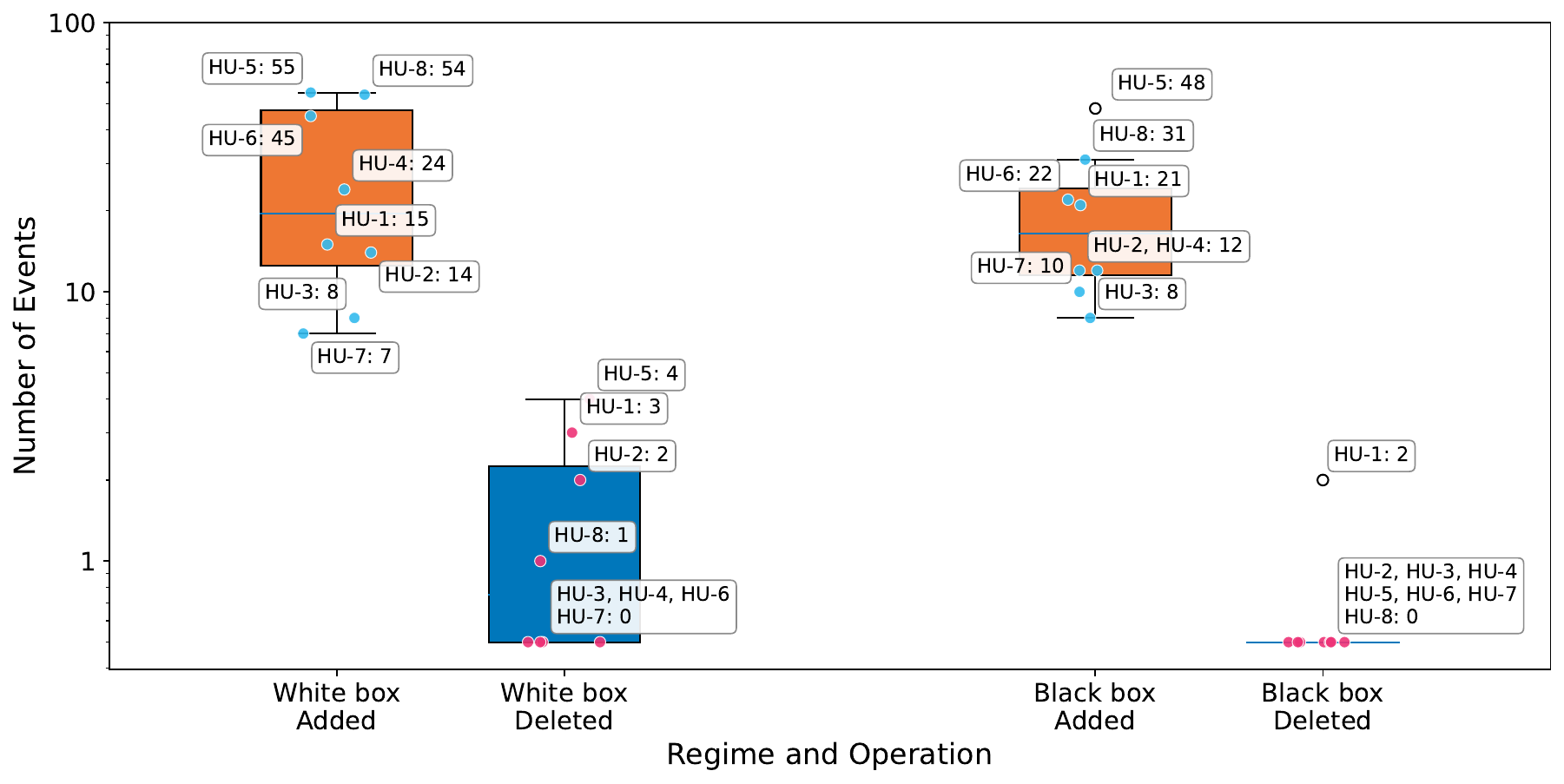}
\caption{\textbf{Distribution of Added and Deleted Events} after humans went through an AI assisted quality control phase. This plot illustrates the number of events added and removed across different regimes (White box and Black box) for each human annotator (HU-n). The distributions are depicted on a logarithmic scale to enhance readability, given the substantial disparity between the number of added and deleted events.}
\label{fig:distribution_added_deleted_events_qa}
\end{figure}

As shown in Table~\ref{tab:performance-metrics-consensus}, the incorporation of AI assistance does not, on average, enhance collaborative performance compared to human solo performance when measured by the F1 score. This finding is further supported by the paired t-test results in Table~\ref{tab:combined-t-tests-consensus}. Although recall improves with AI support, this benefit is offset by a reduction in precision. 
To better understand this outcome, we refer to Figure~\ref{fig:distribution_added_deleted_events_qa}, which depicts the distributions of added and deleted events for human-AI teams compared to human solo performance across all annotators.

The figure reveals a pronounced imbalance between the number of added and deleted annotations across regimes. In the white box regime, annotators added a substantial number of events, with HU-5 and HU-8 contributing most prominently. Conversely, annotation deletions were infrequent, with half of the annotators not deleting any events. In the black box regime, the tendency to add annotations persisted, though generally at a lower rate than in the white box regime, and only one annotator deleted any events. These patterns suggest that white box AI suggestions exert a stronger influence on annotators than black box suggestions. Moreover, black box AI support appears to further diminish the already limited inclination of scorers to critically reassess their own annotations compared to white box support.

Given that Table~\ref{tab:performance-metrics-consensus} demonstrates the AI model introduces a substantial number of false positives, we conclude that, when evaluated against the consensus ground truth, human annotators are generally unable to effectively leverage AI suggestions for quality control. On average, they are unable to reliably distinguish between correct and incorrect AI-generated annotations.

Finally, we highlight a notable outlier among the human annotators: HU-5. This annotator produced substantially more annotations than the other experts (see Table~\ref{tab:performance-metrics-consensus}), leading to low pairwise Cohen's Kappa values in Figure~\ref{fig:pairwise-cohen-kappa-heatmaps} and the lowest specificity for consensus construction (see Figure~\ref{fig:annotator-sensitivities-specificities}). This in turn resulted in the lowest F1 score and precision among all experts (see Table~\ref{tab:performance-metrics-consensus}). 
Interestingly, while AI assistance for quality control reduced HU-5's number of false negatives to the lowest among all experts, it also led to a substantial increase in false positives, since HU-5 predominantly added events suggested by the AI in the quality control phase (see Figure~\ref{fig:distribution_added_deleted_events_qa}). Observationally, HU-5 was also the only expert who, during phase 5 of the study, chose to validate only the AI suggestions when annotating with black box AI assistance from the outset, effectively skipping all time steps not highlighted by the AI. This behavior indicates a particularly high level of trust and reliance on the AI model.
We will further investigate this annotator in subsequent sections.

\subsection{Performance Under the CPS Ground Truth}\label{sec:original-results}

Having established that AI assistance does not enhance human scorer performance when evaluated against the consensus ground truth, we now turn to a related question: To what extent can human-AI teams align with the specific standard on which the AI model was trained, i.e.\ the CPS ground truth, compared to unaided human annotators?

The CPS ground truth serves as the reference standard for the CPS dataset, derived from routine clinical practice. For each subject, annotations were provided by a single medical expert, with the specific expert varying across subjects.

Although this ground truth is not entirely neutral, as it is reflecting the individual styles of the contributing scorers, it provides a concrete standard for assessing alignment. In the previous section, the CPS ground truth was represented by one human expert (\textit{HU-9}). This perspective shifts the research question: Do human-AI teams conform more closely to a given standard than human annotators working independently?

\subsubsection{Comparison of Human-AI Team, Human Solo, and AI Solo Performance}\label{sec:performance-metrics-original}

Table~\ref{tab:performance-metrics-original} presents the comparative results.

\begin{table}[htbp]
\centering
\caption{\textbf{Performance Comparison with CPS Ground Truth} used for AI Model Training. Performance metrics are reported for each human annotator (HU-n), human-AI team (HU-n\&AI), and the AI model. HU and HU\&AI denote the averages across all humans and all human-AI teams, respectively. 
}
\label{tab:performance-metrics-original}
\begin{tabular}{lccccccc}
\toprule
\textbf{Expert} & \textbf{F1 Score} & \textbf{F2 Score} & \textbf{Precision} & \textbf{Recall} & \textbf{TP} & \textbf{FP} & \textbf{FN} \\
\midrule
HU-1 & 0.49 & 0.43 & 0.63 & 0.40 & 64 & 38 & 96 \\
HU-2 & 0.36 & 0.32 & 0.47 & 0.29 & 47 & 52 & 113 \\
HU-3 & 0.55 & 0.51 & 0.63 & 0.49 & 78 & 46 & 82 \\
HU-4 & 0.39 & 0.32 & 0.61 & 0.29 & 46 & 30 & 114 \\
HU-5 & 0.44 & 0.48 & 0.39 & 0.51 & 82 & 130 & 78 \\
HU-6 & 0.32 & 0.28 & 0.43 & 0.26 & 41 & 54 & 119 \\
HU-7 & 0.56 & 0.49 & 0.72 & 0.46 & 73 & 29 & 87 \\
HU-8 & 0.29 & 0.22 & 0.62 & 0.19 & 30 & 18 & 130 \\
\midrule
HU Avg. & 0.42 & 0.38 & 0.56 & 0.36 & 58 & 50 & 102 \\
\midrule
HU-1\&AI & 0.59 & 0.56 & 0.65 & 0.54 & 86 & 47 & 74 \\
HU-2\&AI & 0.44 & 0.41 & 0.50 & 0.39 & 62 & 61 & 98 \\
HU-3\&AI & 0.57 & 0.55 & 0.61 & 0.54 & 86 & 54 & 74 \\
HU-4\&AI & 0.54 & 0.49 & 0.65 & 0.46 & 73 & 39 & 87 \\
HU-5\&AI & 0.56 & 0.70 & 0.43 & 0.83 & 133 & 178 & 27 \\
HU-6\&AI & 0.53 & 0.54 & 0.53 & 0.54 & 86 & 76 & 74 \\
HU-7\&AI & 0.61 & 0.56 & 0.71 & 0.53 & 85 & 34 & 75 \\
HU-8\&AI & 0.61 & 0.58 & 0.67 & 0.56 & 89 & 43 & 71 \\
\midrule
HU\&AI Avg. & 0.56 & 0.55 & 0.60 & 0.55 & 88 & 66 & 72 \\
\midrule
AI & 0.70 & 0.76 & 0.62 & 0.81 & 130 & 80 & 30 \\
\bottomrule
\end{tabular}
\end{table}

Both the AI model and the average human-AI team outperform the average human solo annotator, primarily due to higher recall, with a modest improvement in precision as well. 
The substantial performance gap between the AI model and the human solo average is statistically significant (see Table~\ref{tab:combined-t-tests-original}), which is expected given that the AI was trained on this distribution. 

\begin{table}[htbp]
\centering
\caption{\textbf{Event-level performance (F1-score) evaluated against the CPS ground truth.}
Pairwise comparisons follow the methodology outlined in Section~\ref{sec:pairwise-inference-cps-primary}, with each comparison based on $n=8$ participants. Permutation $p$-values are computed using exact two-sided sign-flip tests, and 95\% confidence intervals (CIs) are estimated via the percentile bootstrap method. No adjustment for multiple comparisons is necessary, as each primary question constitutes a single statistical family. As all human-AI teams outperform their respective human solo counterparts and the AI solo outperforms all others (see Table~\ref{tab:performance-metrics-original}), the sign of all differences is the same, resulting in identical permutation $p$-values across comparisons. In all cases, the null hypothesis is rejected at the 5\% significance level.}
\label{tab:combined-t-tests-original}
\begin{tabular}{@{}lccccc@{}}
\toprule
Comparison (A vs B) & A mean & B mean & Diff (A$-$B) [95\% CI] & $t(\mathrm{df})$ & $p_{\text{perm}}$ \\
\midrule
HU+AI vs HU & 0.56 & 0.42 & 0.13 [0.075, 0.20] & 3.9 (7) & 0.012 \\
HU+AI vs AI & 0.56 & 0.70 & $-0.15$ [$-0.19$, $-0.11$] & $-7.4$ (7) & 0.012 \\
HU vs AI    & 0.42 & 0.70 & $-0.28$ [$-0.34$, $-0.21$] & $-7.8$ (7) & 0.012 \\
\bottomrule
\end{tabular}
\end{table}

Table~\ref{tab:inter-rater-agreement} further reveals only fair agreement between one participant (annotator 7) and the CPS ground truth (annotator 9), and slight agreement for four other participants (annotators 1, 3, 4, and 8).

A key question is whether human-AI team performance differs significantly from human solo performance. To address this, we conducted a paired t-test, with results shown in Table~\ref{tab:combined-t-tests-original}.

The p-value of $0.012$ is below the significance threshold, providing evidence that human-AI team performance is significantly different from human solo performance when evaluated against the CPS ground truth. This finding stands in stark contrast to the results based on the consensus ground truth (Section~\ref{sec:consensus-results}), where no such improvement was observed. Here, AI-generated suggestions appear to guide human annotators toward the style or criteria embodied in the CPS ground truth.

These results indicate that, when the objective is to align with a specific established standard (such as the CPS ground truth used for AI training), AI assistance can indeed help human annotators move closer to that benchmark.

\paragraph{Detailed Analysis of Human-AI Team Performance}

\begin{figure}[h!]
\centering
\includegraphics[width=\textwidth]{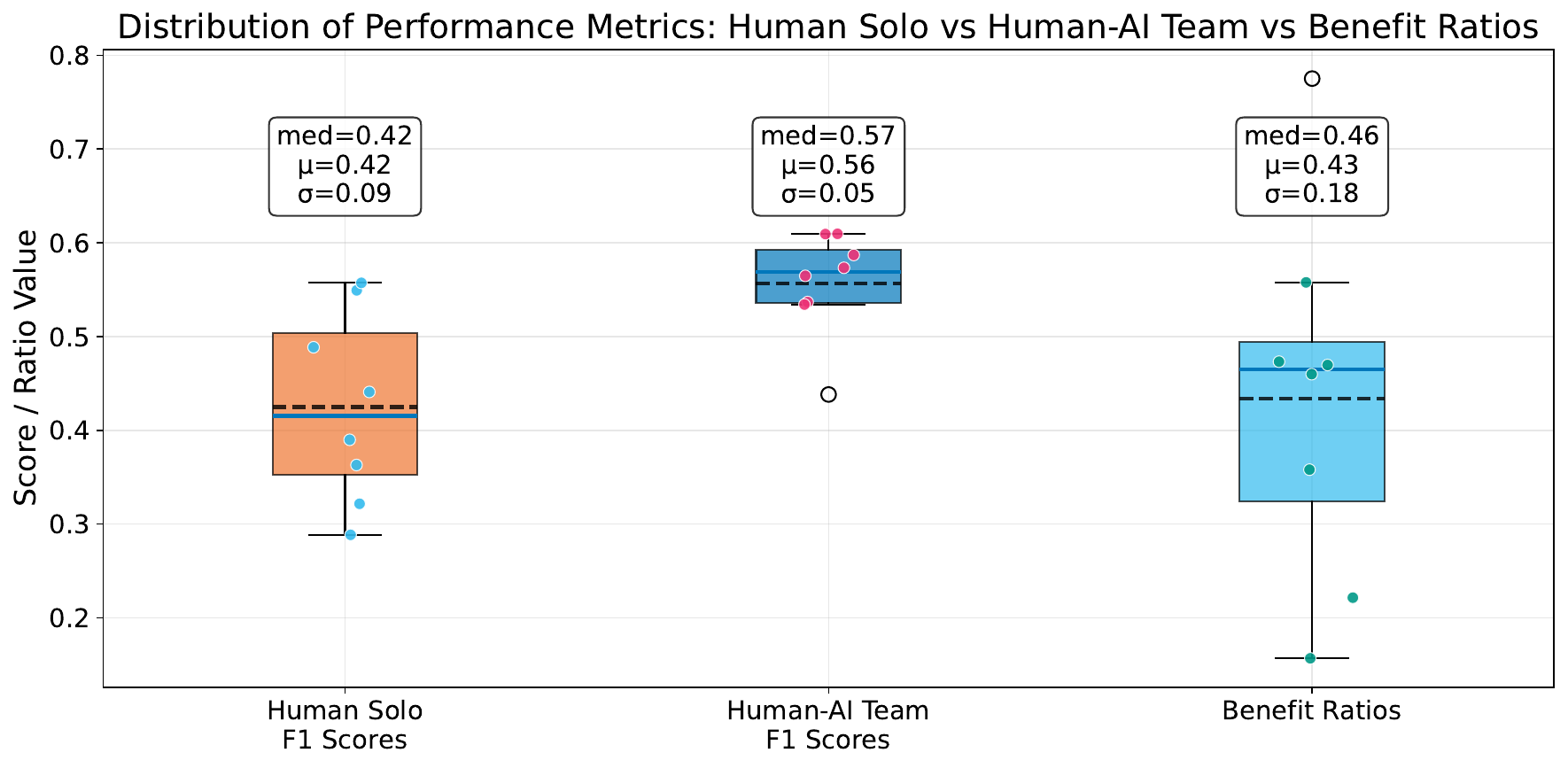}
\caption{\textbf{Distribution of Performance Metrics:} F1 scores for human solo annotation, human-AI team collaboration, and the corresponding benefit ratios. Boxplots display the mean (black dashed line), interquartile range (box), whiskers, individual data points (colored circles), and outliers (empty circles). The data reveal that AI assistance substantially improves human performance and consistency, though benefit ratios vary considerably across teams.}
\label{fig:performance_metrics_boxplot}
\end{figure}

The boxplot in Figure~\ref{fig:performance_metrics_boxplot} offers further insight into the performance distributions of human solo annotators, human-AI teams, and the benefit ratio, which quantifies the effectiveness of AI integration (see Section~\ref{sec:per-event-performance}).

The distribution of human solo F1 scores has a mean ($\mu=0.42$) and standard deviation ($\sigma=0.09$), reflecting notable variability in individual performance. The median is approximately 0.41, with the interquartile range (IQR) spanning roughly 0.36 to 0.50.

In contrast, human-AI team F1 scores exhibit a higher mean ($\mu=0.56$) and a lower standard deviation ($\sigma=0.05$), indicating that AI collaboration not only enhances performance but also yields more consistent results across teams. The median team F1 score is around 0.57, with a narrower IQR from approximately 0.53 to 0.60. This upward shift in both mean and consistency underscores the positive impact of AI assistance. An outlier near 0.43 suggests that, in at least one instance, a human-AI team performed closer to the human solo baseline.

The benefit ratio distribution has a mean ($\mu=0.43$) numerically close to the human solo F1 mean, but its interpretation is distinct. Notably, the benefit ratio exhibits a much larger standard deviation ($\sigma=0.18$) than either F1 distribution, indicating substantial variability in how effectively different teams leveraged AI support. The median benefit ratio is approximately 0.46, suggesting that, on average, teams reached about 46\% of the potential alignment improvement offered by the AI. The presence of an outlier at approximately 0.78 indicates that at least one team was highly effective in adopting the AI's alignment, while the lower end of the distribution reflects cases with minimal benefit.

In summary, these findings demonstrate that human-AI collaboration generally results in higher and more consistent F1 scores compared to human solo annotation. However, the broad distribution of benefit ratios highlights considerable variability in the extent to which teams capitalize on AI assistance. This suggests that factors beyond the AI's intrinsic capabilities, such as the design of human-AI interaction, user training, or task complexity, may substantially influence the realized benefits of AI support.

\subsubsection{Analysis of AI Assistance Regimes}\label{sec:analysis-ai-assistance-regimes}

As outlined in Section~\ref{sec:procedure}, we aggregated human-AI team performance across experimental phases into AI support regimes. Table~\ref{tab:performance-comparison-regimes} presents detailed outcomes for each regime, including both atomic and composite regimes (the latter formed by concatenating samples across phases which leads to metric scores on the micro-level). The table reports the average human-AI team F1 score, the standalone AI F1 score, and the \emph{relative F1 scores} $\mathcal{R} = \mathrm{F1}_{\text{HU+AI Avg.}} / \mathrm{F1}_{\text{AI}}$ (see also Section~\ref{sec:per-event-performance}).

\begin{table}[htbp]
\centering
\caption{\textbf{F1 score comparison across AI support regimes}, ordered by the relative F1 scores $\mathcal{R}$. Abbreviations: Start denotes AI support from the outset; QC indicates AI support used exclusively for quality control after initial manual scoring; BB and WB refer to black box and white box AI support, respectively. Regimes with a ``+'' are composite regimes, constructed by concatenating samples from the respective phases. All others are atomic regimes. $\mathrm{F1}_{\text{HU+AI Avg.}}$ is the average human performance with AI support, $\mathrm{F1}_{\text{AI}}$ is the AI solo performance, and $\mathcal{R}$ are the relative F1 scores of $\mathrm{F1}_{\text{HU+AI Avg.}}$ to $\mathrm{F1}_{\text{AI}}$. Phases are detailed in Section~\ref{sec:procedure}. Values are means with $95\%$ confidence intervals.}
\label{tab:performance-comparison-regimes}
\begin{tabular}{@{}l@{}llccl@{}}
\toprule
\multicolumn{2}{c}{Regime} & Phase & $\mathrm{F1}_{\text{HU+AI Avg.}}$ & $\mathrm{F1}_{\text{AI}}$ & $\mathcal{R}$ \\
\midrule
\multicolumn{1}{r}{QC,}         & WB       & 4       & 0.59 [0.53, 0.65] & 0.66 & 0.90 [0.81, 0.98] \\
\multicolumn{1}{r}{Start+QC,}   & WB       & 4, 6    & 0.57 [0.53, 0.61] & 0.65 & 0.87 [0.81, 0.94] \\
\multicolumn{1}{r}{Start,}      & WB       & 6       & 0.54 [0.50, 0.58] & 0.64 & 0.84 [0.78, 0.89] \\
\multicolumn{1}{r}{Start,}      & BB+WB    & 5, 6    & 0.56 [0.52, 0.61] & 0.70 & 0.80 [0.73, 0.86] \\
\multicolumn{1}{r}{QC,}         & BB+WB    & 3, 4    & 0.56 [0.51, 0.60] & 0.70 & 0.79 [0.73, 0.86] \\
\multicolumn{1}{r}{Start,}      & BB       & 5       & 0.58 [0.51, 0.65] & 0.74 & 0.78 [0.68, 0.88] \\
\multicolumn{1}{r}{Start+QC,}   & BB       & 3, 5    & 0.55 [0.49, 0.60] & 0.75 & 0.73 [0.66, 0.80] \\
\multicolumn{1}{r}{QC,}         & BB       & 3       & 0.52 [0.47, 0.57] & 0.75 & 0.69 [0.63, 0.75] \\
\bottomrule
\end{tabular}
\end{table}

\paragraph{Analysis Overview}
We analyze the log ratio of relative F1 scores $\mathcal{R} = \mathrm{F1}_{\text{HU+AI Avg.}} / \mathrm{F1}_{\text{AI}}$ with a small offset ($\varepsilon=10^{-8}$) to avoid zeros, following the statistical modeling approach described in Section~\ref{sec:statistical-modeling-and-analysis-scale}.

\paragraph{Primary Analysis: RM-ANOVA with Planned Orthogonal Contrasts}
Following the methodology described in Section~\ref{sec:repeated-measures-anova-and-planned-contrasts}, we conduct RM-ANOVA and planned orthogonal contrasts. The results are summarized in Table~\ref{tab:rm-anova-main-effects}.

\begin{table}[p]
\centering
\caption{\textbf{RM-ANOVA summary for main and interaction effects.} Each effect has $df_\text{effect}=1$ and $df_\text{error}=7$. Effect sizes are reported as partial eta-squared $\eta^2_{\mathrm{p}}$. We find that the AI main effect and interaction effect are significant, while the timing main effect is not significant.}
\label{tab:rm-anova-main-effects}
\begin{tabular}{@{}lccc@{}}
\toprule
Effect & $F(1,7)$ & $p$ & $\eta^2_{\mathrm{p}}$ \\
\midrule
\textbf{AI main (WB vs.\ BB)} & \textbf{15.268} & \textbf{0.0058} & \textbf{0.686} \\
Timing main (Start vs.\ QC)& 0.596 & 0.466 & 0.078 \\
\textbf{AI $\times$ Timing Interaction} & \textbf{10.540} & \textbf{0.014} & \textbf{0.601} \\
\bottomrule
\end{tabular}
\end{table}

The RM-ANOVA reveals a significant main effect of \textit{AI} and a significant \textit{AI$\times$Timing} interaction, both associated with large effect sizes. In contrast, the main effect of \textit{Timing} does not reach significance. The ANOVA employs parametric F-tests, while the planned contrasts use permutation tests with Holm correction, accounting for the different p-values between approaches.

\paragraph{Secondary Analysis: Planned-contrast t-tests}
We conduct planned-contrast tests for the main effects and their interaction, following the methodology described in Section~\ref{sec:permutation-testing-and-multiple-comparisons}. Given the significant interaction in the ANOVA, we examine \emph{simple effects} (comparing WB vs.\ BB at both Start and QC, and Start vs.\ QC within both WB and BB conditions) using the approach outlined in Section~\ref{sec:simple-effects-analysis}. Results are presented in Tables~\ref{tab:anova-contrast-summary} and~\ref{tab:simple-effects-summary}.

\begin{table}[p]
\centering
\caption{\textbf{Planned-contrast test results for main and interaction effects in the $2{\times}2$ repeated-measures design.} Each row reports the paired $t$-test statistic, permutation $p$-value with familywise error control via Holm correction (applied to the two main effects while the interaction uses raw permutation $p$), the estimated effect ratios (with bootstrap $95\%$ confidence interval), and within-subject Cohen's $d_z$ effect size (with bootstrap $95\%$ confidence interval). These results align with the RM-ANOVA findings, demonstrating that white box (WB) AI assistance yields a statistically significant and substantial improvement in relative F1 performance compared to black box (BB) assistance (an average increase of approximately \textbf{18\%}), and that the interaction effect is also statistically significant.}
\label{tab:anova-contrast-summary}
\begin{tabular}{@{}lcccc@{}}
\toprule
Contrast & $t(7)$ & $p_{\text{perm}}$ & Ratio [95\% CI] & $d_z$ [95\% CI] \\
\midrule
\textbf{AI main} & \textbf{3.91} & \textbf{0.039} & \textbf{1.18} [\textbf{1.09}, \textbf{1.27}] & \textbf{1.38} [\textbf{0.59}, \textbf{4.68}] \\
Timing main & 0.77 & 0.46 & 1.03 [0.96, 1.08] & 0.27 [--0.37, 2.47] \\
\textbf{Interaction} & \textbf{--3.25} & \textbf{0.027} & \textbf{0.83 [0.75, 0.92]} & \textbf{--1.15 [--2.51, --0.60]} \\
\bottomrule
\end{tabular}
\end{table}

\begin{table}[p]
\centering
\caption{\textbf{Simple-effect contrasts examining the significant interaction in the $2{\times}2$ repeated-measures design.} Each row reports a paired $t$-test for a specific simple effect and reports Holm-corrected permutation $p$-values. Effect sizes are given as within-subject Cohen’s $d_z$ with bootstrap-derived $95\%$ confidence intervals. Effect ratios are accompanied by $95\%$ bootstrap confidence intervals. We see that the benefit of white box (WB) AI assistance over black box (BB) is most pronounced during the quality control (QC) phase, with an average improvement of \textbf{30\%}.}
\label{tab:simple-effects-summary}
\begin{tabular}{@{}lcccc@{}}
\toprule
Simple effect & $t(7)$ & $p_{\text{perm, Holm}}$ & Ratio [95\% CI] & $d_z$ [95\% CI] \\
\midrule
WB vs.\ BB at Start & 1.27 & 0.41 & 1.08 [0.96, 1.20] & 0.45 [--0.23, 2.22] \\
\textbf{WB vs.\ BB at QC} & \textbf{6.27} & \textbf{0.047} & \textbf{1.30 [1.20, 1.40]} & \textbf{2.22 [1.57, 4.49]} \\
Start vs.\ QC with WB & --1.40 & 0.41 & 0.94 [0.86, 1.02] & --0.50 [--1.80, 0.19] \\
Start vs.\ QC with BB & 2.99 & 0.11 & 1.13 [1.04, 1.20] & 1.06 [0.31, 3.45] \\
\bottomrule
\end{tabular}
\end{table}

Our findings indicate that white box (WB) assistance improves relative F1 scores compared to black box (BB) assistance by approximately \textbf{18\%} on average (AI main effect), with a large within-subject effect size ($d_z=1.38$), reflecting substantial enhancement in relative F1 performance. The interaction effect is also pronounced ($d_z=1.15$ in absolute value), and the simple effects analysis reveals that the benefit of AI transparency is most pronounced during the QC phase. Specifically, at QC, WB outperforms BB by about \textbf{30\%} on average, with a very large effect size ($d_z=2.22$). In contrast, the difference between WB and BB at Start is small to moderate ($d_z=0.45$) and not statistically significant. The timing difference within BB is large ($d_z=1.06$), reaching significance prior to multiplicity correction, but not after familywise error control.

\subsection{Count-based evaluation of arousal scoring regimes}
\label{sec:count_based_eval}

\begin{figure}[htbp]
\centering
\includegraphics[width=\textwidth]{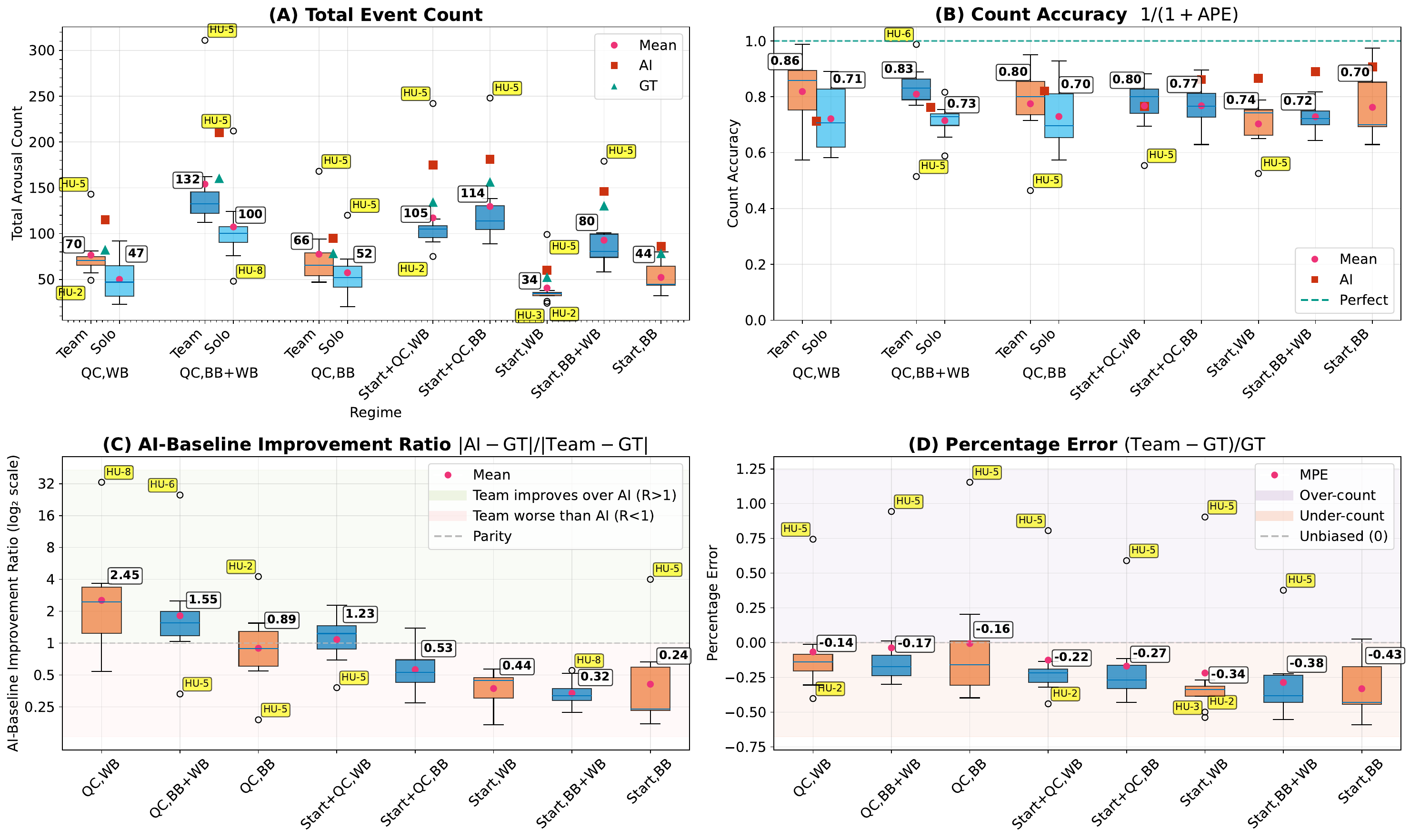}
\caption{\textbf{Count-based evaluation across regimes.}
Panel (\textbf{A}) displays the total arousal counts per recording for each scoring source: human solo scorers ($C_{\mathrm{HU}}$, available only in the QC regime), human-AI teams ($C_{\mathrm{HU+AI}}$), AI-only outputs ($C_{\mathrm{AI}}$, represented by red squares), and the ground truth ($C_{\mathrm{GT}}$, represented by green triangles). 
Panel (\textbf{B}) presents the ground truth count accuracy $A_{x,\mathrm{GT}}$. 
Panel (\textbf{C}) illustrates the ground truth error reduction, visualized as the base-2 logarithm of the error reduction ratio, $\log_2 R_{\mathrm{GT}}$. 
Panel (\textbf{D}) shows the percentage error $\mathrm{PE}$, where the mean value corresponds to the mean percentage error ($\mathrm{MPE}$). 
In all panels, medians are indicated by annotations, means are depicted as magenta points, and outliers are labeled by scorer ID.}
\label{fig:combined_regime_analysis}
\end{figure}

\paragraph{Rationale and link to objectives}
While event-level performance evaluation is valuable for understanding scorer behavior during the diagnostic arousal annotation task of polysomnographic patient data, clinical decision-making typically relies on more aggregated statistics, most notably, the arousal index, defined as the number of arousals per hour of sleep~\citep{berry2012rules, ehrlich2024state}. 
To address the primary objectives \textit{PO2} and \textit{PO3} (see Section~\ref{sec:objectives}) we therefore analyze absolute arousal counts against the CPS ground truth on three measures (see Section~\ref{sec:total-event-count-performance} for definitions): (i) \emph{count accuracy} $A_{x,\mathrm{GT}}$, (ii) the \emph{AI-Baseline Improvement Ratio} $R_{\mathrm{GT}}$ comparing the team to the AI (analyzed on $y_{\mathrm{RGT}}{=}\log R_{\mathrm{GT}}$), and (iii) the \emph{percentage error} $\mathrm{PE}$ giving a measure of systematic count bias toward over- or under-counting. The results are visualized in Figure~\ref{fig:combined_regime_analysis}.

\paragraph{Descriptive patterns across scoring regimes}
Analysis of the distributional patterns in Figure~\ref{fig:combined_regime_analysis} reveals systematic differences in scoring behavior across experimental conditions. Panel A demonstrates that the AI consistently overestimates arousals relative to ground truth by approximately 10--40\%, whereas human-AI teams tend to underestimate by about 20--50\%. This contrast reflects a fundamental difference in approach: human scorers adopt a conservative strategy, while the AI model is optimized for recall. The provision of explanations (WB) reduces inter-participant variability, with this stabilizing effect most pronounced in QC conditions.

\paragraph{Performance hierarchy across atomic regimes}
Figure~\ref{fig:combined_regime_analysis}B demonstrates that the QC-WB regime yields the highest median accuracy among all conditions, with teams achieving approximately 86\% accuracy compared to 80\% for the QC-BB and 74\% for the Start-WB regimes. Panel C indicates that teams operating in the QC-WB regime achieve the greatest error reduction relative to the solo AI (mean ratio of 2.45), while teams in Start-based regimes perform worse than the AI alone (ratios below 1.0). Panel D reveals that teams in QC regimes experience substantially reduced under-counting bias compared to those utilizing AI assistance from the start, with QC-WB showing the least negative bias (-14\%) compared to Start-BB (-43\%).

\paragraph{Scope of inference and design}
Following the statistical modeling approach described in Section~\ref{sec:statistical-modeling-and-analysis-scale}, we analyze the log-transformed AI-Baseline Improvement Ratio $y_{\mathrm{RGT}}=\log R_{\mathrm{GT}}$ and percentage error $\mathrm{PE}$. Within QC we additionally compare team to human solo via paired $t$-tests on $A_{\mathrm{GT}}$ using the methodology outlined in Section~\ref{sec:permutation-testing-and-multiple-comparisons}.

All inferential results are restricted to the four atomic regimes (Start/QC $\times$ WB/BB), following the experimental design outlined in Section~\ref{sec:experimental-design-and-scope-of-inference}.

\paragraph{Statistical Analysis of AI-Baseline Improvement Ratio}
The omnibus analyses in Table~\ref{tab:count_rgt_anova} reveal a significant main effect of Timing, accompanied by a large effect size ($\eta^2_{\mathrm{p}}{=}0.66$), while neither the AI main effect nor the interaction reach statistical significance.

\begin{table}[p]
\centering
\caption{\textbf{RM-ANOVA results for $y_{\mathrm{RGT}}{=}\log R_{\mathrm{GT}}$.} All effects have $df_\text{effect}{=}1$ and $df_\text{error}{=}7$. Partial eta-squared ($\eta^2_{\mathrm{p}}$) quantifies effect size. Only the Timing main effect is significant.}
\label{tab:count_rgt_anova}
\begin{tabular}{@{}lccc@{}}
\toprule
Effect & $F(1,7)$ & $p$ & $\eta^2_{\mathrm{p}}$ \\
\midrule
AI main(WB vs.\ BB) & 2.2 & 0.18 & 0.24 \\
\textbf{Timing main (Start vs.\ QC)} & \textbf{14} & \textbf{0.0075} & \textbf{0.66} \\
Timing $\times$ AI Interaction& 2.2 & 0.18 & 0.24 \\
\bottomrule
\end{tabular}
\end{table}

The planned contrasts in Table~\ref{tab:count_rgt_contrasts} confirm that, on average, teams reduce the error rate in the \emph{QC} condition relative to \emph{Start} by approximately a factor of four (ratio = 0.26, 95\% CI: [0.15, 0.53]), as indicated by the Timing main contrast. This substantial improvement aligns with the visual patterns observed in Figure~\ref{fig:combined_regime_analysis}C, where QC regimes consistently show ratios above 1.0 (indicating team improvement over AI) while Start regimes fall below 1.0. The statistical methodology is described in Section~\ref{sec:repeated-measures-anova-and-planned-contrasts}.

\begin{table}[p]
\centering
\caption{\textbf{Planned-contrast test results for $y_{\mathrm{RGT}}$ (back-transformed as ratios) in the $2{\times}2$ repeated-measures design.} Each row presents the paired $t$-test statistic, permutation $p$-value with familywise error control via Holm correction (applied to the two main effects while the interaction uses raw permutation $p$), the estimated effect ratios (with bootstrap $95\%$ confidence intervals), and within-subject Cohen's $d_z$ effect size (with bootstrap $95\%$ confidence intervals). The results align with the RM-ANOVA findings, demonstrating a significant main effect of Timing. Specifically, the QC condition improves over the Start condition on average by approximately a factor of four.}
\label{tab:count_rgt_contrasts}
\begin{tabular}{@{}lcccc@{}}
\toprule
Contrast & $t(7)$ & $p_{\text{perm}}$ & Ratio [95\% CI] & $d_z$ [95\% CI] \\
\midrule
AI (WB vs.\ BB) & 1.5 & 0.18 & 1.6 [0.87, 2.8] & 0.53 [--0.15, 2.1] \\
\textbf{Timing (Start vs.\ QC)} & \textbf{--3.7} & \textbf{0.039} & \textbf{0.26 [0.15, 0.53]} & \textbf{--1.3 [--5.0, --0.45]} \\
Interaction & --1.5 & 0.25 & 0.32 [0.079, 1.2] & --0.53 [--1.3, 0.15] \\
\bottomrule
\end{tabular}
\end{table}

\paragraph{Statistical Analysis of Percentage Error $\mathrm{PE}$}
The statistical results in Tables~\ref{tab:count_bias_anova} and \ref{tab:count_bias_contrasts} indicate a significant effect of Timing (Start vs.\ QC) on percentage error $\mathrm{PE}$, which quantifies systematic over- or under-counting relative to ground truth. 

\begin{table}[p]
\centering
\caption{\textbf{RM-ANOVA for percentage error $\mathrm{PE}$.} Each effect has $df_\text{effect}{=}1$ and $df_\text{error}{=}7$. Only the Timing main effect is significant.}
\label{tab:count_bias_anova}
\begin{tabular}{@{}lccc@{}}
\toprule
Effect & $F(1,7)$ & $p$ & $\eta^2_{\mathrm{p}}$ \\
\midrule
AI main (WB vs.\ BB) & 0.30 & 0.60 & 0.041 \\
\textbf{Timing main (Start vs.\ QC)} & \textbf{17} & \textbf{0.0044} & \textbf{0.71} \\
Timing $\times$ AI Interaction & 0.75 & 0.41 & 0.097 \\
\bottomrule
\end{tabular}
\end{table}

\begin{table}[!htbp]
\centering
\caption{\textbf{Planned-contrast test results for percentage error $\mathrm{PE}$ (raw scale) in the $2{\times}2$ repeated-measures design.} Each row reports the paired $t$-test statistic, permutation $p$-value with Holm correction for the two main effects (the interaction uses the raw permutation $p$), the estimated effect differences with bootstrap-derived $95\%$ confidence intervals, and within-subject Cohen's $d_z$ effect size (also with bootstrap $95\%$ confidence intervals). These results are consistent with the RM-ANOVA, revealing a significant main effect of Timing. On average, the percentage error is significantly lower in the QC condition compared to Start, with a reduction of approximately $24$ percentage points.}
\label{tab:count_bias_contrasts}
\begin{tabular}{@{}lcccc@{}}
\toprule
Contrast & $t(7)$ & $p_{\text{perm}}$ & Estimate [95\% CI] & $d_z$ [95\% CI] \\
\midrule
AI (WB vs.\ BB) & 0.55 & 0.60 & 0.026 [--0.057, 0.12] & 0.19 [--0.84, 0.96] \\
\textbf{Timing (Start vs.\ QC)} & \textbf{--4.1} & \textbf{0.023} & \textbf{--0.24 [--0.34, --0.14]} & \textbf{--1.5 [--2.8, --1.1]} \\
Interaction & 0.87 & 0.47 & 0.17 [--0.16, 0.56] & 0.31 [--0.56, 0.97] \\
\bottomrule
\end{tabular}
\end{table}

The RM-ANOVA shows a highly significant main effect of Timing ($p = 0.0044$), indicating a large effect size ($\eta^2_{\mathrm{p}} = 0.71$). This is supported by the planned-contrast analysis, where the Timing main contrast yields a Holm-adjusted permutation $p$-value of $0.023$ and a substantial effect estimate of -0.24 (95\% CI: [-0.34, -0.14]). This means that, on average across AI transparency levels, QC regimes reduce under-counting by about 24 percentage points on the $B$-scale, relative to Start (absolute shift towards zero), consistent with the visual patterns in Figure~\ref{fig:combined_regime_analysis}D. In contrast, neither the AI transparency main effect (WB vs.\ BB) nor the Timing $\times$ AI interaction reach statistical significance in either the omnibus or planned-contrast tests at this sample size.

\paragraph{Comparative Analysis: Teams versus Human Solo Performance}

\begin{table}[!htbp]
\centering
\caption{\textbf{Team vs.\ solo on GT accuracy} $A_{\mathrm{GT}}$ \textbf{within QC}. Differences are team minus solo. $p_{\text{perm,Holm}}$: Holm-adjusted within the two QC cells.}
\label{tab:count_team_solo}
\begin{tabular}{@{}lcccccc@{}}
\toprule
Regime & $n$ & Team mean & Solo mean & Diff [95\% CI] & $t(7)$ & $p_{\text{perm,Holm}}$ \\
\midrule
QC, WB & 8 & 0.82 & 0.72 & 0.098 [--0.042, 0.22] & 1.4 & 0.46 \\
QC, BB & 8 & 0.78 & 0.73 & 0.046 [--0.040 0.12] & 1.0 & 0.46 \\
\bottomrule
\end{tabular}
\end{table}

Qualitative inspection of raw annotated arousal counts (Figure~\ref{fig:combined_regime_analysis}A) reveals that, in both QC regimes, the median arousal count for teams is closer to the ground truth than that of human solo scorers. Notably, in the QC-WB condition, the distribution of team counts is substantially tighter than for solo scoring, with reduced inter-participant variability. 

Quantitatively, the coefficient of variation (CV) in QC-WB decreases from 0.47 for human solo to 0.37 for teams, and the mean absolute error (MAE) relative to the ground truth drops from 35 to 21. In the QC-BB condition, improvements are more modest, with the CV decreasing from 0.53 to 0.51 and the MAE reducing from 31 to 27. These improvements are consistent with the visual patterns in Figure~\ref{fig:combined_regime_analysis}B, where QC-WB shows the highest accuracy values.

The results from Table~\ref{tab:count_team_solo} on the \textit{count accuracy} confirm that teams tend to exceed human solo performance in QC, especially with WB (mean difference = 0.098, 95\% CI: [-0.042, 0.22]), but the differences do not reach statistical significance at this sample size.

\paragraph{Individual Scorer Heterogeneity and Adaptive Strategies}
Participant heterogeneity is evident across all panels of Figure~\ref{fig:combined_regime_analysis} and provides valuable insights into regime effectiveness. A particularly illustrative case is scorer \textit{HU-5}, whose behavior reveals complex patterns of AI interaction and adaptive strategies.

\textit{HU-5} consistently appears as an outlier across all panels, exhibiting systematic over-counting behavior. In Panel A (Total Event Count), \textit{HU-5} records the highest arousal counts across most regimes, with especially pronounced values under QC conditions. This over-counting pattern is mirrored in Panel D (Percentage Error), where \textit{HU-5} consistently displays positive values, indicating persistent overestimation relative to ground truth.

Consequently, \textit{HU-5} achieves lower accuracy than other teams (Panel B) and performs worse than the AI alone in several regimes (Panel C). A notable exception occurs in the Start-BB regime, where \textit{HU-5} achieves the highest performance among all teams. This outcome results from an adaptive strategy documented through direct observation: in this final regime, \textit{HU-5} reviewed only AI-suggested events, leaving intervals between suggestions unscored and accepting nearly all AI-suggested events. Given the AI's strong performance in the Start-BB regime, this selective reliance strategy proves particularly effective.

This case demonstrates the dynamic interplay between user trust, strategy selection, and regime design, highlighting how individual differences in AI reliance can substantially influence performance outcomes.

\paragraph{Summary and Clinical Implications}
The count-based evaluation establishes a clear hierarchy of factors influencing human-AI team performance in arousal scoring. \emph{Timing} emerges as the primary determinant, with QC-based workflows substantially outperforming Start-based approaches across all evaluated metrics. Teams utilizing AI during QC achieve greater error reduction relative to AI alone ($R_{\mathrm{GT}}$ ratio of 0.26, $p = 0.032$) and exhibit significantly reduced count bias compared to ground truth ($B = -0.24$, $p = 0.023$). In contrast, \emph{AI transparency} confers only modest, non-significant benefits, particularly for error reduction in QC contexts. The interaction between Timing and AI transparency is likewise non-significant, indicating that the advantage of QC is robust across both WB and BB conditions. 

These findings have important implications for clinical implementation. QC-based workflows produce the most favorable distributional patterns.
The visual evidence in Figure~\ref{fig:combined_regime_analysis} demonstrates that QC-WB brings teams closer to ground truth than any other regime, supporting a workflow that separates an initial human pass from a quality control step, with preliminary evidence suggesting that transparency may provide additional stabilization without undermining human judgment.

\subsection{Time Demand and Efficiency of Arousal Scoring}\label{sec:time-demand}
\begin{figure}[htbp]
\centering
\includegraphics[width=\textwidth]{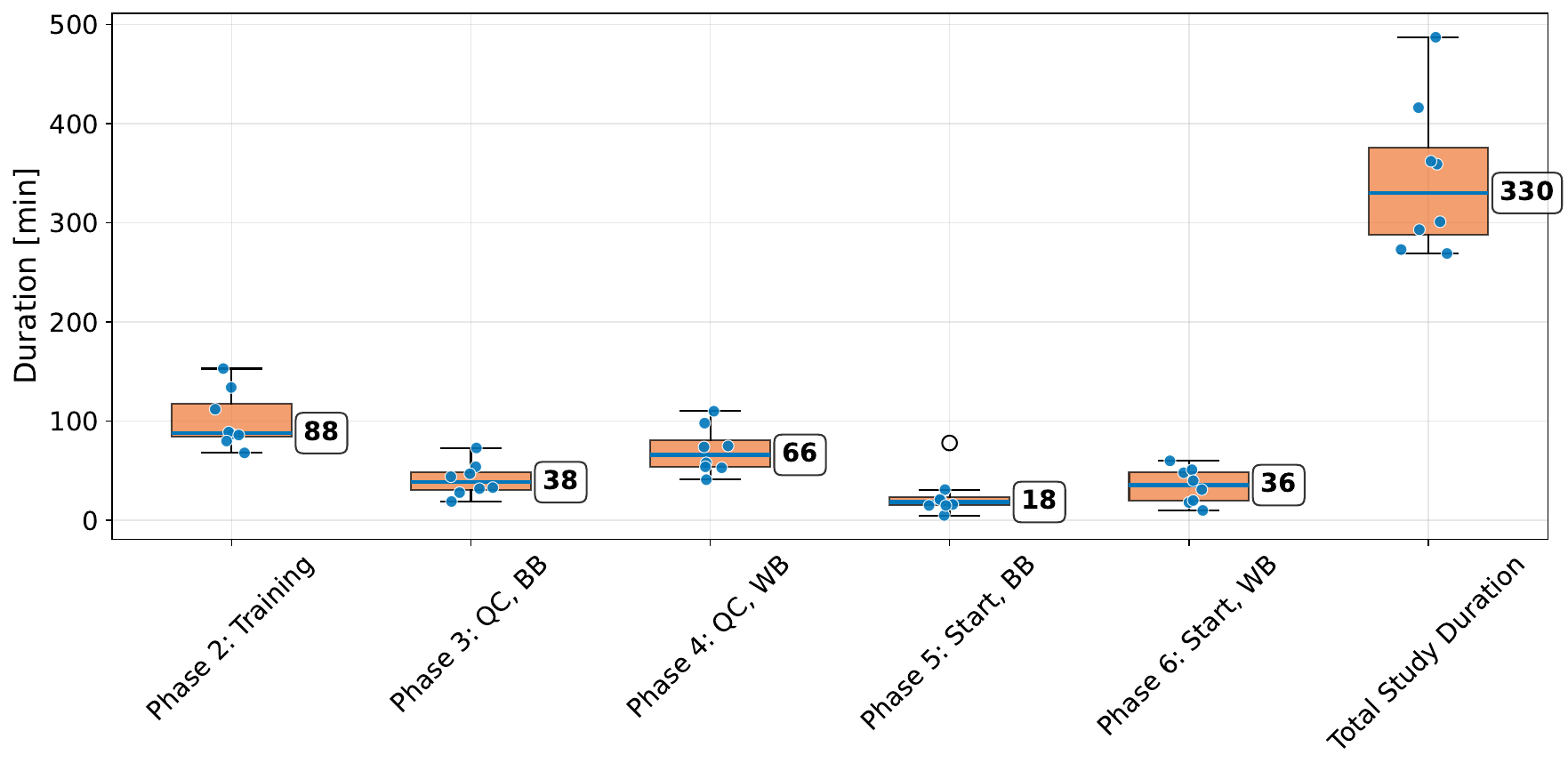}
\caption{\textbf{Time Demand} across user study phases. The boxplots illustrate the distribution of durations for each phase, with inlier points represented by blue circles and outliers by empty circles. The labels denote the median duration in minutes for each phase. Notably, AI support provided during the quality control (QC) phase results in scoring sessions that are approximately twice as long compared to when AI support is given right from the start.}
\label{fig:time-demand}
\end{figure}

To understand the practical implications of AI assistance on workflow efficiency, we examined the time demand associated with each phase of the user study. Figure~\ref{fig:time-demand} presents boxplots summarizing the duration of all relevant scoring phases across the cohort. The figure shows the median (red line), interquartile range (box), whiskers, and individual data points (blue circles), with outliers highlighted in red. Annotations mark the medians for each distribution.

Phases 1 and 8, covering the introductory video (approx.\ 13 minutes) and the closing questionnaire, are excluded from the plot. The final column, ``Total Study Duration,'' captures the cumulative time per participant, including these omitted phases as well as breaks and technical delays. While the overall study lasted around 5.5 hours per participant, individual differences are considerable, likely reflecting differences in user affinity with digital tools and adaptive behavior under fatigue.

The results reveal two prominent patterns: First, across both task settings (quality control (QC) and assistance from the start) white box (WB) AI support required notably more time than black box (BB) support. Specifically, the median durations of the WB phases (66 and 36 minutes) are approximately twice those of their BB counterparts (38 and 18 minutes, respectively). This increased time demand is consistent with the cognitive overhead of interpreting explanations and the study protocol's requirement that participants meaningfully engage with these explanations. Our observations indicated that participants increased their speed after becoming familiar with the initial explanations. Notably, the WB durations also show greater dispersion, indicating individual variation in speed. Some participants experienced similar time demands for both WB and BB support, indicating that efficiency can be achieved with increased familiarity.

Second, we observe that QC phases are generally more time-consuming than start-support phases, with QC sessions taking almost twice as long regardless of AI explainability. This makes sense given the sequential workflow: participants had to reassess the data in full, rather than make decisions immediately based on AI input. The increased duration thus stems from re-scanning the timeline, either by revisiting specific events flagged by the AI or conducting a comprehensive second pass through the data. Importantly, this QC overhead represents a conservative upper bound for time demand. In real-world deployments, practice and interface optimization would likely reduce this burden significantly.

One particularly revealing case merits discussion: a participant in the final phase (\textit{Start, BB}) completed the task in just 5 minutes, the shortest session across all participants and conditions. This user had developed enough trust in the AI to only evaluate AI-suggested events, effectively skipping segments without flagged arousals. This behavior illustrates the impact of AI design on user trust and reliance: the system, which prioritizes recall over precision, may have led users to over-trust its coverage. While encouraging from a workflow efficiency standpoint, this behavior underscores the importance of monitoring how AI shapes user attention and vigilance.

In summary, while white box AI introduces an initial time cost, its efficiency may converge toward black box levels as users gain fluency. Similarly, although QC demands more time than direct assistance, it may offer benefits in user autonomy and error-checking that justify this investment, especially during onboarding. These dynamics are crucial for deployment planning in time-sensitive environments such as clinical diagnostics or real-time monitoring.

\subsection{Evaluation of Questionnaires}\label{sec:eval-questionnaires}

\begin{figure}[htbp]
\centering
\includegraphics[width=\textwidth]{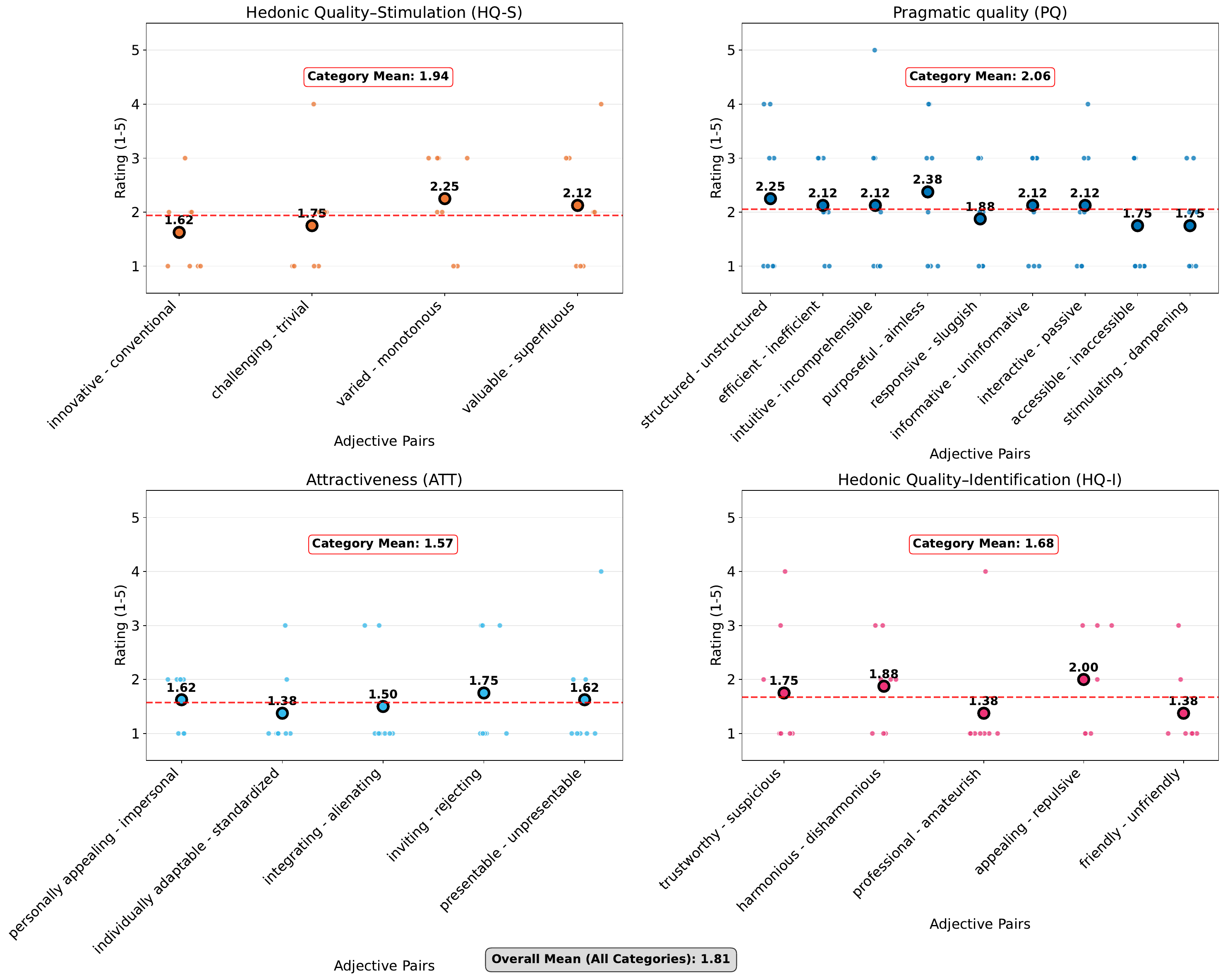}
\caption{\textbf{Evaluation of User Experience} utilizing the AttrakDiff questionnaire. This figure presents participant ratings across four dimensions: Hedonic Quality-Stimulation (HQ-S), Pragmatic Quality (PQ), Attractiveness (ATT), and Hedonic Quality-Identification (HQ-I). It emphasizes both the category means and the overall mean, with ratings on a 1-5 scale, where lower values indicate more favorable assessments. Participants rated all dimensions with a mean score of 2 or lower.}
\label{fig:attrakdiff-questionnaire}
\end{figure}

\begin{figure}[htbp]
\centering
\includegraphics[width=\textwidth]{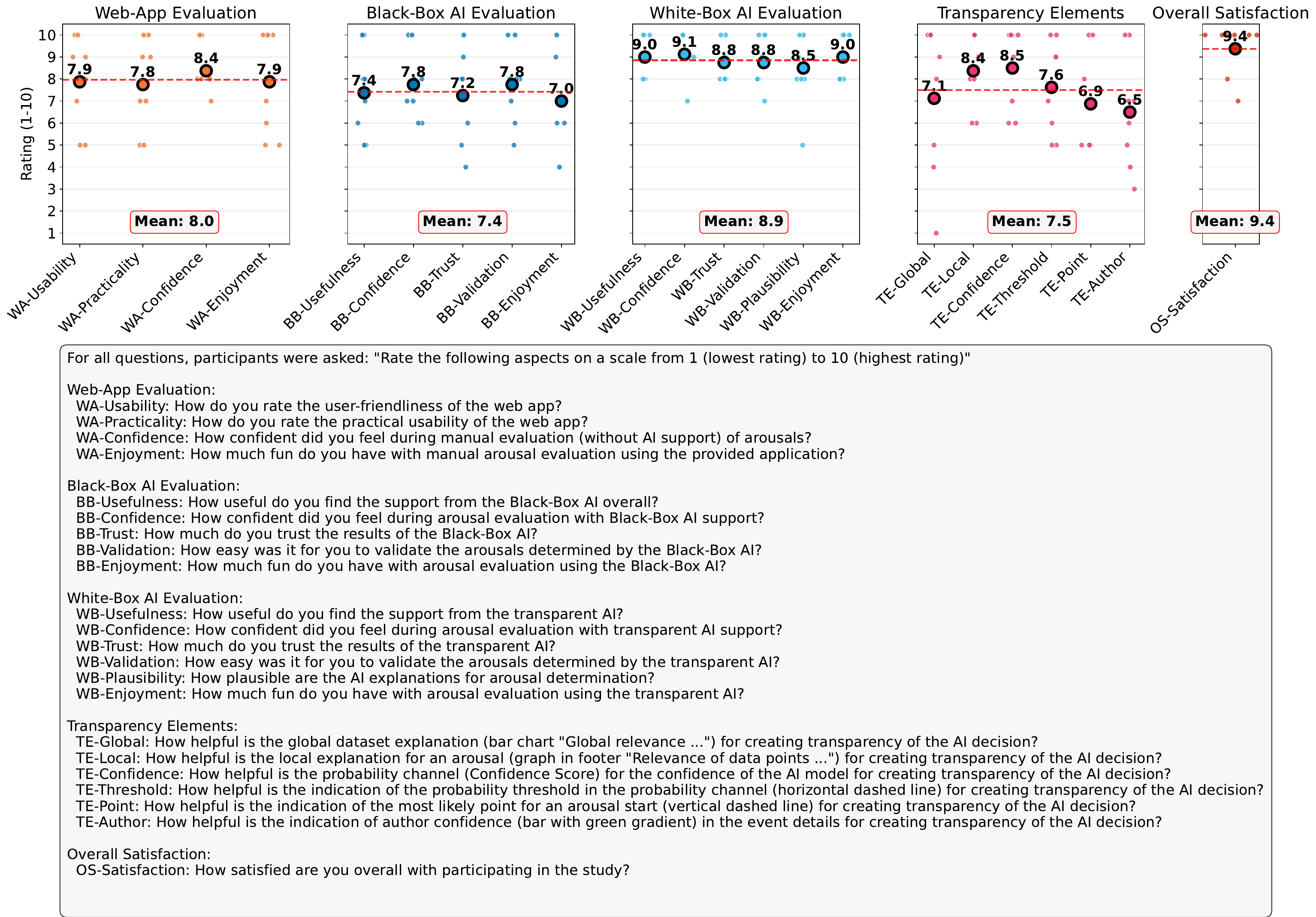}
\caption{\textbf{Analysis of Rating Questions} derived from participant feedback. This figure presents evaluations for the Web-App, Black Box, and White Box AI, as well as Transparency Elements and Overall Satisfaction, with the mean values indicated by a red line for each category. Ratings are on a 1-10 scale, where higher values denote greater favorability. Notably, ratings for the white box AI approach a score of 9, demonstrating a preference for transparency and explanatory features, as they are over 40\% closer to the ideal score of 10 compared to the black box AI ratings.}
\label{fig:rating-questions}
\end{figure}

\begin{figure}[htbp]
\centering
\includegraphics[width=\textwidth]{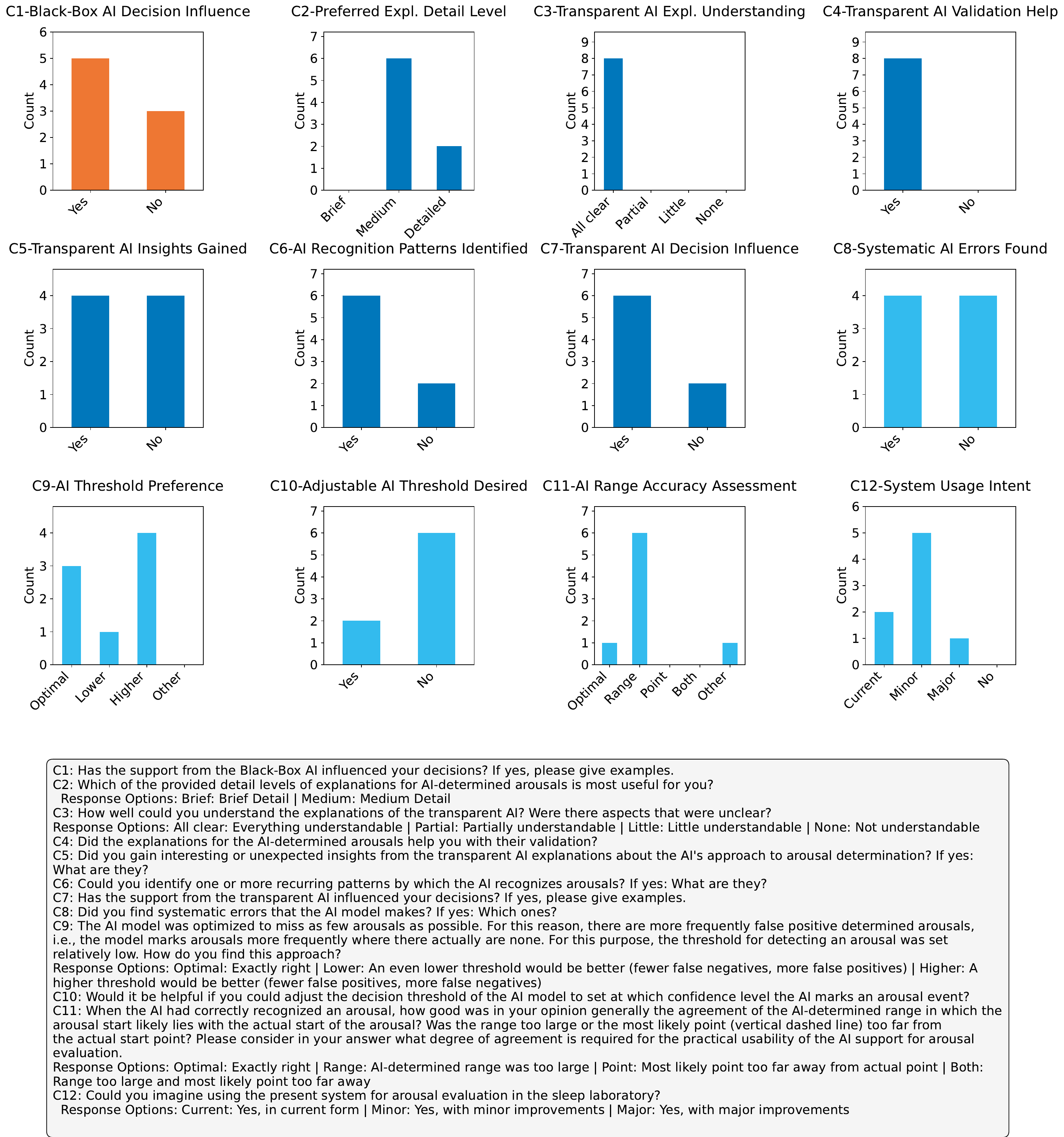}
\caption{\textbf{Analysis of Categorical Responses} derived from participant feedback. This figure depicts the distribution of categorical responses to various inquiries concerning black box (orange) and white box (dark blue) AI support, as well as different elements related to AI and the web application (light blue). Remarkably, all participants reported a complete understanding of the provided explanations, and unanimously found the transparent AI support beneficial for validating AI-suggested arousal events. Additionally, seven out of eight participants expressed a willingness to utilize the AI-powered Decision Support System (DSS) with minor modifications or in its current state, reflecting a high degree of acceptance and usability.}
\label{fig:categorical-responses}
\end{figure}

This section primarily delves into the secondary objectives \textit{SO1-SO9}, while also offering valuable insights into the primary objective \textit{PO3} (refer to Section~\ref{sec:objectives}). The numerical findings from the questionnaires are illustrated in Figures~\ref{fig:attrakdiff-questionnaire},~\ref{fig:rating-questions}, and~\ref{fig:categorical-responses}. Free-text excerpts have been slightly edited for fluency, with the original responses available in Appendix~\ref{appendix:free-text-responses}. All questions and responses have been translated from German to English.

\subsubsection{Usability and User Experience of the Web App}\label{sec:usability-and-user-experience-of-the-web-app}

Before delving into the AI support, we first examine the usability and user experience of the web app. It is important to note that the AttrakDiff ratings are on a scale from 1 to 5, where lower scores are preferable, while the rating questions use a scale from 1 to 10, where higher scores are more favorable. The web app's usability is commendable, as evidenced by the Pragmatic Quality ratings in the AttrakDiff questionnaire (Figure~\ref{fig:attrakdiff-questionnaire}), which show an average score of 2.06. Additionally, participants awarded the web app an explicit usability rating of 7.9 and a practicality score of 7.8 (Figure~\ref{fig:rating-questions}).

The user experience is also highly rated, as reflected in the excellent hedonic quality ratings, with an average stimulation score of 1.94 and an identification score of 1.68 on AttrakDiff, alongside an average attractiveness rating of 1.57 (Figure~\ref{fig:attrakdiff-questionnaire}). These findings align with the explicit ratings for web app confidence at 8.4 and web app enjoyment at 7.9 (Figure~\ref{fig:rating-questions}).

The positive user experience of the web app is further supported by participant feedback in the form of free-text responses. One participant remarked that the web interface felt \textit{very detailed and professionally created once one gets an overview}, while another simply stated: \textit{I find the web app user-friendly}.

In conclusion, the web app is perceived as ranging from good to very good by participants, which is a crucial prerequisite. This perception suggests that AI assistance is not hindered by poor overall usability and user experience of the web app.

\subsubsection{PO3 - AI Assistance Timing}\label{sec:po3-ai-assistance-timing}
In the free-text responses, participants were asked about their preference for starting with AI support or beginning with manual scoring followed by AI-based quality control. Six out of eight participants expressed a clear preference for initiating with AI support, with one participant highlighting \textit{increased confidence and speed} as the rationale. Another participant offered a more nuanced perspective, noting that \textit{AI support from the start is certainly easier - for experienced evaluators! (somewhat tempting for newcomers)}. This insight points to a potential risk for newcomers, who may become overly reliant on AI suggestions or find it challenging to critically evaluate them. While this study does not explore this aspect in depth, it remains a crucial consideration for introducing AI assistance in real-world applications.

\subsubsection{PO3 - AI Assistance Usefulness of Transparency}\label{sec:po3-ai-assistance-transparency}
Participants rated the usefulness of AI support on a \(1-10\) scale, with black box support receiving a score of \(7.4\) and transparent (white box) support achieving a higher score of \(9.0\) (Fig.~\ref{fig:rating-questions}). This preference for white box AI is echoed in free-text responses, where several participants expressed a clear preference for it, stating comments like \textit{White Box was better for me} or \textit{Transparent White Box AI support was my favorite}, while no such endorsements were made for black box AI.

Furthermore, a significant majority reported that AI, particularly the transparent variant, influenced their decisions (75\% for white box compared to 62.5\% for black box). When asked for examples to illustrate this influence, responses regarding black box AI assistance were mostly generic (e.g., \textit{Regarding confidence} or \textit{one or the other arousal}). In contrast, feedback on white box AI assistance included both general comments (e.g., \textit{occasionally}, or \textit{positively influenced}) and specific examples (e.g., \textit{PLM seen and I didn't but the AI was right.} or \textit{Breathing pattern /PLM}, where PLM stands for Periodic Leg Movement), indicating a deeper level of user engagement and insights gained with white box AI. These aspects are further discussed in the following sections.

\subsubsection{SO1 - Trust in AI Predictions}\label{sec:so1-trust-in-ai-predictions}

Trust scores mirror the utility pattern. Manual scoring confidence was already high (mean \(=8.4\)), but black box trust dropped slightly to \(7.3\). Transparent AI, by contrast, achieved a mean trust rating of \(8.8\) meaning that participants overall expressed slightly more trust in the transparent AI than in their own manual scoring. The AttrakDiff adjective pair ``trustworthy – suspicious'' gives the application a rating of 1.75 (on a \(1\text{–}5\) scale where lower values are positive) leaning strongly towards the first adjective, suggesting that the interface as a whole felt trustworthy (Fig.~\ref{fig:attrakdiff-questionnaire}). Taken together, the data indicate that explanations bolster user trust even beyond the baseline confidence derived from domain expertise. 
Several free-text notes connect confidence and trust directly to the presence of explanations, with a spectrum of reliance and trust from high to low: 
\begin{itemize}
    \item \textit{95\% correctly marked}
    \item \textit{The explanation partially convinced me to score an arousal}
    \item \textit{Explanations were the confirmation for me}
    \item \textit{In case of uncertainty explanations definitely helped and were often conclusive}
    \item \textit{Explanations helped for arousals where I wasn't quite sure}
    \item \textit{I thought about why the blackbox might have marked this now and with the whitebox I looked at why it marked it and then as always decided who was right}
    \item \textit{The explanations helped but were not decisive in my decisions}
\end{itemize}

We wish to emphasize that the statement \textit{95\% correctly marked} originates from participant \textit{HU-5}, who is identified as an outlier both at the event-based level (refer to Section~\ref{sec:human-ai-team-performance}) and in terms of total event count (refer to Section~\ref{sec:count_based_eval}). Both analyses also reflect sub-par performance for this participant. The combination of these quantitative results and the statement strongly indicates misplaced trust.

\subsubsection{SO2 - Perceived Comfort}\label{sec:so2-perceived-comfort}

Self-reported confidence while working with the AI aligns with the trust pattern: \(7.8\) for black box versus \(9.1\) for white box assistance. Overall study satisfaction was exceptionally high (mean \(=9.4\)), implying that the experimental workload and tooling were acceptable. Thus, transparent assistance not only inspired trust but may also have fostered a comfortable working atmosphere. This is backed by the good to very good ratings on Usability and User Experience of the Web App (Section~\ref{sec:usability-and-user-experience-of-the-web-app}).

\subsubsection{SO3 - Ease of Validation}\label{sec:so3-ease-of-validation}

Participants found black box predictions reasonably easy to validate (mean \(=7.8\)), yet transparent predictions were judged even easier (mean \(=8.8\)). All participants answered ``yes'' when asked whether explanations helped them in validation (C4 in Fig.~\ref{fig:categorical-responses}), underscoring the perception that explanations accelerated or simplified verification. This tendency is backed by the response of one participant, stating \textit{With the White Box AI, the evaluation went much faster because you can rely very much on the AI having marked everything important, unlike manual evaluation}. The participant was actually 1.5 times faster with WB AI support compared to BB AI support when assistance was provided from the beginning. However, during the QC phase, the participant was approximately twice as slow. Since the analysis in Section~\ref{sec:time-demand} shows that overall, phases with WB support took on average about twice as long as BB support phases, the experience of faster validation with WB AI seems to be subjective than backed by objective results.

Beyond the textual indicators of the value of explanations for validation discussed in \textit{S2} (Section~\ref{sec:so1-trust-in-ai-predictions}), a participant noted that \textit{explanations support and you take a closer look at it again}, and another participant who prefered transparent AI stated: \textit{One actually gets influenced and questions one's decisions again}, both indicating that explanations encourage a more thorough validation.

A participant highlighted a notable benefit of the local explanations for validation: \textit{Due to lack of space, additional channels 
could simply be viewed in the explanations, especially when the AI has 
already classified them as relevant, they are very interesting.} This observation emphasizes the utility of local relevance plots, which prioritize channels from most to least relevant, thereby reinforcing the overall value of the explanations.

\subsubsection{SO4 - Plausibility of Explanations}\label{sec:so4-plausibility-of-explanations}

Plausibility ratings for white box explanations averaged \(8.5\). Every participant declared the explanations completely understandable (C3 in Fig.~\ref{fig:categorical-responses}); no one selected ``partially'' or ``little'' understandable. Within individual transparency elements (see Fig.~\ref{fig:rating-questions}), local explanations (mean \(=8.4\)) and the confidence channel (mean \(=8.5\)) were deemed most helpful, whereas global relevance statistics, the most-likely event start indicator, and aggregate-confidence scores were less helpful (means \(=7.1\), 6.9, and \(6.5\), respectively). This pattern suggests that detailed, event-specific visuals resonate more strongly than aggregate information.
Concerning the transparency elements, one participant stated \textit{I missed the threshold/confidence} when being left with black box AI support.
In the free-text answers on how to improve the transparent AI support, one participant suggested that the AI model should prioritize EEG frequency increases. This would be in accordance with American Academy of Sleep Medicine guidelines, wherease the current modelling approach is guideline-agnostic. 
Another requested \textit{shorter markings}, hinting at interval-length issues discussed in \textit{SO9} (Section~\ref{sec:so9-temporal-accuracy-of-predictions}).

\subsubsection{SO5 - Satisfaction and Enjoyment}\label{sec:so5-satisfaction-and-enjoyment}

Enjoyment ratings again favour transparency: \(7.0\) under black box versus \(9.0\) under white box conditions. When combined with the high overall-satisfaction score, these findings imply that explainability may enhance intrinsic motivation, a non-trivial benefit given the repetitive nature of arousal scoring.

One scorer expressed that the AI system \textit{marked 95\% of the arousals correctly}, orally stating \textit{No other sleep system can match this}, capturing the enthusiasm behind the numbers.

\subsubsection{SO6 - Preferred Level of Detail}\label{sec:so6-preferred-level-of-detail}

Six of eight participants preferred a \emph{medium} level of explanatory detail, while the remaining two opted for \emph{detailed}. Nobody chose ``brief,'' indicating that terse justifications are insufficient, and very granular output is favored by some users. This suggests that concise but information-rich visuals strike the optimal balance, where the slight tendency towards the more detailed level of detail is consistent with the findings of \textit{SO4} (\ref{sec:so4-plausibility-of-explanations}). No free
text reported confusion about over-complexity.

\subsubsection{SO7 - Insights into AI Reasoning}\label{sec:so7-insights-into-ai-reasoning}

Half of the cohort reported gaining novel insights from the explanations, and 75\% could articulate recurring patterns with respect to various modalities that trigger arousal predictions: \textit{Pulse rate}, \textit{PLMs (don't always lead to an arousal) and desaturations with breathing events}, \textit{Heart rate increase}, \textit{Thorax and abdomen breathing, EMG}, \textit{EEG}, \textit{Breathing pattern /PLM}. Such meta-knowledge can foster calibrated trust, and may gradually shift human expertise toward more consistent scoring standards.
The free text responses to the question about insights gained from explanations about transparent AI arousal determination methods indicate that participants observed the AI's comprehensive use of all available data and noted a significant divergence in the data focus compared to human evaluators: \textit{The AI really uses all available data}, \textit{The data that the AI uses for assessment differs greatly from the data that an evaluator normally pays attention to}. 
One participant expressed surprise that the AI marked arousals when the patient was awake,
indicating potential areas for enhancing AI alignment with current scoring standards.

\subsubsection{SO8 - Decision-Threshold Preference}\label{sec:so8-decision-threshold-preference}

Only three participants considered the low decision threshold appropriate; four wanted it higher, one lower. Yet six of eight opposed a user-adjustable slider (C9 and C10 in Fig.~\ref{fig:categorical-responses}), suggesting that participants prefer a little less conservative default but are reluctant to fine-tune model sensitivity themselves. Future iterations could therefore expose two or three vetted presets rather than a free slider.

Free-text answers reveal that the reasons of those who favored a higher threshold are related to time-efficiency of the workflow: \textit{Too many false arousals. I get somewhat delayed by this}, \textit{I score an arousal rather quickly, rather than having to sight and then possibly decide}, and \textit{I'd rather draw a few ``forgotten'' arousals myself than having to delete many again.}.

\subsubsection{SO9 - Temporal Accuracy of Predictions}\label{sec:so9-temporal-accuracy-of-predictions}

Six of eight participants felt the predicted arousal range was too wide, while merely one person considered it \textit{exactly right}. No respondent complained that the most-likely point was systematically misplaced, pointing to interval width, not point accuracy, as the main usability bottleneck. 

Those, wishing for a smaller range, expressed either a size of \textit{$\pm$3 seconds} or a factor of \textit{0.3 to 0.5} of the current range.
Two participants clearly opted for only displaying the most likely point stating \textit{the line was exactly right}, and that an improvement would be a \textit{limitation of the marked area to the blue line}.

An obvious improvement would therefore be to only display the most likely point, and to keep the whole range through the confidence channel which was highly appreciated (see \textit{SO4}, Section~\ref{sec:so4-plausibility-of-explanations}).

\subsubsection{Conclusion of the Questionnaire Study}

The questionnaire study yields three overarching take-aways that extend and corroborate the quantitative performance results.

\textbf{1.~The platform is ready for AI integration.}
Participants rated both pragmatic and hedonic qualities of the web application as \emph{good to very good} (Section \ref{sec:usability-and-user-experience-of-the-web-app}),
removing basic usability as a confounder when interpreting the effects of algorithmic support.

\textbf{2.~Transparency is the dominant driver of acceptance.}
Across all key dimensions -- perceived usefulness and trust
(Sections~\ref{sec:po3-ai-assistance-transparency},~\ref{sec:so1-trust-in-ai-predictions}),
comfort (Section~\ref{sec:so2-perceived-comfort}), validation ease
(Section~\ref{sec:so3-ease-of-validation}), explanatory plausibility (Section~\ref{sec:so4-plausibility-of-explanations}), and enjoyment (Section~\ref{sec:so5-satisfaction-and-enjoyment}) -- the
white box assistant outperformed the black box by one to two Likert points.
Free-text notes confirm that explanations not only foster trust but also stimulate critical reflection rather than blind acceptance.

\textbf{3.~Clinically relevant refinements are still required.}
Users welcome AI support from the start (Section~\ref{sec:po3-ai-assistance-timing}), yet half of them request a \emph{higher} decision threshold and three-quarters deem the current onset window too wide (Section~\ref{sec:so8-decision-threshold-preference},
Section~\ref{sec:so9-temporal-accuracy-of-predictions}). They also ask for stronger emphasis on EEG features and prevent the AI from scoring arousals when the patient is awake to align the model with American Academy of Sleep Medicine guidelines (Sections~\ref{sec:so4-plausibility-of-explanations} and~\ref{sec:so7-insights-into-ai-reasoning}). These points can be addressed with an optimization closer to the F1 than the F2 score, a narrower or even point-based visual range focused on the most-likely start, having one EEG channel always prominently visible in all explanations, and (in case of events during wakefulness) more extensive post-processing of the AI's predictions.

\paragraph{Practical implication.} A transparent, event-centred explanation layer is not an optional add-on but a prerequisite for clinical adoption. With minor interface and model tweaks, the system could move from promising prototype to a trusted co-scorer in routine sleep-laboratory workflows. 
The majority of participants concur with this evaluation, as reflected in the results C12 in Fig.~\ref{fig:categorical-responses}. Specifically, two participants are ready to adopt the system in its current state, five participants suggest minor enhancements, and only one participant advocates for significant modifications.

\section{Discussion}\label{sec:discussion}

Our study addresses three fundamental research 
questions about human-AI collaboration in 
clinical arousal scoring (see Section~\ref{sec:introduction}), which we summarize as follows:

\begin{enumerate}[label=\textbf{RQ\arabic*:}, leftmargin=2.5em]
\item Is AI assistance preferable to unaided human scoring?
\item Does transparent (white-box, WB) assistance outperform opaque (black-box, BB) assistance?
\item Does it matter whether the AI is consulted from the start or only in a post-hoc quality-control (QC) pass?
\end{enumerate}

The evaluation of the effectiveness of AI assistance depends critically on the reference standard and evaluation measures employed. Our comprehensive analysis integrates event-level performance, count-based clinical measures, temporal efficiency, and user experience to provide insights into human-AI collaboration that would be missed by any single evaluation approach (Sections~\ref{sec:consensus-results} -- \ref{sec:eval-questionnaires}). This multi-faceted evaluation is particularly valuable for clinical decision support systems, where both technical performance and user acceptance are critical for successful implementation.

\subsection{The Critical Role of Evaluation Standards}

Our dual-ground-truth analysis shows that conclusions about AI assistance effectiveness, effectiveness, addressing \textbf{RQ1}, are contingent on the evaluation standard. Under the relatively neutral \emph{consensus} reference, the analyses (Section~\ref{sec:consensus-clusters}) reveal substantial inter-rater heterogeneity and clustering structure, indicating that the consensus aggregate is not closely aligned with the \emph{CPS} standard used to train the AI, even though that standard entered the pool via participant HU-9. In this setting, AI assistance does not confer measurable gains when judged against the consensus benchmark (Table~\ref{tab:combined-t-tests-consensus}).

By contrast, when evaluated against the \emph{CPS} ground truth, both AI and human-AI teams significantly outperform human solo scoring (Table~\ref{tab:combined-t-tests-original}). Concretely, the distribution of human solo F1 scores under CPS has mean $\mu{=}0.42$ and standard deviation $\sigma{=}0.09$, whereas human-AI teams improve to $\mu{=}0.56$ with reduced variability $\sigma{=}0.05$ (Figure~\ref{fig:performance_metrics_boxplot}; Section~\ref{sec:performance-metrics-original}). The pattern is consistent with AI assistance shifting human scoring toward the standard underlying model training. Given the typically poor inter-rater reliability in arousal scoring (in a recent study, a multi-center intraclass correlation of $\mathrm{ICC}{=}0.41$ was reported~\citealp{pitkanen2024multi}, indicating ``poor'' inter-rater reliability), such a shift is not unexpected and, importantly, is accompanied by a clinically desired reduction in variability.

Taken together, for \textbf{RQ1}, AI assistance for arousal detection is advantageous when the objective is alignment with a clearly specified clinical standard (here, CPS), whereas it offers no benefit under a neutral but heterogeneous consensus reference. These findings place heightened importance on the careful curation and governance of the training standard for AI intended for clinical decision support. Moreover, our evidence of multiple coexisting scoring tendencies among experts (Section~\ref{sec:consensus-clusters}) suggests that consolidating and harmonizing standards should precede broad deployment.

The count-based analysis adds the perspective of clinical decision-making (Section~\ref{sec:count_based_eval}; Figure~\ref{fig:combined_regime_analysis}). We observe systematic biases: the AI \emph{overestimates} arousals by roughly 10–40\% while humans \emph{underestimate} by 20–50\%. The outcome for the AI is expected since it was optimized for the F2 score. Human-AI teams reduce, but do not eliminate, this gap and magnitude of the reduction is regime-dependent. In the most favorable regime, transparent AI assistance as quality control, both variability and absolute error show a downward trend, as the coefficient of variation decreases from 0.47 (human solo) to 0.35 (team), and MAE drops from 34.5 to 20.8 events (Section~\ref{sec:count_based_eval}). Also, the median team discrepancy falls below 15\% of the ground-truth total (cf.\ Section~\ref{sec:count_based_eval}). While this pattern in the count-based evaluation mirrors the improvements observed in the event-level analysis, it does not reach statistical significance (Table~\ref{tab:count_team_solo}).

\subsection{Impact of Transparency on Performance and User Experience}

Findings on event-level performance (Section~\ref{sec:analysis-ai-assistance-regimes}) directly address \textbf{RQ2}: transparent assistance consistently outperforms opaque assistance. Statistically, white-box (WB) assistance yields an \emph{average} 18\% improvement in relative F1 over black-box (BB) assistance with a significant AI transparency$\times$Timing interaction (Table~\ref{tab:anova-contrast-summary}). The transparency benefit is most pronounced in quality control (QC), where WB exceeds BB by about 30\% on average (Table~\ref{tab:simple-effects-summary}).

Count-based evaluation shows that transparency provides modest, although non-significant additional benefits. Specifically, the QC-WB teams achieved the highest accuracy and lowest bias for underestimation of arousal counts, but the advantage of transparency over black-box assistance did not reach statistical significance at this sample size. Nonetheless, visual patterns suggest that transparency may help stabilize team performance and reduce inter-participant variability, particularly in QC workflows.

User-experience results align with these effects. Participants rate WB 1–2 Likert points higher across usefulness, confidence, trust, ease of validation, and enjoyment (Section~\ref{sec:eval-questionnaires}). On the 1–10 usefulness scale, WB narrows the gap to the ideal score of 10 by \emph{more than 40\%} relative to BB (Figure~\ref{fig:rating-questions}; Section~\ref{sec:po3-ai-assistance-transparency}). Together, these findings indicate that transparency enhances both performance and acceptance, key prerequisites for clinical adoption.

\subsection{Timing Effects and Workflow Strategy}

Our findings provide nuanced answers to \textbf{RQ3}. For event-level measures, although timing is not a significant main effect (Table~\ref{tab:anova-contrast-summary}), the significant AI$\times$Timing interaction indicates that transparency matters most at quality control (QC) timing (Table~\ref{tab:simple-effects-summary}).

For clinically oriented count-based measures, timing is the dominant factor (Section~\ref{sec:count_based_eval}): Compared to Start timing, QC regimes improve the AI-baseline Improvement Ratio by roughly a factor of four on average (the timing ratio is about 0.26, see Table~\ref{tab:count_rgt_contrasts}) and substantially reduce systematic under-counting (the timing effect on the percentage error $\mathrm{PE}$ is reduced by approximately $24$ percentage points, see Table~\ref{tab:count_bias_contrasts}).

Survey feedback on the preferred timing of AI assistance (see Section~\ref{sec:po3-ai-assistance-timing}) indicates a clear inclination among participants to utilize AI support from the outset of scoring, rather than reserving it for a subsequent quality control step. This early integration was seen as beneficial for workflow efficiency and user confidence, particularly among experienced scorers. Notably, one participant highlighted the need for caution, as immediate AI involvement could potentially foster over-reliance or diminish critical assessment, especially for less experienced users. 

Taken together, these findings highlight a key tension: while objective performance metrics point to the greatest accuracy gains when AI assistance is used as a quality-control step, many users express a preference for integrating AI support from the outset to enhance workflow efficiency and confidence. This underscores the challenge of designing systems that not only optimize measurable outcomes but also align with user preferences and promote sustained engagement.

\subsection{Clinical Implications, Time Efficiency, and Workflow Integration}

Combining transparency with a QC timing yields the most favorable outcomes across measures. In QC-WB, teams achieve the highest count accuracy (Figure~\ref{fig:combined_regime_analysis}B), the strongest improvement over the AI baseline (Figure~\ref{fig:combined_regime_analysis}C), and the smallest residual bias (Figure~\ref{fig:combined_regime_analysis}D). This regime also shows reduced inter-participant variability (Section~\ref{sec:count_based_eval}).

These gains however come with time costs. Median duration approximately doubles for WB versus BB and, independently, QC sessions are roughly twice as long as Start sessions (Figure~\ref{fig:time-demand}; Section~\ref{sec:time-demand}). Some participants, however, already achieved WB speeds comparable to the fastest BB (Section~\ref{sec:time-demand}), suggesting learning effects may mitigate transparency overhead. Furthermore, one participant explicitly expressed a preference for white-box over black-box due to enhanced speed, a preference substantiated by the data for this participant.
By contrast, QC's extra pass is inherently costly and unlikely to diminish substantially with experience.

\paragraph{White-box vs.\ black-box.}
Given the large performance and acceptance advantages of WB over BB (Sections~\ref{sec:analysis-ai-assistance-regimes},~\ref{sec:eval-questionnaires}), we judge the additional time to be justifiable. Several participants also appeared to consult explanations selectively. With familiarity, brief verification via explanations may be even faster than speculating about a black-box suggestion.

\paragraph{Quality control vs.\ start-based.}
Six of eight participants preferred starting with AI (Section~\ref{sec:po3-ai-assistance-timing}), citing confidence and speed, but QC delivered the most accurate counts, the largest error reduction relative to AI, and the smallest bias (Section~\ref{sec:count_based_eval}). A practical compromise is therefore context-dependent: For routine scoring, a faster yet less accurate support from the start may suffice. Conversely, for auditing, educational purposes, or complex cases, the more time-consuming quality control approach may prove beneficial.

\subsection{Human-AI Collaboration: Beyond Simple Automation}

The benefit-ratio analysis indicates teams adopted about 46\% of the AI’s alignment signal on the median, with wide individual spread (0.10–0.78; Figure~\ref{fig:performance_metrics_boxplot}, Section~\ref{sec:performance-metrics-original}). While this may indicate a level of critical evaluation which is beneficial, it may also include missed opportunities for improvement. 
Notably, Sections~\ref{sec:human-ai-team-performance} and~\ref{sec:count_based_eval} document pitfalls of strong alignment and over-reliance (e.g., scorer HU-5), including acceptance of false positives and insufficient correction. 
Despite such cases, and the fact that teams typically did not \emph{exceed} the solo AI at the event level (Table~\ref{tab:performance-comparison-regimes}), there are strong reasons to retain humans in the loop:
QC-WB teams frequently surpass the AI on clinically salient count-based outcomes (Figure~\ref{fig:combined_regime_analysis}C), humans cope better with out-of-distribution cases, and human oversight supports accountability, bias mitigation, and iterative model improvement.

In summary, transparent AI assistance, particularly when applied as a quality-control step, empowers human experts to harness the strengths of artificial intelligence while actively mitigating its errors. This collaborative approach produces results that are not only more consistent than those achieved by unaided humans, but also more reliable and trustworthy than those generated by AI alone.

\subsection{Limitations}\label{sec:limitations}

Several limitations warrant consideration when interpreting our findings.

The study's reliance on eight participants, while adequate for detecting large effects, limits the ability to identify more subtle interactions and may introduce individual biases that affect generalizability. The small sample size also constrains our ability to detect smaller but potentially meaningful effects, particularly in the interaction analyses.

The dual-ground-truth approach, while methodologically rigorous, introduces its own limitations. The consensus labels were generated by the same annotators involved in the study, potentially introducing residual bias. The CPS ground truth, while representing established clinical practice, reflects individual scoring styles that may not generalize across institutions or time periods. Future research should consider involving independent expert panels to enhance ground truth validity.

The AI model's optimization for high recall (F2 score) may have influenced human adoption patterns and interaction behaviors. A model tuned for precision might yield different collaboration dynamics, suggesting the need for further exploration of alternative optimization strategies and their impact on human-AI interaction.

The study utilized a single explanation type (causal explanations in the form of local relevance plots together with confidence traces), limiting our understanding of how different explanation modalities might affect collaboration. Investigating other types of explanations, such as contrastive or counterfactual explanations, could provide additional insights into optimal explanation design for clinical applications.

Although the study design incorporates counterbalancing and different subjects per phase to mitigate learning effects, some learning curve may still be present. The limited exposure duration (primarily first-time usage) means that long-term effects on user trust, fatigue, and adaptation remain unexplored.

Finally, the user interface design significantly influences collaboration effectiveness, and our specific implementation choices may have affected the results. The iterative design process incorporated expert feedback, but alternative interface designs could yield different interaction patterns and performance outcomes.

\subsection{Conclusion}\label{sec:final-conclusion}

The central lesson of this study is that 
human-AI collaboration in arousal detection significantly enhances team alignment with the reference standard used to train the AI: teams become more accurate and more consistent, and the key lever is \emph{transparency}. Explanations transform the AI from a mere source of suggestions into a provider of actionable evidence, yielding the most reliable results when transparent support is applied as a targeted \emph{quality-control} step.

Beyond quantitative performance, clinicians express a strong willingness to adopt such systems. Transparent assistance receives higher ratings for usefulness, trust, ease of validation, and enjoyment, with most participants indicating they would use the system with at most minor modifications. However, these benefits come at the cost of increased time, as both transparency and quality control introduce workflow overhead. This underscores the need for \emph{configurable workflows}: transparent AI assistance for quality support should be employed when accuracy is critical, while start-time assistance can address throughput demands, provided safeguards are in place to prevent over-reliance.

Collectively, these findings position human-AI arousal scoring as a \emph{design} challenge as much as a modeling one. Systems that integrate clinically meaningful explanations, flexible timing, calibrated decision thresholds, and precise onset visualization can facilitate alignment with the chosen standard while maintaining human agency. In this way, transparent AI functions as a reliable co-scorer, supporting rather than supplanting expert judgment.

\bibliographystyle{plainnat}
\bibliography{sample}

\newpage
\begin{appendix}

\section{Table of Notation}\label{appendix:table-of-notation}

Table \ref{tab:table_of_notation} lists the notation used in the main text.

\begin{table}[!htbp]
\caption{Table of notation}
\label{tab:table_of_notation}
\setlength{\tabcolsep}{4pt}
\begin{center}
\begin{tabular}{ll}

\toprule
Symbol & Meaning \\
\cmidrule(lr){1-2}

$t_{k,i}$ & Start time of event $i$ annotated by annotator $k$ \\
$C_j$ & Cluster $j$ of temporally close annotations \\
$\varepsilon$ & Temporal distance threshold for clustering \\
$S$ & Kneedle sensitivity parameter for knee detection \\
$m^{\text{Cluster}}$ & Minimum annotations required to form a cluster \\
$A_{k,j}$ & Indicator that annotator $k$ contributed to cluster $j$ \\
$s_k$ & Sensitivity (true positive rate) of annotator $k$ \\
$p_k$ & Specificity (true negative rate) of annotator $k$ \\
$P_j$ & Probability that cluster $j$ contains a true event \\
$P_j^{\text{prev}}$ & Prior event probability for cluster $j$ from previous iteration \\
$\log(L_j)$ & Log-likelihood under event-present hypothesis for cluster $j$ \\
$\log(M_j)$ & Log-likelihood under event-absent hypothesis for cluster $j$ \\
$D_{1j}$ & Unnormalized log-posterior for event present in cluster $j$ \\
$D_{2j}$ & Unnormalized log-posterior for event absent in cluster $j$ \\
$\tau$ & Threshold for selecting consensus events ($P_j \geq \tau$) \\
$F1$ & F1 score (harmonic mean of precision and recall) \\
$F2$ & F2 score (recall-weighted F-score) \\
$TP,\ FP,\ FN$ & True positives, false positives, false negatives \\
$\mathrm{HU}$ & Human solo\\
$\mathrm{AI}$ & AI solo\\
$\mathrm{HU+AI}$ & Human-AI collaboration\\
$\mathcal{B}$ & Benefit ratio: $\frac{F1^{\text{HU+AI}}-F1^{\text{HU}}}{F1^{\text{AI}}-F1^{\text{HU}}}$ \\
$\mathcal{R}$ & Relative F1 score: $\frac{F1_{\text{HU+AI}}}{F1_{\text{AI}}}$ \\
$C_{\mathrm{GT}}$ & Ground-truth arousal count per recording \\
$C_x$ & Arousal count of source $x\in\{\mathrm{AI},\,\mathrm{HU{+}AI},\,\mathrm{HU}\}$ \\
$D_{x\to\mathrm{GT}}$ & Absolute deviation $\lvert C_x - C_{\mathrm{GT}}\rvert$ \\
$A_{x,\mathrm{GT}}$ & Count accuracy $\frac{1}{1+\mathrm{APE}_{x,\mathrm{GT}}}$ \\
$\mathrm{APE}_{x,\mathrm{GT}}$ & Absolute percentage error $\frac{D_{x\to\mathrm{GT}}}{\max\{C_{\mathrm{GT}},\epsilon\}}$ \\
$R_{\mathrm{GT}}$ & AI-baseline improvement ratio \\
$y_{\mathrm{RGT}}$ & Log improvement ratio $\log R_{\mathrm{GT}}$ \\
$\mathrm{PE}$ & Percentage error $\frac{C_{\mathrm{HU+AI}}-C_{\mathrm{GT}}}{\max\{C_{\mathrm{GT}},\epsilon\}}$ \\

\bottomrule
\end{tabular}
\end{center}
\end{table}

\section{Comparison of Feature Attribution Methods}\label{appendix:comparison-of-feature-attribution-methods}

This appendix provides an analysis of post-hoc feature attribution methods applied to a randomly selected arousal event from subject S2 (selected within the white-box quality-control regime; cf. Table~\ref{tab:sample-selection}). Our goals are to: 
(i) qualitatively compare local attributions across input channels and methods, 
(ii) quantify cross-method agreement using multiple similarity measures, and 
(iii) relate local agreement to global channel importance rankings derived from aggregated attributions.

\paragraph{Methods in Brief}
We employ two gradient-based explainability methods to generate attributions for our neural time-series model.
DeepLIFT is our primary method for the user study (see Section~\ref{sec:explanation-methods}). It attributes relevance by contrasting activations to a reference activation in a computationally efficient manner.
GradientSHAP, used here as a complementary method, builds on the SHAP framework by using gradients and random perturbations to estimate feature attributions~\citep
{lundberg2017unified}. It approximates SHAP values by averaging gradients taken along random paths from reference inputs (baselines) to the target input. In each of many trials it (i) slightly perturbs the input, (ii) randomly selects a baseline, (iii)
picks a random point on the straight line between that baseline and the (noisy) input, and (iv) computes the target output's gradient there. 
Averaging over trials yields SHAP values that are essentially the expected gradient weighted by the input -- baseline differences.
We use the standard implementation from the \texttt{captum} library and use the same baselines as for DeepLIFT (see Section~\ref{sec:local-explanations-design}).

\begin{figure}[!htbp]
    \centering
    \includegraphics[width=1.0\linewidth]{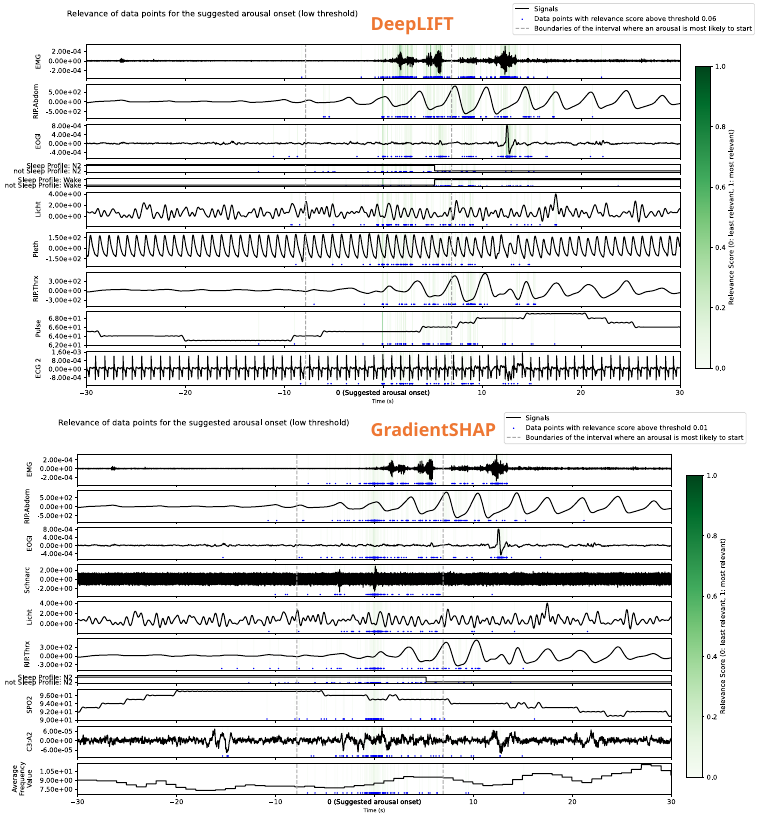}
    \caption{\textbf{Local feature attributions (DeepLIFT vs. GradientSHAP).} Both methods highlight consistent event-proximal relevance spikes in several respiratory and EMG-derived channels, while also revealing method-specific differences in sparsity and spread of relevance.}
    \label{fig:comparison-of-local-attribution-plots}
\end{figure}

\paragraph{Local Attributions}
Figure~\ref{fig:comparison-of-local-attribution-plots} displays the per-channel attributions for DeepLIFT and GradientSHAP for a 60-second window centered on the model’s predicted event onset (dashed gray line). 

This matches the user-study visualization with \textit{low} threshold settings which reveal the most fine-grained local contributions (see Section~\ref{sec:local-explanations-design}). Importantly, these thresholds influence the computation of global channel relevance but not the local cross-method similarity analysis, which uses all available attributions.

\begin{figure}[!htbp]
\centering
\includegraphics[width=1.0\linewidth]{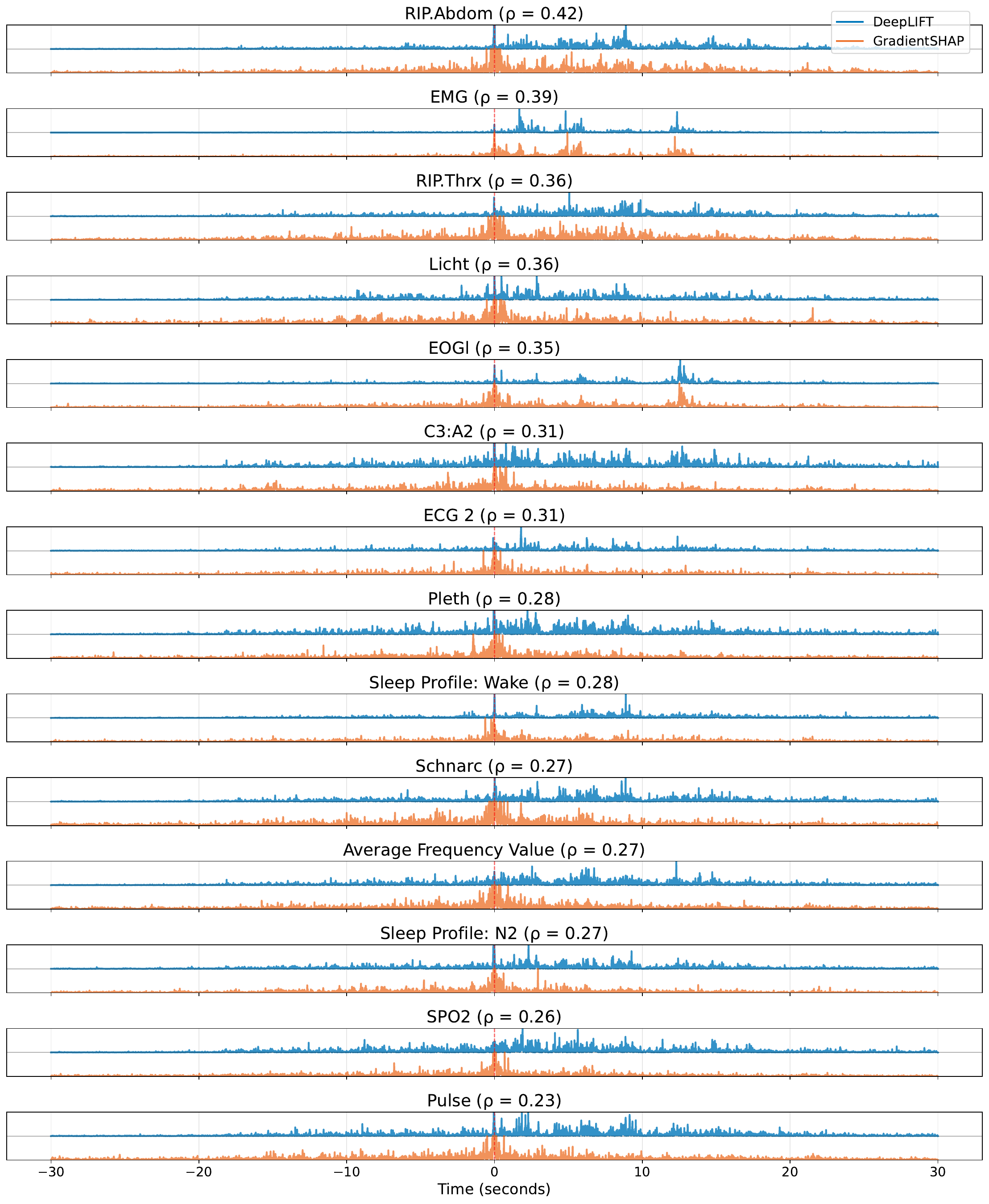}
\caption{\textbf{Side-by-side temporal comparison of normalized attributions (DeepLIFT vs. GradientSHAP).} Channels are ordered by cross-method Spearman correlation $\rho$. Normalization excludes $\pm$\,0.5\,s around the predicted onset to avoid peak inflation. The correlations are computed on the full signals, not the masked signals used for normalization.}
\label{fig:normalized-attribution-comparison}
\end{figure}

To visually compare temporal attributions directly and prepare for the correlation analysis, Figure~\ref{fig:normalized-attribution-comparison} overlays normalized DeepLIFT and GradientSHAP traces for the same channels ordered by Spearman correlation over the normalized signals. Normalization is performed per channel using min-max scaling with an exclusion window of $\pm 0.5\,$s around the prediction point to prevent spike domination at $t=0$. This window is \emph{not} removed from the signals used for the correlation analysis, it only affects the normalization parameters.

The ordering by $\rho$ reflects which channels display most similar attribution patterns (RIP.Abdom, EMG). Visually, high-$\rho$ channels show aligned bursts across methods, while lower-$\rho$ channels exhibit timing or spread differences, indicating method-specific sensitivities.
Based on Spearman $\rho$, the correlation for most channels is weak, while one channel (RIP.Abdom) reaches moderate correlation.

\begin{figure}[!htbp]
\centering
\includegraphics[width=1.0\linewidth]{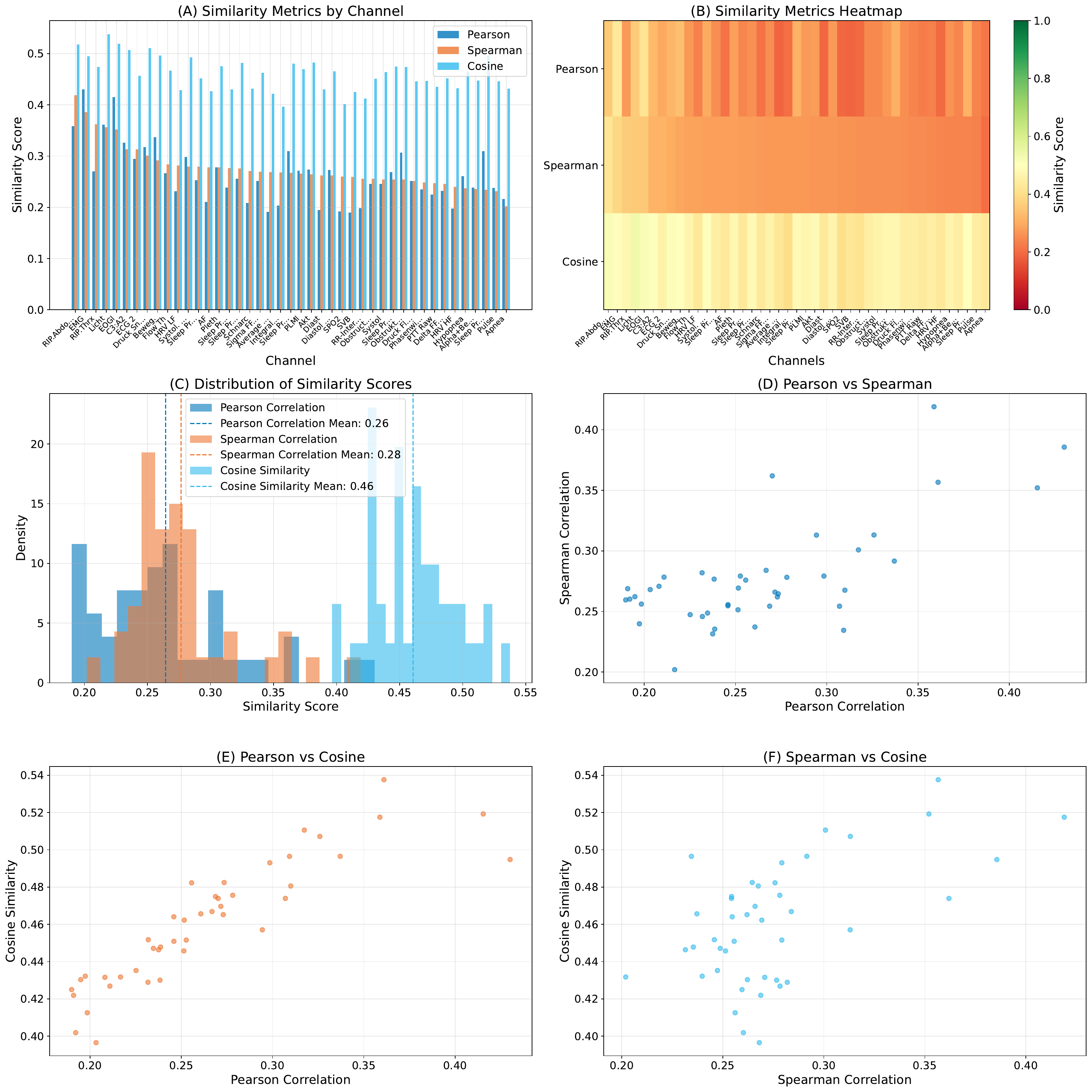}
\caption{\textbf{Similarity between methods across channels for three different correlation measures.} Plots (A) and (B) show corresponding channel-wise similarity scores ordered by Spearman correlation. Pearson correlation and cosine similarity tend to co-vary, while Spearman correlation differs markedly, emphasizing monotone agreement in temporal structure.}
\label{fig:similarity-metrics-comprehensive}
\end{figure}

Before proceeding, we test the robustness of our approach by comparing alternative correlation measures for quantifying agreement between the two attribution methods.
Figure~\ref{fig:similarity-metrics-comprehensive} summarizes agreement between the two attribution methods across channels. We report Spearman rank correlation (monotonic agreement), Pearson correlation (linear agreement), and cosine similarity (directional alignment).

Panels A, B and E reveal that Pearson correlation and cosine similarity typically rank channels similarly, reflecting shared sensitivity to overall shape and directional alignment. Spearman correlation, by focusing on order rather than magnitude, penalizes localized discrepancies (e.g., minor shifts in peaks or differing spike widths). Because our interest is whether methods agree on \textit{when} a channel is relevant (focusing on the temporal structure of the attributions) rather than exact amplitude scaling, we use \textit{Spearman correlation} as the primary agreement measure. This choice is further supported by Figure~\ref{fig:normalized-attribution-comparison}, where channels with high $\rho$ consistently exhibit similar patterns across methods.

\paragraph{Global Attributions}\label{sec:global-attributions}
We next relate local agreement to global feature importance. We compute global relevance by aggregating the positive attributions for each channel over time at \textit{low} threshold selection.

\begin{figure}[!htbp]
    \centering
    \includegraphics[width=1.0\linewidth]{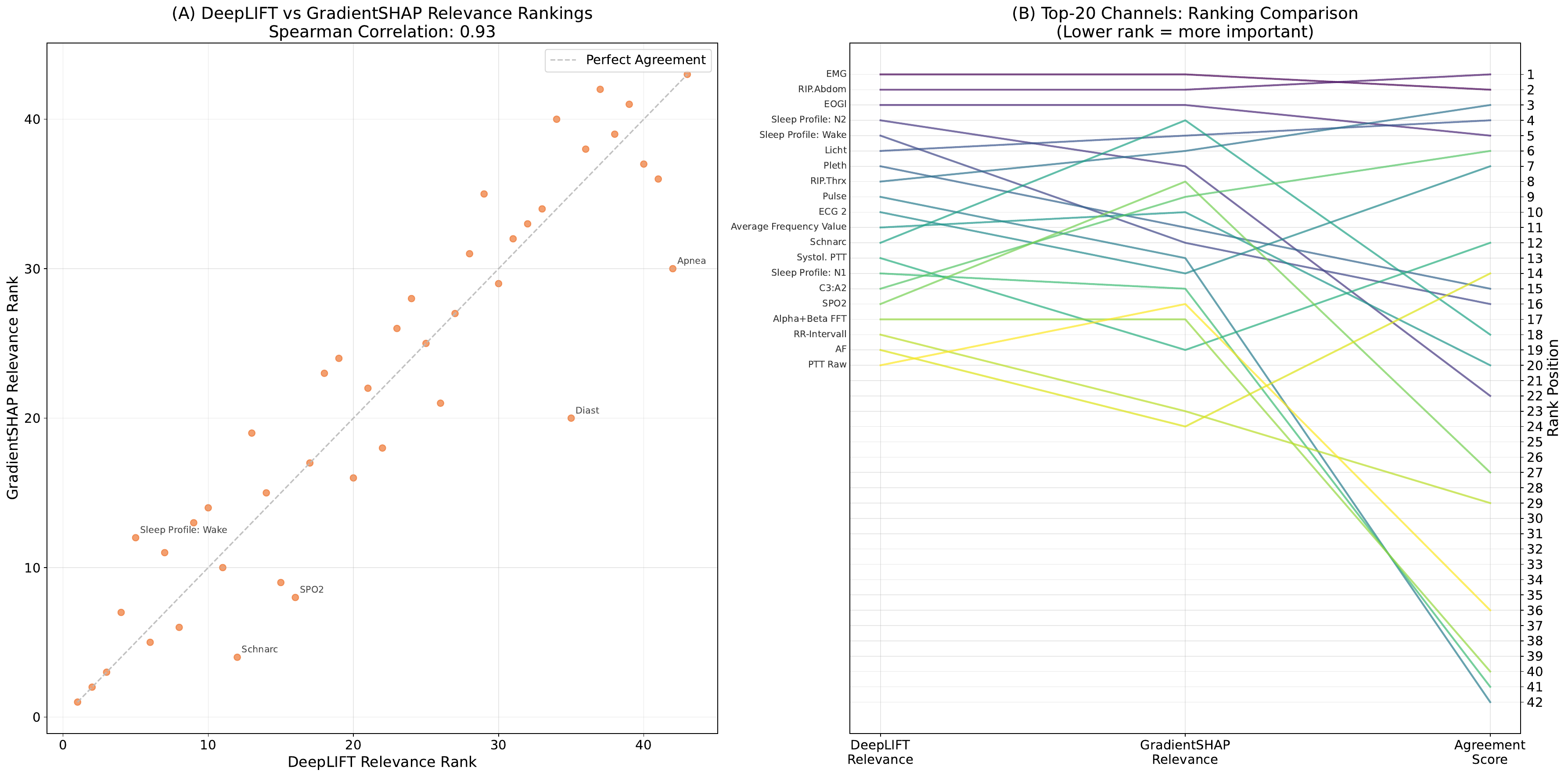}
    \caption{\textbf{Global relevance and cross-method correspondence.} Panel A shows a very strong rank correlation between DeepLIFT and GradientSHAP global relevance across channels. Panel B depicts a slope graph for the top-20 channels by DeepLIFT relevance: both methods agree on the leading channels, while a few channels show notable divergence or high global relevance but comparatively low local agreement.}
    \label{fig:combined-attribution-ranking}
\end{figure}

Figure~\ref{fig:combined-attribution-ranking} combines two complementary views. Panel A shows the correlation of global relevance rankings between DeepLIFT and GradientSHAP across all channels. Panel B displays a slope graph for the top-15 channels (by DeepLIFT relevance), comparing their DeepLIFT relevance rank, GradientSHAP relevance rank, and agreement-score (Spearman) rank.

Panel A demonstrates a \emph{very strong} correlation between global relevance rankings from both methods, indicating robust agreement on which channels matter most overall. Panel B shows that (i) the top-three channels are consistently ranked highly by both approaches, and (ii) a small number of channels (e.g., ``Schnarc'') deviate substantially between methods, or are globally important but show relatively low local attribution agreement (e.g., ``Pulse''). This pattern shows that, even when global importance aligns, temporal attribution structure can still differ.

\paragraph{Discussion}
Overall, DeepLIFT and GradientSHAP provide a largely coherent view of feature attribution patterns in this exploratory analysis. 
Notably, both methods identify the same three channels (EMG, RIP.Abdom, and EOGl; see Figures~\ref{fig:comparison-of-local-attribution-plots} and~\ref{fig:combined-attribution-ranking}B) as most important globally. These channels also demonstrate the highest level of cross-method agreement, as measured by Spearman correlation. 
Although the agreement scores for these channels are only moderate or on the borderline of moderate, visual inspection (Figure~\ref{fig:combined-attribution-ranking}B) reveals a clear and meaningful alignment in their overall temporal attribution patterns.
Conversely, channels deemed less important by both methods (e.g., Pulse or Average Frequency Value) tend to show lower agreement scores and more pronounced visual discrepancies in attribution patterns (see Figure~\ref{fig:normalized-attribution-comparison}). In fact, the time-series for Pulse and Average Frequency Value do not display any distinctive features from a layman's perspective in contrast to e.g. EMG, RIP.Abdom, or EOGl (see Figure~\ref{fig:comparison-of-local-attribution-plots}), which may help explain the lack of consistent attribution.
Importantly, the global rank correlation across all channels remains very strong ($\rho=0.93$), underscoring robust overall concordance between the two attribution methods.
Of particular note is the ``Pulse'' channel, which demonstrates an intriguing discrepancy: while it ranks as the second most globally relevant channel according to DeepLIFT across the entire dataset (see Figure~\ref{fig:global-explanation}), it shows the lowest consistency in its attribution pattern when compared to GradientSHAP (see Figure~\ref{fig:combined-attribution-ranking}B). Determining whether this inconsistency is merely an anomaly specific to the event examined here, or indicative of a broader phenomenon, would require more extensive analysis across a larger set of events. Such an investigation, as well as the evaluation of additional post-hoc attribution methods, remains a valuable direction for future work but lies beyond the scope of this study.

Collectively, these findings support the robustness of our choice of \textit{DeepLIFT} for the user study: It delivers clear, interpretable local explanations and demonstrates good concordance with GradientSHAP, showing clear visually consistent patterns of agreement for the most important channels at the local level and excellent consistency in global relevance across channels.

\section{Mapping of Subject IDs}\label{appendix:mapping-of-subject-ids}
Table \ref{tab:mapping-of-subject-ids} shows the mapping of subject IDs, as used in Table~\ref{tab:sample-selection}, to file IDs as specified in the documentation of the CPS dataset~\citep{kraft2024cps}.

\begin{table}[htbp]
\caption{\textbf{Mapping of Subject IDs to File IDs}. Subject IDs from Table~\ref{tab:sample-selection} are mapped to the corresponding file IDs as specified in the CPS dataset documentation~\citep{kraft2024cps}.}\label{tab:mapping-of-subject-ids}
\begin{center}
\begin{tabular}{ll}
\toprule
Subject-ID & File-ID \\
\midrule
S1 & FLsgQZoIGHx1G3LmdD7jtICMik2EKRKN \\
S2 & Su02hndUSYGSKJmcSqroKmtDjXIJ4y60 \\
S3 & kKDzUlAprXqDz84Nrw9UP1W0jpgUKkhN \\
S4 & YFQX33c8EEoapTndd2084KbUuUmtj7xF \\
S5 & LcsapTberZwzU7qyEr11andO59HTOVCv \\
S6 & tIgyhF8T1BOZnu7h6jb58igU5MAGgdo9 \\
S7 & CzwqE37s81YahjNICSXI2Tb4Fmp6bclE \\
S8 & vHJMSYFIl1TfLweQ5DWMGN5f47ULFNxe \\
S9 & leySrSnra9yAO3eTJIGB55nrjRS3RqIW \\
S10 & KB84bUmLWOrKCKkISCn8QuNBhF5mg0L8 \\
S11 & MT1zW5iB0h1bxF42QBpyDqotQk7NcnHw \\
S12 & RpARZ17l5osnFUcqIj2aTOsgRBMDutoA \\
S13 & mXHZZ887A9fcZgOmnxhnPVHwu5ECljDG \\
S14 & oEJ7fslCTL7s0OfIe7nYIPqo7Il4rMjI \\
\bottomrule
\end{tabular}
\end{center}
\end{table}

\section{Construction of Consensus Ground-Truth}\label{appendix:consensus-ground-truth}

\begin{figure}[!htbp]
    \centering
    \includegraphics[width=1.0\linewidth]{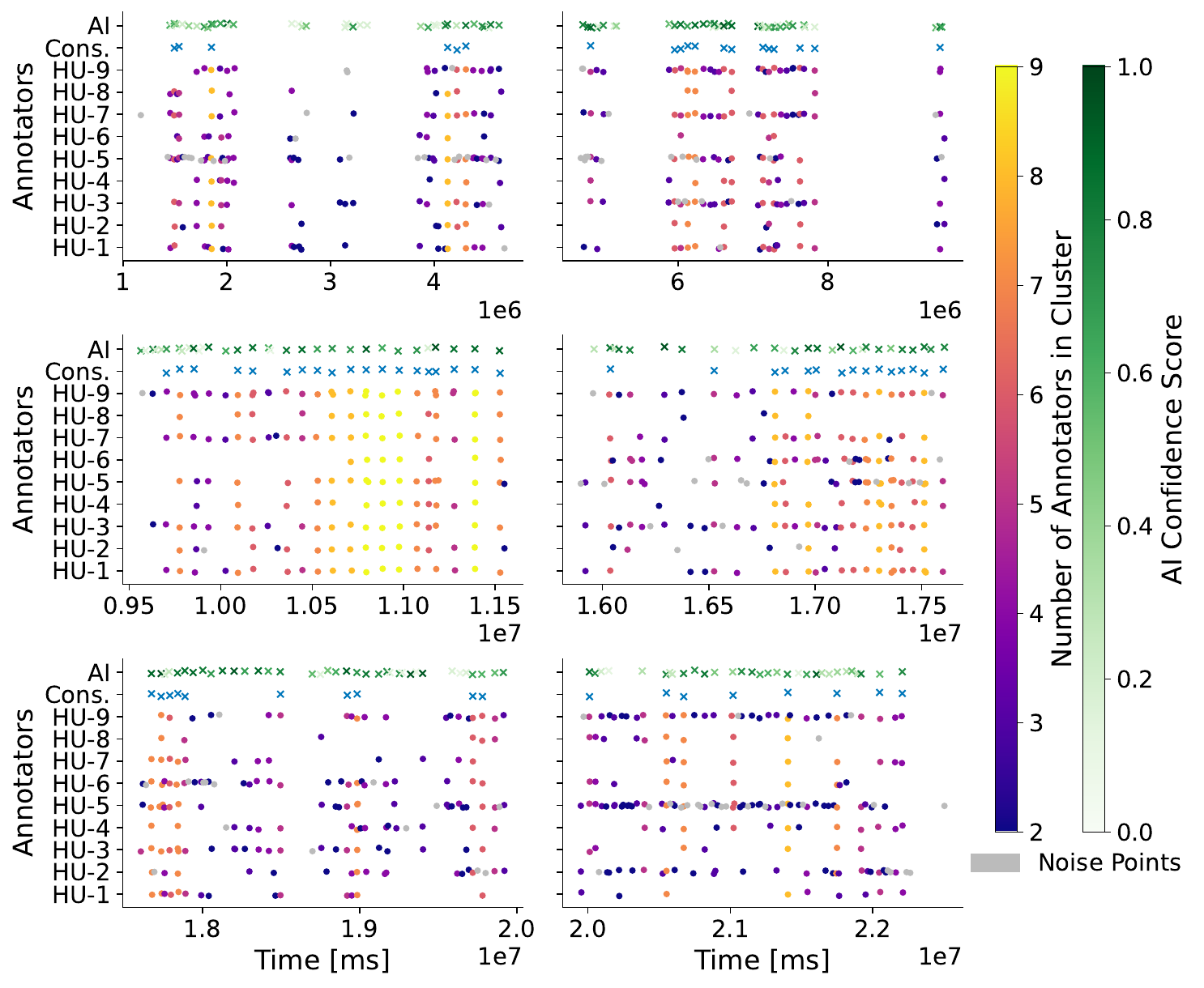}
    \caption{\textbf{Faceted Timeline Plot of Clustering Results using Agglomerative Clustering}. The plot displays human solo annotations across multiple panels, with data from patient samples \textit{S2} and \textit{S4} concatenated into approximately six hours of recording time. Annotations of the same color and vertical alignment belong to the same cluster. The horizontal displacement is for visual clarity only. Large intervals without annotations have been omitted. Consensus annotations (\textit{Cons.}) are marked as blue crosses, while AI annotations are marked as green crosses, with the hue representing the AI model's confidence score. The color scale on the right indicates the number of annotators in each cluster. Annotator \textit{HU-9} represents the human annotations used as the CPS ground truth for AI model training.}
    \label{fig:faceted-timeline-plot-consensus-agglomerative}
\end{figure}

Figure \ref{fig:faceted-timeline-plot-consensus-agglomerative} shows the human annotations over multiple time ranges for the two manually annotated samples S2 and S4 (see Table~\ref{tab:sample-selection}). The annotations are clustered using Agglomerative Clustering (see Section~\ref{sec:data-preparation}). The consensus annotations are constructed using expectation maximization (see Section~\ref{sec:em-truth-quality}).

\section{Free-Text Responses}\label{appendix:free-text-responses}

This section contains questions with free-text responses from the participants, translated into English. If questions refer to former questions, the reference is given as well.

\subsection*{Describe your experiences with the web app. What did you like and what could be improved?}

\begin{enumerate}[nosep]
    \item Keyboard shortcuts
        \\ Full screen utilization
    \item Training beforehand would be good
        \\ Then handling becomes easier over time
    \item Good signal resolution, many adjustment options for evaluation
    \item It is very comprehensive and individually adjustable but the many submenus make it very complicated. The expandable menus are annoying. The "Event created" info is unnecessary.
    \item Initially it seems complicated. Once you get an overview it is very clear. Very detailed and professionally created.
    \item Good
    \item I find the web app user-friendly.
    \item More additional markings desired
\end{enumerate}

\subsection*{Here is space for explanations of your choice in the last question.}
\textit{Reference: Could you imagine using the present system for arousal evaluation in the sleep laboratory?}

\begin{enumerate}[nosep]
    \item Multiple relevant channels at a glance would be better
\end{enumerate}

\subsection*{Here is space for explanations of your choice in the last question.}
\textit{Reference: Which approach do you prefer in connection with AI support: (1) You have access to the AI-determined arousals and explanations from the beginning or (2) You first evaluate manually without AI support and use the AI support for quality control afterwards? Does your preference differ depending on whether the AI support is a Black Box or transparent?}

\begin{enumerate}[nosep]
    \item One actually gets influenced and questions one's decisions again
\end{enumerate}

\subsection*{Has the support from the Black Box AI influenced your decisions? If yes, please give examples. [Other]}

\begin{enumerate}[nosep]
    \item Regarding confidence
    \item Checked own decision again when no arousal was given
    \item one or the other arousal
    \item Generally
    \item positive
\end{enumerate}

\subsection*{Do you have suggestions for improving the Black Box AI support? [Other]}

\begin{enumerate}[nosep]
    \item a bit faster
    \item Programming of AASM rules e.g. no arousals while awake
    \item limitation of the marked area to the blue line
\end{enumerate}

\subsection*{Here is space for explanations of your choice in the last question.}
\textit{Reference: How did your approach differ between the three modes: (1) manual evaluation of arousal events, (2) evaluation with Black Box AI support and (3) evaluation with transparent White Box AI support?}

\begin{enumerate}[nosep]
    \item Decisions become plausible through the other channels and simplify the selection
\end{enumerate}

\subsection*{Justify your answer to the last question.}
\textit{Reference: Did the explanations for the AI-determined arousals help you with their validation?}

\begin{enumerate}[nosep]
    \item The confidence and probability give me some more insight
    \item In case of uncertainty it definitely helped and was often conclusive
    \item it supports and you take a closer look at it again
    \item Due to lack of space, additional channels could simply be viewed in the explanations, especially when the AI has already classified them as relevant, they are very interesting.
    \item The explanations helped but were not decisive in my decisions
    \item Explanation was the confirmation for me
    \item For arousals where I wasn't quite sure
    \item Confidence
\end{enumerate}

\subsection*{Did you gain interesting or unexpected insights from the transparent AI explanations about the AI's approach to arousal determination? If yes: What are they? [Other]}

\begin{enumerate}[nosep]
    \item that the AI really uses all available data
    \item The data that the AI uses for assessment differs greatly from the data that an evaluator normally pays attention to
    \item Arousals are not differentiated here
    \item I was surprised that AI marked arousals that I saw as awake
\end{enumerate}

\subsection*{Could you identify one or more recurring patterns by which the AI recognizes arousals? If yes: What are they? [Other]}

\begin{enumerate}[nosep]
    \item Pulse rate
    \item PLMs (don't always lead to an arousal) and desaturations with breathing events
    \item Heart rate increase
    \item Thorax and abdomen breathing, EMG
    \item EEG
    \item Breathing pattern /PLM
\end{enumerate}

\subsection*{Has the support from the transparent AI influenced your decisions? If yes, please give examples. [Other]}

\begin{enumerate}[nosep]
    \item The explanation based on partially convinced me to score an arousal
    \item occasionally
    \item PLM seen and I didn't but she was right.
    \item Generally
    \item positively influenced
    \item Breathing pattern /PLM
\end{enumerate}

\subsection*{Did you find systematic errors that the AI model makes? If yes: Which ones? [Other]}

\begin{enumerate}[nosep]
    \item Distance of arousals, too long events
    \item Arousals sometimes too close together without enough sleep in between
    \item Arousal drawn but patient awake
    \item see above
\end{enumerate}

\subsection*{Do you have suggestions for improving the transparent AI support?}

\begin{enumerate}[nosep]
    \item pay more attention to EEG frequency increases and only then the other parameters.
    \item better differentiation of arousal
\end{enumerate}

\subsection*{Justify your answer to the last question.}
\textit{Reference: The AI model was optimized to miss as few arousals as possible. For this reason, there are more frequently false positive determined arousals, i.e., the model marks arousals more frequently where there actually are none. For this purpose, the threshold for detecting an arousal was set relatively low. How do you find this approach?}

\begin{enumerate}[nosep]
    \item Too many false arousals. I get somewhat delayed by this
    \item you reflect on your evaluation again and possibly correct yourself again
    \item I score an arousal rather quickly, rather than having to sight-decide and then possibly decide
    \item I'd rather draw a few "forgotten" arousals myself than having to delete many again.
    \item In my opinion, the AI was almost too generous with marking arousals
    \item 95\% correctly marked
    \item Skipped
    \item without justification
\end{enumerate}

\subsection*{When the AI had correctly recognized an arousal, how good was in your opinion generally the agreement of the AI-determined range in which the arousal start likely lies with the actual start of the arousal? Was the range too large or the most likely point (vertical dashed line) too far from the actual start point? Please consider in your answer what degree of agreement is required for the practical usability of the AI support for arousal evaluation. [Other]}

\begin{enumerate}[nosep]
    \item last recording the areas fit perfectly, with the others the point was a bit too far away
\end{enumerate}

\subsection*{If the accuracy was not good enough for you in the last question: How much tighter would the predicted range have to be or how much closer would the most likely point have to be to the actual start point of an arousal?}
\textit{Reference: When the AI had correctly recognized an arousal, how good was in your opinion generally the agreement of the AI-determined range in which the arousal start likely lies with the actual start of the arousal? Was the range too large or the most likely point (vertical dashed line) too far from the actual start point? Please consider in your answer what degree of agreement is required for the practical usability of the AI support for arousal evaluation.}

\begin{enumerate}[nosep]
    \item The start point should be closer to the highest point of the curve, arousals all start too early
    \item very different, but the start point of the arousal is very important
    \item +/- 3 sec
    \item the line was exactly right
    \item 0.3 -0.5 factor
\end{enumerate}

\subsection*{How did your approach differ between the three modes: (1) manual evaluation of arousal events, (2) evaluation with Black Box AI support and (3) evaluation with transparent White Box AI support?}

\begin{enumerate}[nosep]
    \item 1. I thought about what the AI would rate now
        \\ 2. I missed the threshold/confidence
        \\ 3. I could best verify the AI
    \item all 3 the same,
        \\ checked the same for the given arousals
        \\ the others afterwards
    \item Setting the modes and then starting
    \item Not at all, except that I thought about why the blackbox might have marked this now and with the whitebox I looked at why it marked it and then as always decided who was right.
    \item With the White Box AI, the evaluation went much faster because you can rely very much on the AI having marked everything important, unlike manual evaluation
    \item White Box was better for me
    \item Not at all
    \item transparent White Box AI support was my favorite
\end{enumerate}

\subsection*{Which approach do you prefer in connection with AI support: (1) You have access to the AI-determined arousals and explanations from the beginning or (2) You first evaluate manually without AI support and use the AI support for quality control afterwards? Does your preference differ depending on whether the AI support is a Black Box or transparent?}

\begin{enumerate}[nosep]
    \item 1. is certainly easier - for experienced evaluators! (somewhat tempting for newcomers)
    \item would prefer the transparent AI from the beginning
    \item 1. with AI support
    \item 1 and transparent
    \item I prefer approach 1. Yes the preference differs
    \item 2
    \item From the beginning with AI
    \item With AI support more confidence and speed
\end{enumerate}

\subsection*{If you wished for improvements in the last question or would not use the system, justify your answer.}
\textit{Reference: Could you imagine using the present system for arousal evaluation in the sleep laboratory?}

\begin{enumerate}[nosep]
    \item as already mentioned above
    \item shorter markings
    \item speed
    \item The user interface would need to be simplified. Ideally, the AI would integrate into existing evaluation systems.
    \item An option to set preferences regarding how strict the AI is when marking arousals would be very helpful.
    \item more color markings
\end{enumerate}

\subsection*{Would you participate in a similar study again in the future? Why or why not?}

\begin{enumerate}[nosep]
    \item Yes
    \item yes, was very interesting
    \item yes, but crucial is who guides you through the study
    \item Yes, of course.
    \item Yes. Professional development.
    \item Yes
    \item I would like to participate in a similar study
    \item Yes
\end{enumerate}

\end{appendix}
\end{document}